\documentclass[lettersize,journal]{IEEEtran}
\usepackage{amsmath,amsfonts}
\usepackage{array}
\usepackage{caption}
\usepackage{subcaption}
\usepackage[numbers]{natbib}
\usepackage{textcomp}
\usepackage{stfloats}
\usepackage{multicol}
\usepackage{url}
\usepackage{verbatim}
\usepackage{graphicx}
\usepackage{algorithmic}
\usepackage[ruled,vlined]{algorithm2e}
\usepackage{amsmath, amsthm, amssymb}
\newtheorem{theorem}{Theorem}
\newtheorem{proposition}{Proposition}
\newtheorem{definition}{Definition}
\newtheorem{lemma}{Lemma}
\newtheorem{property}{Property}
\newtheorem{corollary}{Corollary}

\def\BibTeX{{\rm B\kern-.05em{\sc i\kern-.025em b}\kern-.08em
    T\kern-.1667em\lower.7ex\hbox{E}\kern-.125emX}}
    
\usepackage{balance}
\begin{document}
\title{Robot Body Schema Learning from Full-body Extero/Proprioception Sensors}
\author{Shuo Jiang, Jinkun Zhang, Lawson Wong
\thanks{Manuscript created October, 2020; This work was developed by the IEEE Publication Technology Department. This work is distributed under the \LaTeX \ Project Public License (LPPL) ( http://www.latex-project.org/ ) version 1.3. A copy of the LPPL, version 1.3, is included in the base \LaTeX \ documentation of all distributions of \LaTeX \ released 2003/12/01 or later. The opinions expressed here are entirely that of the author. No warranty is expressed or implied. User assumes all risk.}}

\markboth{Journal of \LaTeX\ Class Files,~Vol.~18, No.~9, September~2020}%
{Robot Body Schema Learning from Full-body Extero/Proprioception Sensors}

\maketitle

\begin{abstract}
For a robot, its body structure is an \textit{a priori} knowledge when it is designed. However, when such information is not available, can a robot recognize it by itself? In this paper, we aim to grant a robot such ability to learn its body structure from exteroception and proprioception data collected from on-body sensors. By a novel machine learning method, the robot can learn a binary Heterogeneous Dependency Matrix from its sensor readings. We showed such matrix is equivalent to a Heterogeneous out-tree structure which can uniquely represent the robot body topology. We explored the properties of such matrix and the out-tree, and proposed a remedy to fix them when they are contaminated by partial observability or data noise. We ran our algorithm on 6 different robots with different body structures in simulation and 1 real robot. Our algorithm correctly recognized their body structures with only on-body sensor readings but no topology prior knowledge.
\end{abstract}


\section{Introduction}
\noindent
As individuals, humans possess an innate understanding of their bodily configuration characterized by the interlink of four limbs, a central trunk, and a head. Within the human nervous system, the somatosensory cortex plays a central role in the apprehension of such body schema. This term denotes the personal awareness of their body, encompassing precise cognition of their temporal and spatial interrelationships of body parts, and the holistic functional integrity \cite{bullard2013encyclopedia, carter2009human}. The establishment and maintenance of this schema relies on exteroception and proprioception signals acquired from receptors distributed throughout the body. Exteroception encompasses sensory information originating from external sources, including tactile, visual, auditory, olfactory, and other sensory modalities. Proprioception perceives internal body movement and its spatial orientation. The somatosensory cortex synthesizes these incoming sensory signals into a comprehensive representation of the entire body, which encapsulates essential aspects such as the body's topological structure, sensory receptor distribution, posture, and gestures \cite{2015understanding, flor2002phantom}.

\begin{figure}[!t]
 \centering
\includegraphics[width=.8\linewidth]{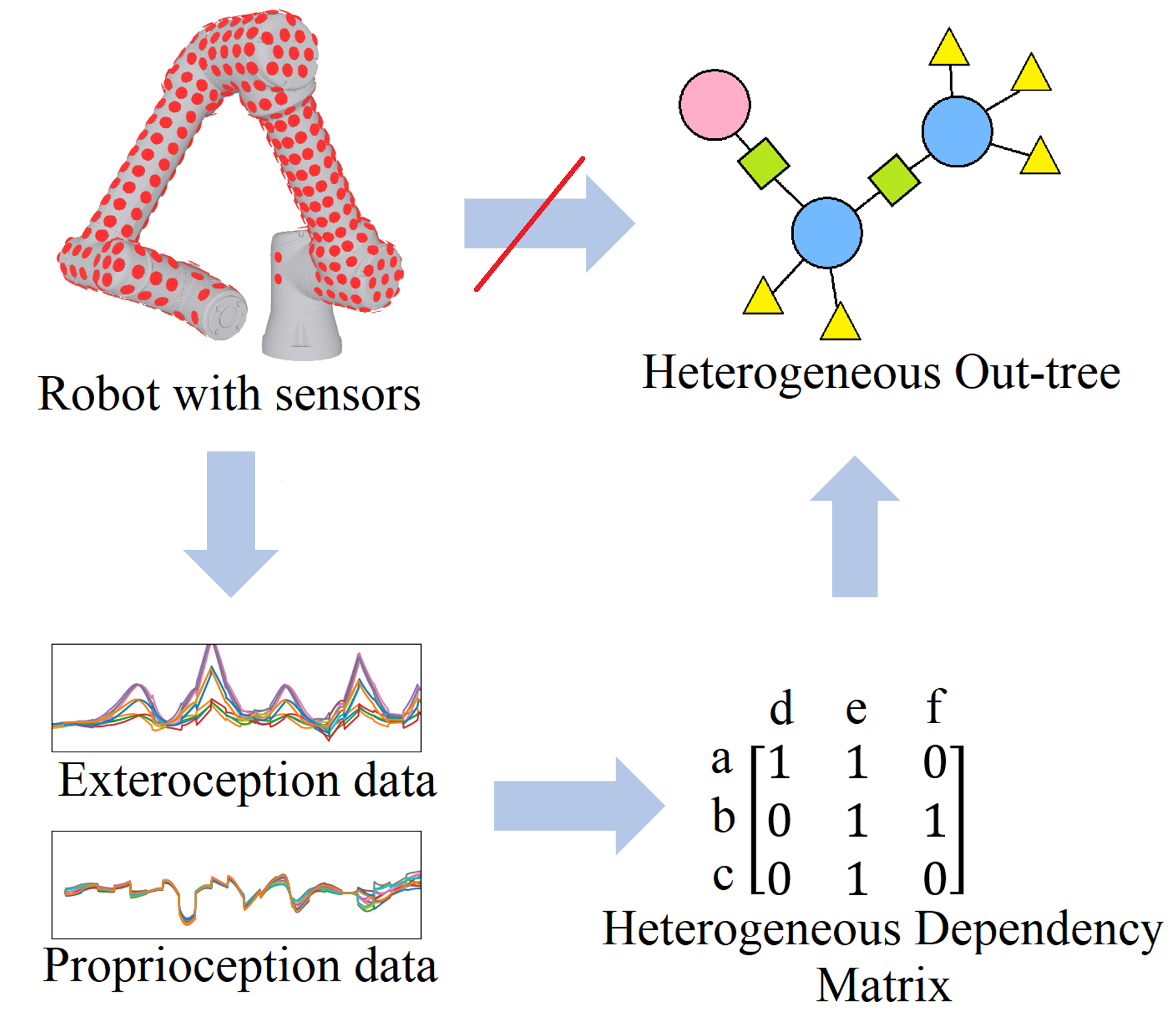}
 \caption{System flowchart: we cannot directly observe robot topology (represented by heterogeneous out-tree), but we can infer it by extracting a heterogeneous dependency matrix from exteroception and proprioception data and use the matrix-tree equivalence to infer the unobserved tree structure.}
 \label{fig_19}
 \vspace{-10pt}
\end{figure}
With the advancement of AI in robotics, we are actively exploring the role of AI in augmenting a robot's environmental and self-awareness, which encompasses the development of novel perception mechanisms and algorithms. We have observed an increasing adoption of whole-body sensors by a diverse range of robots, enabling them to perform tasks such as motion safety \cite{rogelio2020plantar} and object recognition \cite{jiang2022active}. Concurrently, the proliferation of diverse robot morphologies, exemplified by the emergence of reconfigurable robots, has introduced heightened complexities in modeling and kinematic analysis. Motivated by the principles of biological body schema learning, we attempted to propose innovative machine learning theories and algorithms for robot body schema learning, which has been less explored. These innovations are aimed at endowing robots with the capability to autonomously perceive their current body structures in real-time, leveraging information gathered from exteroceptive and proprioceptive sensors distributed across their bodies. The concept of our proposed solution is depicted in Fig. \ref{fig_19}. We assume that the robot's body schema can be effectively described by a heterogeneous out-tree structure on the interconnections among joints, links, and sensors. However, this underlying topology cannot be directly observed (e.g. by ego-centric vision). To overcome this limitation, we employ an array of exteroceptive sensors, such as Inertial Measurement Units (IMUs), and proprioceptive sensors like joint encoders positioned on the robot's body. Through analysis of the sensor data, we can extract a heterogeneous dependency matrix that encapsulates critical information pertaining to the robot's topology. By means of a rigorous equivalence analysis between this matrix and the out-tree structure, the robot's body schema (in the form of the out-tree) can be implicitly discerned.

\section{BACKGROUND}\label{sec_REL}
\noindent
Evidence has substantiated the notion that humans maintain a comprehensive body representation within their brains \cite{head1911sensory}. This representation effectively maps multi-modal sensory inputs, such as tactile, visual, and proprioceptive cues, to the spatial poses of various body parts. Within robotics, the concept of body schema demonstrates an intricate interplay with kinematic identification processes. The acquisition and maintenance of a body schema assume paramount importance, particularly when a robot undergoes alterations in its body morphology due to mechanical failures, repair, or material fatigue. The body schema, in this context, endows the robot with the capacity to discern the morphological alteration and facilitate real-time adjustments in the corresponding body control strategies. This adaptive capability is instrumental in averting potentially catastrophic consequences. For a more comprehensive summary robot body representation, readers can refer to the work by \cite{hoffmann2010body}.

In the study by \cite{mimura2017bayesian}, a Bayesian approach was introduced for the estimation of body schema based on readings from distributed tactile sensors. Their methodology involved an extension of Dirichlet process Gaussian mixture models (DPGMM) \cite{gorur2010dirichlet}, which clusters sensor data while simultaneously learning a latent tree structure that represents the robot's body topology. With their approach only relied on exteroceptive information, our framework assumed that proprioceptive data also encodes crucial information about body structure, offering the potential to simplify the inference process. Additionally in their work, accessing global position and velocity data for each sensor can pose challenges in real-world, which is typically supported by a network of costly base stations \cite{van2018accuracy}. This limitation also constrains vision-based systems, as evidenced in prior research \cite{sturm2009body, chen2022fully}. For instance,  \cite{sturm2009body} developed a Bayesian network to model the forward and inverse kinematic structure of a robot that established the relationship between action commands and body poses. This system relied on vision trackers attached to various links, providing 6-D pose information for inference. Nevertheless, the vision tracking approach was reported with self-occlusion issues. \cite{chen2022fully} employed five depth cameras situated around the robot, in conjunction with joint encoders, to learn the mapping from joint angles to robot morphology. However, their approach focused on learning the robot's body shape, without addressing the topological aspects. Furthermore, \cite{bongard2006automated} introduced an evolutionary algorithm for body schema identification. Their approach utilized multi-modal sensor data, including foot-tip touch sensors, tilt sensors, and body clearance sensors. Their study demonstrated the successful identification of the body structure of a quadrupedal robot but did not present theoretical validation.

\section{METHOD}\label{sec_MET}
\noindent
As our goal is to learn the body structure from exteroception and proprioception data, in subsection \ref{sec_ExEs}, we will use the two data sources to learn a proposed global pose for each sensor. In the second subsection, we show the Jacobian of such global pose with respect to all the joints encodes the dependency information. Such dependencies from all sensors are concatenated as a binary matrix, we name it Heterogeneous Dependency Matrix. The third subsection discusses how such matrix can be learned by a data-driven method. In the fourth subsection, we go back to the robot's body, we show that a tree structure can be extracted from an open-chain robot, we call it Heterogeneous Out-tree. Subsection \ref{sec_TME} shows the matrix and the out-tree are equivalent representations of the robot topology. We proposed two theorems to discuss the conditions when any binary matrix is a Heterogeneous Dependency Matrix. Subsection \ref{sec_PO} discusses the case of partial observability that some links have no sensor mounted, while subsection \ref{sec_contra} discusses another case when an erroneous matrix is extracted, how can it be fixed. The last subsection summarizes our method and shows how such a framework can be used to find robot body topology.
\subsection{Exteroception Estimation}\label{sec_ExEs}
The first thing we need to find is the relation between proprioception and exteroception data in expressing the FK. Then the Jacobian of such FK can encode the dependency information. Assuming we have a second-order differentiable function approximator $f_{\phi }\left ( \theta  \right )$
\begin{equation}
    f_{\phi }\left ( \theta  \right )=\mathbf{T}_{\phi }\left ( \theta  \right )=\begin{bmatrix}
\mathbf{R}\left ( \theta  \right ) & \mathbf{b}\left ( \theta  \right ) \\ 
\mathbf{0} & 1
\end{bmatrix}=\begin{bmatrix}
\mathbf{n}_{x} & \mathbf{n}_{y} & \mathbf{n}_{z} & \mathbf{b}\\ 
0 & 0 & 0 & 1
\end{bmatrix}
\label{eq_8}
\end{equation}
which is parameterized by $\phi$. The function approximator is used to map the input joint angle vector $\theta$ to a homogeneous transformation $\mathbf{T}_{\phi }\left ( \theta  \right )\in \mathbb{R}^{4\times 4}$ in SE(3), that describes the transformation that sensor frame $B$ described in global frame $G$. Reader should keep in mind that for each sensor on the robot body, there will be a unique $f_{\phi }\left ( \theta  \right )$ to represent the pose of the sensor, we only discuss one of them, and the rest are used in the same way. Due to lacking of the knowledge of the robot body structure, the ground-truth homogeneous transformation is unknown, that is why we use such an approximator to learn it from data. We denote the first and second-order derivative of $f_{\phi }\left ( \theta  \right )$ as $\frac{\partial f_{\phi }\left ( \theta  \right )}{\partial \theta}=\mathbf{J}_{\phi}\left ( \theta  \right )\in \mathbb{R}^{4\times 4\times N}$ and $\frac{\partial^{2} f_{\phi }\left ( \theta  \right )}{\partial \theta^{2}}=\mathbf{H}_{\phi}\left ( \theta  \right )\in \mathbb{R}^{4\times 4\times N\times N}$, where $N$ is the number of joints. We can also measure the proprioception as the angles, angular velocities and angular accelerations for all the joint angles (though we do not know their hierarchy) as $\theta$, $\dot{\theta }$, $\ddot{\theta }$. The exteroception data are the vectors $\left [ \alpha _{B},\beta   _{B}\right ]^{T}$ of linear acceleration and angular velocity (by roll-pitch-yaw) described in sensor frame, where $\alpha _{B}\in \mathbb{R}^{3}$, $\beta   _{B}\in \mathbb{R}^{3}$. We will ignore the symbol of dependency of $\theta$ for all variables to make the result neat. The first and second-order time derivative of $f_{\phi }\left ( \theta  \right )$ are shown in Equations \ref{eq_12} and \ref{eq_7} as $\frac{\partial f_{\phi } }{\partial t}\in \mathbb{R}^{4\times 4}$ and $\frac{\partial ^{2}f_{\phi }}{\partial t^{2}}\in \mathbb{R}^{4\times 4}$.
\begin{equation}
    \frac{\partial f_{\phi } }{\partial t}=\begin{bmatrix}
\mathbf{\dot{R}} & \mathbf{\dot{b}}\\ 
\mathbf{0} & 1
\end{bmatrix}=\mathbf{J}_{\phi }\dot{\theta}
\label{eq_12}
\end{equation}
\begin{equation}
    \begin{split}
        \frac{\partial ^{2}f_{\phi }}{\partial t^{2}}=\begin{bmatrix}
\mathbf{\ddot{R}} & \mathbf{\ddot{b}}\\ 
\mathbf{0} & 1
\end{bmatrix}=\mathbf{H}_{\phi }\dot{\theta}\dot{\theta}+\mathbf{J}_{\phi }\ddot{\theta}
    \end{split}
    \label{eq_7}
\end{equation}
The two equations show that we can estimate $\mathbf{\dot{R}}$, $\mathbf{\dot{b}}$, $\mathbf{\ddot{R}}$ and $\mathbf{\ddot{b}}$ by exteroception and proprioception data. The next step is to find the relation between $\mathbf{R}$, $\mathbf{b}$, $\mathbf{\dot{R}}$, $\mathbf{\dot{b}}$, $\mathbf{\ddot{R}}$,  $\mathbf{\ddot{b}}$ and  $\left [ \alpha  _{B}, \beta  _{B}\right ]^{T}$.

For any point $r$ in space, we have
\begin{equation}
    r_{G}=\mathbf{T}_{\phi }\cdot r_{B}
\end{equation}
$r_{G}$ and $r_{B}$ are coordinates of the same $r$ in global and sensor frame respectively. By taking the second-order time derivative of such relation, we have the acceleration of point $r$ in sensor frame:
\begin{equation}
\small
    \begin{split}
        a_{B} = \mathbf{T}_{\phi }^{-1}a_{G}=\mathbf{T}_{\phi }^{-1}\ddot{r}_{G}=\mathbf{T}_{\phi }^{-1}\cdot \mathbf{\ddot{T}}_{\phi }\cdot r_{B}
=\begin{bmatrix}
\mathbf{R}^{T}\mathbf{\ddot{R}} & \mathbf{R}^{T}\mathbf{\ddot{b}}\\ 
\mathbf{0} & 0
\end{bmatrix}\cdot r_{B}
    \end{split}
\end{equation}
We conclude that the linear part $\alpha  _{B}$ equals $\mathbf{R}^{T}\mathbf{\ddot{b}}$ because they are both in $\mathbb{R}^{3}$. Then we need to find the angular part. Define operation
\begin{equation}
    \omega  _{B}^{\wedge }=\begin{bmatrix}
w_{1}\\ 
w_{2}\\ 
w_{3}
\end{bmatrix}^{\wedge }=\begin{bmatrix}
0 & -w_{3} & w_{2}\\ 
w_{3} & 0 & -w_{1}\\ 
-w_{2} & w_{1} & 0
\end{bmatrix}
\end{equation}

To find the expression of $\beta_{B}$, we already know the expression of angular velocity in body frame, as $\omega  _{B}^{\wedge }=\mathbf{R}^{T}\mathbf{\dot{R}}$.
However, the corresponding vector $\omega  _{B}$ is not $\beta _{B}$ because $\omega  _{B}$ is the rotation velocity vector described by axis-angle system and $\beta _{B}$ is the rotation velocity vector described by roll-pitch-yaw system. To align the two representations, we know the two velocity vectors should cause the same rotation effect in a short period of time $T_{s}$. For $\omega  _{B}$, it can be decomposed to a time derivative of an angle $\dot{\vartheta }\in \mathbb{R}$ and a rotation axis vector $\vec{u}\in \mathbb{R}^{3}$, with $\omega  _{B}=\dot{\vartheta }\vec{u}$ and $\left |\vec{u}   \right |=1$. The rotated angle in $T_{s}$ can be approximated as $\dot{\vartheta }T_{s}$. By Rodrigues' rotation formula, the corresponding rotation matrix in $T_{s}$ is 
\begin{equation}
    \mathbf{R}_{1}=\mathbf{I}+\textup{sin}\left ( \dot{\vartheta }T_{s} \right )\omega  _{B}^{\wedge }+\left [ 1-\textup{cos}\left (  \dot{\vartheta }T_{s}\right ) \right ]\omega  _{B}^{\wedge }\cdot \omega  _{B}^{\wedge }
\end{equation}
For $\beta _{B}$, it represents the roll, pitch, yaw velocities as $\beta _{B} = \left [  \beta _{B1} ,\beta _{B2} ,\beta _{B3} \right ]^{T}$. The rotated angle in $T_{s}$ can be approximated by $\left [  \beta _{B1} \cdot T_{s} ,\beta _{B2}\cdot T_{s} ,\beta _{B3}\cdot T_{s} \right ]^{T}$, so the rotation matrix can be calculated from standard roll-pitch-yaw rotation matrices as
\begin{equation}
    \begin{split}
        \mathbf{R}_{2}&=\mathbf{R}_{z}\left ( \beta _{B3}\cdot T_{s} \right )\mathbf{R}_{y}\left ( \beta _{B2}\cdot T_{s} \right )\mathbf{R}_{x}\left ( \beta _{B1}\cdot T_{s} \right )
    \end{split}
\end{equation}
$\mathbf{R}_{z}$, $\mathbf{R}_{y}$ and $\mathbf{R}_{x}$ are rotation matrices w.r.t. z, y and x axes. For who is not familiar with the context, one can refer to Appendix \ref{sec_APP1} for a more detailed derivation. Apparently $\mathbf{R}_{1}$ and $\mathbf{R}_{2}$ should be the same matrix because they express the same rotation effect in $T_{s}$. As $\mathbf{R}_{1}$ is proposed from $f_{\phi }\left ( \theta  \right )$ and $\mathbf{R}_{2}$ is calculated from measurement $\beta _{B}$, we can learn the parameter $\phi$ by minimizing their difference. For any rotation matrix $\mathbf{R}$, we have $\textup{tr}\left ( \mathbf{R} \right )=1+2\textup{cos}\left ( \rho  \right )$, $\rho$ is the angle of rotation described in angle-axis system, so
\begin{figure}[!t]
 \centering
 \includegraphics[width=.7\linewidth]{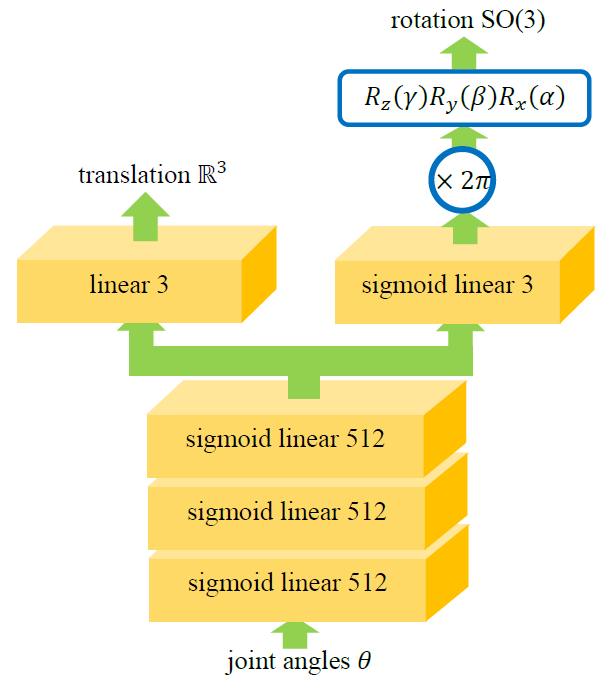}
 \caption{Neural network structure of $f_{\phi }\left ( \theta  \right )$. The roll-pitch-yaw angles to rotation matrix transformation can be found in Equation \ref{eq_18}. The result will be reshaped to a homogeneous transformation matrix of 4 $\times $ 4 by padding zeros and ones as Equation \ref{eq_8}.}
 \label{fig_35}
\end{figure}
\begin{equation}
    \rho=\textup{arccos}\frac{\textup{tr}\mathbf{R}-1}{2}
\end{equation}
In our case, matrix $\mathbf{R}$ is the relative rotation between $\mathbf{R}_{1}$ and $\mathbf{R}_{2}$ which is $\mathbf{R}=\mathbf{R}_{1}^{T}\mathbf{R}_{2}$. When the two matrices are the same, we have $\rho=0$.

Thus, we can find the optimal parameter $\phi^{*}$ by
\begin{equation}
    \phi^{*} =\underset{\phi}{\textup{argmin}}\left [\left | \alpha  _{B}-\mathbf{R}^{T}\mathbf{\ddot{b}} \right |+\textup{arccos}\frac{\textup{tr}\left (\mathbf{R}_{1}^{T}\mathbf{R}_{2}  \right )-1}{2}  \right ]
    \label{eq_9}
\end{equation}
Using gradient descent method can potentially cause some numerical problem by $\textup{arccos}$ term. In practice, we found using the following optimization can also have good performance.\\
\begin{figure*}[!t]
 \centering
 \includegraphics[width=\linewidth]{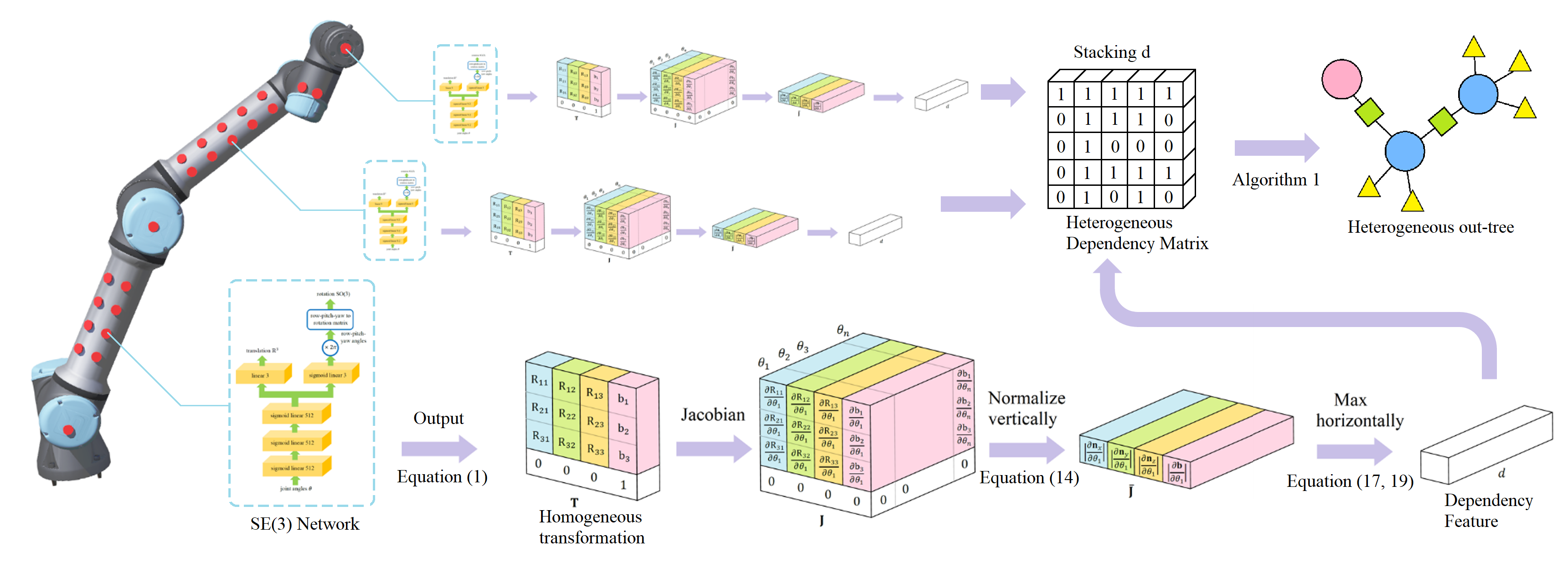}
 \caption{Pipeline to extract robot body structure (represented by a heterogeneous out-tree). (left) Each IMU is linked to a SE(3) neural network as shown in Figure \ref{fig_35} to approximate its global pose given the joint angles. (bottom) each neural network $f_{\phi }\left ( \theta  \right )$ outputs the current homogeneous transformation $\mathbf{T}$; The 'Jacobian' step extracts Jacobian matrix $\mathbf{J}$ of size (4, 4, N) from $\mathbf{T}$; normalize by each column to get Transform Invariant Jacobian $\overline{\mathbf{J}}$; extract Dependency Feature $d$ from $\overline{\mathbf{J}}$ by Equation \ref{eq_10}. (right) stack all Dependency Features from all sensors and merge the duplicates generates a heterogeneous dependency matrix; running algorithm \ref{algo_3} to generate the out-tree from the matrix.}
 \label{fig_34}
\end{figure*}
\begin{equation}
    \phi^{*} =\underset{\phi}{\textup{argmin}}\left [\left | \alpha  _{B}-\mathbf{R}^{T}\mathbf{\ddot{b}} \right |-\textup{tr}\left (\mathbf{R}_{1}^{T}\mathbf{R}_{2}  \right )  \right ]
    \label{eq_20}
\end{equation}
The detailed derivation of this section can be seen in Appendix \ref{sec_APP1}.

In our work, we use a neural network to represent $f_{\phi }\left ( \theta  \right )$, whose structure is shown in Figure \ref{fig_35}. The input of the neural network is the joint angle vector of $\mathbb{R}^{n}$. Through three hidden layers with sigmoid activation function, the translation branch consists of 3 linear neurons predicting global translation in $\mathbb{R}^{3}$. The rotation branch consists of 3 linear neurons predicting the roll-pitch-yaw angles $\left [ \alpha ,\beta ,\gamma  \right ]^{T}$ in $\left [ 0,2\pi  \right ]$. Such angles will by transformed to a rotation matrix with
\begin{equation}
    \begin{split}
        \mathbf{R}_{2}&=\mathbf{R}_{z}\left ( \gamma  \right )\mathbf{R}_{y}\left ( \beta  \right )\mathbf{R}_{x}\left ( \alpha  \right )
    \end{split}
    \label{eq_18}
\end{equation}
The rotation part and translation part of the neural network output will be concatenated as a homogeneous transformation matrix as shown in Equation \ref{eq_8}. Notice that for each exteroception sensor (IMU), there will be a unique neural network attached to it (shown in Figure \ref{fig_34} left). It means if there is 1000 sensors, there will be 1000 such networks trained in parallel. It also indicates the system is linearly scalable with the increasing of number of sensors.
\subsection{Heterogeneous Dependency Matrix} \label{sec_HDM}
In last subsection, we have already learned a proposed homogeneous transformation $\mathbf{T}=f_{\phi }\left ( \theta  \right )$ to approximate the global pose of the sensor linked with it. However, as we have not specified the reference frame, the value of such homogeneous transformation can be arbitrary. We can assume the Jacobian of $f_{\phi }\left ( \theta  \right )$ can encode the true dependency information of how the change of each joint angle affects the change of output, but it is also reference frame dependent. However, when $\mathbf{T}$ is irrelevant to some joint value $\theta$, the corresponding term in Jacobian matrix should be zero, and that is reference frame independent. In this section, we show a way to extract the reference frame independent Jacobian representation. The concatenation of such representations from all sensor frames can form a new matrix, which records the global dependency information.

\begin{definition}
If we have homogeneous transformation $\mathbf{T}$ as defined in Equation \ref{eq_8}, the homogeneous Jacobian w.r.t configuration $\theta$ can be represented as a $12\times N$ matrix as:\\
\begin{equation}
    \frac{\partial \mathbf{T}}{\partial \theta }=\left [\frac{\partial \mathbf{n}_{x}}{\partial \theta },\frac{\partial \mathbf{n}_{y}}{\partial \theta },\frac{\partial \mathbf{n}_{z}}{\partial \theta }, \frac{\partial \mathbf{b}}{\partial \theta } \right ]^{T}
\end{equation}
The Transform Invariant Jacobian is defined as a $4\times N$ matrix as:
\begin{equation}
    \overline{\mathbf{J}}=\left [\left |\frac{\partial \mathbf{n}_{x}}{\partial \theta }  \right |,\left |\frac{\partial \mathbf{n}_{y}}{\partial \theta }  \right |,\left |\frac{\partial \mathbf{n}_{z}}{\partial \theta }  \right |, \left |\frac{\partial \mathbf{b}}{\partial \theta }  \right | \right ]^{T}
    \label{eq_11}
\end{equation}
$\left | \cdot  \right |$ is to normalize by each column. $N$ is the number of joints.
\label{def_1}
\end{definition}
We show (in Appendix \ref{sec_APP2}) that $\overline{\mathbf{J}}$ is invariant to the choice of the reference frame, which means no matter where the sensor is mounted, as long as it is mounted on the same link, such Jacobian should keep the same.
\begin{proposition}
Transform Invariant Jacobian is invariant to any joint-independent transformation $\mathbf{Y}\in SE\left ( 3 \right )$.
\label{prop_2}
\end{proposition}

As Jacobian encodes the dependency of change of pose with respect to joints, we assume it to be a feature of sensor dependency that carries topology information. We can transform such feature into a more concise representation. The idea is we only care about if a frame $i$ has dependency on joint $\theta _{j}$ or not. In such case, we can turn Transform Invariant Jacobian to a binary representation.
\begin{definition}
There is one corresponding Dependency Feature $\mathbf{d}\in \mathbb{R}^{N}$ to each Transform Invariant Jacobian $\overline{\mathbf{J}}$. Dependency Feature is defined as $\mathbf{d}_{n}=0$ if all the elements in $n$-th column of $\overline{\mathbf{J}}$ are zeros. Else $\mathbf{d}_{n}=1$.
\label{def_3}
\end{definition}
For example, 
\begin{equation}
    \overline{\mathbf{J}}=\begin{bmatrix}
3.14 &6.23  & 0 &1.56 \\ 
7.23 &2.64  &0  & 3.25\\ 
2.33 &3.08  & 0 & 6.32\\ 
7.33 & 8.32 &  0&9.17 
\end{bmatrix}\rightarrow \mathbf{d}=\left [ 1, 1, 0,1 \right ]
\label{eq_10}
\end{equation}
Notice that each column of $\mathbf{d}$ has a unique label $\theta _{j}$. The "1" positions in Dependency Feature encodes all the joints that the sensor frame passes as it traverses to the root. The pipeline to extract Dependency Feature is shown in Figure \ref{fig_34} bottom.

\begin{definition}
\textbf{Heterogeneous Dependency Matrix} $\mathbf{D}$ is defined as stacking Dependency Feature $\mathbf{d}$ of all the sensors (not on root node) as rows in a matrix, with duplicate rows merged (only one left). 
\label{def_2}
\end{definition}
The Dependency Feature stacking is shown in Figure \ref{fig_34} right, forming a Heterogeneous Dependency Matrix. The duplicate rows in such matrix are introduced by two sensors mounted on the same link, which we know from proposition \ref{prop_2}. Dependency features of different sensors on the same link are identical because the sensors traverse the same set of joints to the root. Merging the duplicate rows ensures the redundant dependency information is eliminated.

Heterogeneous dependency matrix is a labelled matrix. In our case, the row labels are different sensors and the column labels are different joints. We also have to ensure each entity of sensors and joints are physically distinguishable (apparent in our case because they are measured).
\begin{property}
Rows and columns of $\mathbf{D}$ are interchangeable.
\label{per_1}
\end{property}
\begin{property}
Rows and columns of $\mathbf{D}$ are non-duplicate.
\label{per_3}
\end{property}
Property \ref{per_1} is obvious because rows are labels of sensors without order. For columns, because each row is a Dependency Feature records the binary dependency of joint nodes on the route from the sensor to the root, this dependency is agnostic to the order of joints. Notice that interchanging rows and columns of $\mathbf{D}$ is accompanied by interchanging their labels. Because the rows and columns of $\mathbf{D}$ are labelled, we assume any permutation of rows and columns represents the same matrix (Figure \ref{fig_6}).
Property \ref{per_3} is shown in Appendix \ref{sec_APP3}.

\begin{figure}[!t]
 \centering
 \includegraphics[width=.45\linewidth]{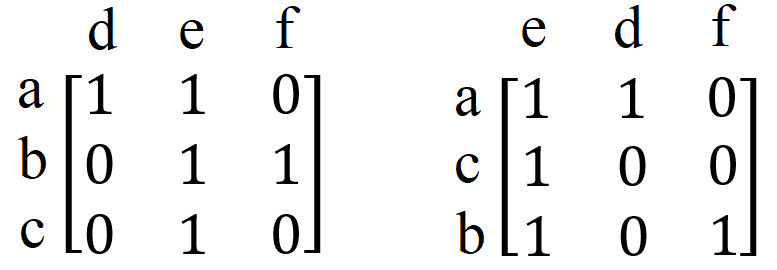}
 \caption{Two matrices transformed by switching rows [b,c] and columns [d,f]; we assume they are equivalent.}
 \label{fig_6}
 \vspace{-10pt}
\end{figure}

\begin{figure*}[!t]
\centering
\begin{subfigure}{.2\linewidth}
    \centering
    \includegraphics[height=.08\textheight]{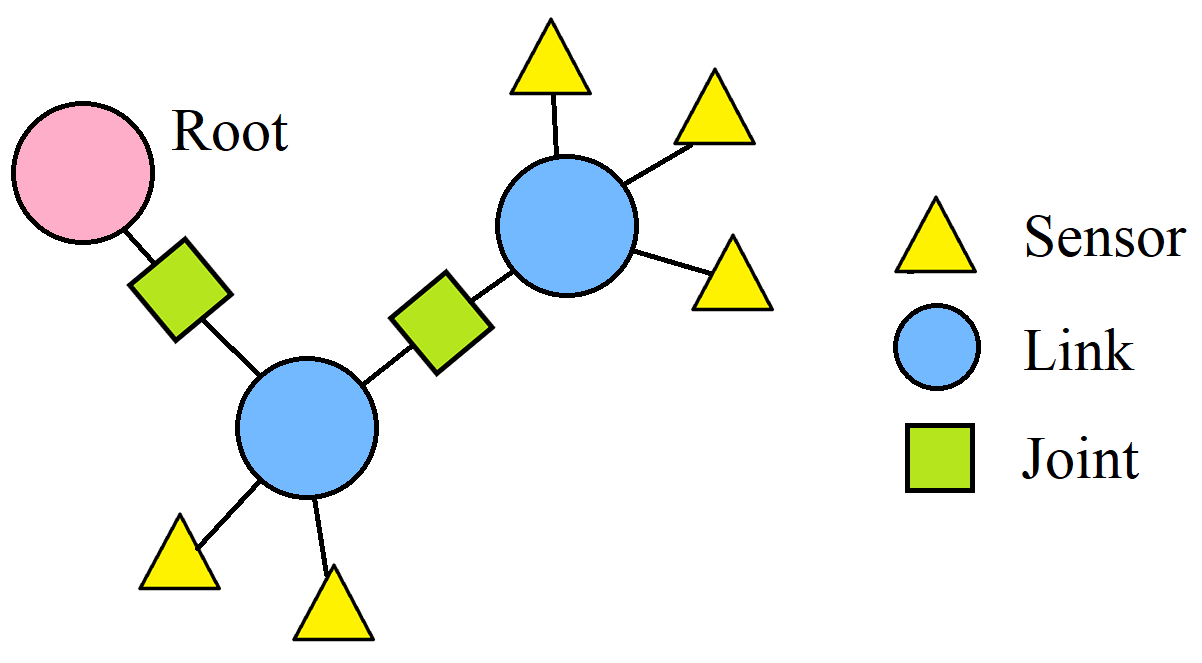}
    \caption{}
    \label{fig_2}
\end{subfigure}
\begin{subfigure}{.45\linewidth}
    \centering
    \includegraphics[height=.08\textheight]{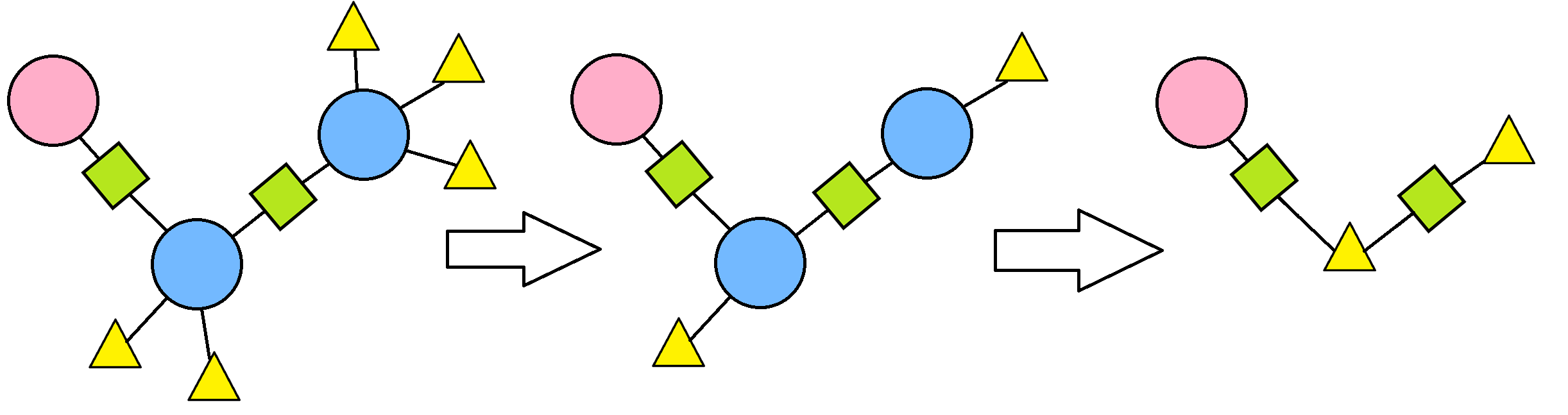}
    \caption{}
    \label{fig_3}
\end{subfigure}
\begin{subfigure}{.25\linewidth}
    \centering
    \includegraphics[height=.08\textheight]{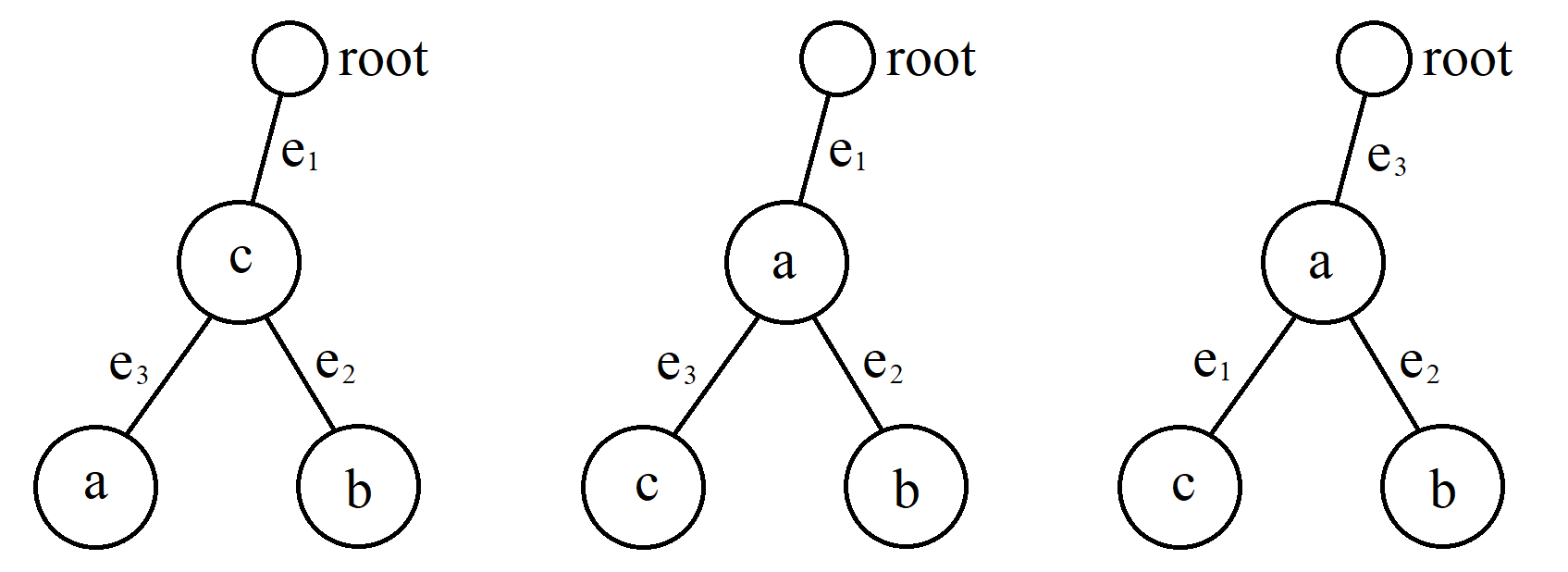}
    \caption{}
    \label{fig_4}
\end{subfigure}
\caption{(a) Topology of open-chain robot. (b) Topology simplification of a robot. The first arrow is because different sensors on the same link provide redundant joint-dependency information. The second arrow is because link and sensor are one-one pair, so it is equivalent to discuss either. (c) Examples of heterogeneous out-tree.}
\end{figure*}

\subsection{Data-driven Remedy} \label{sec_DDR}
Before we continue to the next section, we need to solve some problems when we learn and extract heterogeneous dependency matrix by the data-driven method. The first problem we are faced with is that the Transform Invariant Jacobian is actually a function of $\theta$. However, what we need is a real number matrix as in Example \ref{eq_10}. Here comes a question that which value of $\theta$ we choose to calculate such matrix. Here, we independently sampled $\theta$s, e.g., from a random joint motion or uniformly sample in configuration space and then calculate the variance of their corresponding Jacobians.
\begin{equation}
\begin{split}
    \overline{\mathbf{J}} &= \underset{\theta}{\mathbb{E}}\left [ \left ( \overline{\mathbf{J}}_{\phi}\left ( \theta \right )- \underset{\theta}{\mathbb{E}}\left [ \overline{\mathbf{J}}_{\phi}\left ( \theta \right ) \right ]\right )^{2} \right ] \\
    &= \frac{1}{n}\sum_{i=1}^{n}\left [ \overline{\mathbf{J}}_{\phi}\left ( \theta _{i} \right )-\frac{1}{n}\sum_{i=1}^{n}\overline{\mathbf{J}}_{\phi}\left ( \theta _{i} \right ) \right ]^{2} \nonumber
\end{split}
\label{eq_22}
\end{equation}

We can also use mean or median value, but we found using variance provides the most stable result. The idea behind this is when sensor $i$'s movement is independent of joint $j$, the $j$'s column of $\overline{\mathbf{J}}$ should be all-zero, no matter what $\theta$ is. Thus, the variance of such column must also be zero. Otherwise, there must be some $\theta$s that cause the $j$'s column of $\overline{\mathbf{J}}$ non-zero, then the variance is also non-zero. So, variance also retains the dependency information that Jacobian carries.

Then, we should discuss the problem of Jacobian $\mathbf{J}_{\phi }$ and its underlying function approximator $f_{\phi }$. In ideal case, $f_{\phi }\left ( \theta  \right )$ perfectly catches the FK and the Transform Invariant Jacobian $\overline{\mathbf{J}}$ is noiseless. However noise could be naturally brought by data that the all-0 columns in $\overline{\mathbf{J}}$ will not be zero (as in Example \ref{eq_10}), or duplicate rows can be non-duplicate, which means Heterogeneous Dependency Matrix can have more rows than columns.

For the first problem, we can filter $\overline{\mathbf{J}}$ by a max function instead of Example \ref{eq_10} and normalize such feature.
\begin{equation}
    \mathbf{d}' =\underset{row}{\textup{max}}\overline{\mathbf{J}}/\left | \underset{row}{\textup{max}}\overline{\mathbf{J}} \right |
\end{equation}
For example, 
\begin{equation}
   \begin{split}
        &\overline{\mathbf{J}}=\begin{bmatrix}
3.14 &6.23  & 0 &1.56 \\ 
7.23 &2.64  &0  & 3.25\\ 
2.33 &3.08  & 0 & 6.32\\ 
7.33 & 8.32 &  0&9.17 
\end{bmatrix} \rightarrow \mathbf{d}'=\left [ 7.33, 8.32, 0, 9.17 \right ]\\
&\rightarrow \mathbf{d}'=\left [ 0.5, 0.57, 0, 0.63 \right ]
   \end{split}
\end{equation}
Then we can use a threshold function to get $\mathbf{d}$ as
\begin{equation}
    \mathbf{d}_{i}=\left\{\begin{matrix}
1&\textup{if} \; \mathbf{d}'_{i}>\delta  \\ 
0&\textup{otherwise}
\end{matrix}\right.
\end{equation}

For the second problem (more rows than columns), we can use Dirichlet Process Gaussian Mixture Model (DPGMM \cite{gorur2010dirichlet}) to cluster all rows $\mathbf{d}$, and sample rows (no more than the number of columns) from the DPGMM. Notice the sampled rows are not binary vectors, we have to filter them with a threshold of 0.5 (because the rows in each cluster are all binary). Then the sampled rows form a new Heterogeneous Dependency Matrix.

A new problem is introduced that what is the optimal value of threshold $\delta$? The criterion of choosing $\delta$ is that $\delta$ should ensure among clusters more separated and in each cluster less dispersed. Also, it should ensure the numbers of samples in different clusters are similar. We propose to optimize $\delta$ as
\begin{equation}
\begin{split}
    \mathbf{S}_{i,j} &= \left | \mu _{i}-\mu _{j} \right | \\
    m &= \sum _{\forall c}\sum _{x_{i}\in c}\left ( x_{i}-\mu _{c} \right )^{T}\left ( x_{i}-\mu _{c} \right ) \\
    p_{c} &= \frac{n_{c}}{\sum _{c}n_{c}} \\
    \delta^{*} &= \underset{\delta}{\textup{argmax}}\frac{det\left ( \mathbf{S} \right )}{m}-\lambda \sum _{c}p_{c}\cdot \textup{log}\left (p_{c}  \right )
\end{split}
    \label{eq_23}
\end{equation}

$\lambda$ is a weighting factor. Here, matrix $\mathbf{S}$ stores the distance information between means of any two clusters. $m$ captures the dispersion of data in each cluster by summing up the dispersion in each cluster. $p_{c}$ is the number of data points in cluster $c$ based on the total number of data points. The optimal $\delta^{*}$ maximizes $\mathbf{S}$ to separate clusters, minimizes $m$ to reduce data dispersion in each cluster, and maximizes the entropy of $p_{c}$ to make numbers of samples in different clusters similar.

\subsection{Heterogeneous Out-tree} \label{sec_HOT}
In the previous section, we discussed how to extract heterogeneous dependency matrix $\mathbf{D}$ from data, we aim to find the relation between $\mathbf{D}$ and the robot forward kinematics (body structure). For an open-chain robot with sensors mounted on the body links, we can have its body representation as in Figure \ref{fig_2}.
We can easily find each dependency feature actually encodes all the joint nodes on the route from the sensor to the root. If we merge all the same rows, the equivalent robot structure is that there is exactly one sensor on each link. In such case, the sensors and links are one-one pair and rigidly mounted, so they will share the same dependency. Then we do not need to discuss three different types of nodes in such topology, but only two types are enough (joints and sensors). Such simplification is shown in Figure \ref{fig_3}. We keep the sensor nodes in the final tree structure because they are directly observable, but not for link nodes. Notice such tree has a direction from root to all leaf nodes but is not shown in the figure. Also, every sensor node or joint node is unique (reader can imagine they are labelled), which means if any two nodes switch, then it results in a different tree. We can also assume sensors are nodes and joints are edges.

\begin{definition}
\textbf{Heterogeneous Out-Tree} is defined as an out-tree with all edges labelled.
\label{def_4}
\end{definition}
Different from out-trees, heterogeneous out-tree not only considers the parent of each node, but also considers via which particular edge, the node inherits from its parent. Figure \ref{fig_4} shows some heterogeneous out-trees with nodes $\left [ a,b,c,\cdots  \right ]$ and edges $\left [ e_{1},e_{2},e_{3},\cdots  \right ]$, they are different trees.
\begin{definition}
Two heterogeneous out-trees are semi-equivalent if any node with the same label in both trees has the same parent node.
\label{def_7}
\end{definition}
Semi-equivalent tree neglects the dependency for nodes of edges. For example, tree 2 and tree 3 in Figure \ref{fig_4} are semi-equivalent but not equivalent.
\begin{property}
In a heterogeneous out-tree, the number of nodes is $N+1$ and the number of edges is $N$.
\label{per_2}
\end{property}
\begin{property}
$\sum _{j}\mathbf{D}_{ij}$ is the depth of node $i$ in the tree.
\label{per_4}
\end{property}
\begin{property}
There are $\left ( N+1 \right )^{N-1}\cdot N!$ different tree structures with $N$ nodes.
\label{per_5}
\end{property}
Property \ref{per_5} is proved in Appendix \ref{sec_APP4}. The following lemma (proof in Appendix \ref{sec_APP5}) links such an out-tree with heterogeneous dependency matrix introduced in section \ref{sec_HDM}.
\begin{lemma}
From each heterogeneous out-tree, one can extract a determined heterogeneous dependency matrix.
\label{lemma_1}
\end{lemma}

\begin{figure*}[!t]
\hfill
\begin{subfigure}{.35\linewidth}
    \centering
    \includegraphics[height=.11\textheight]{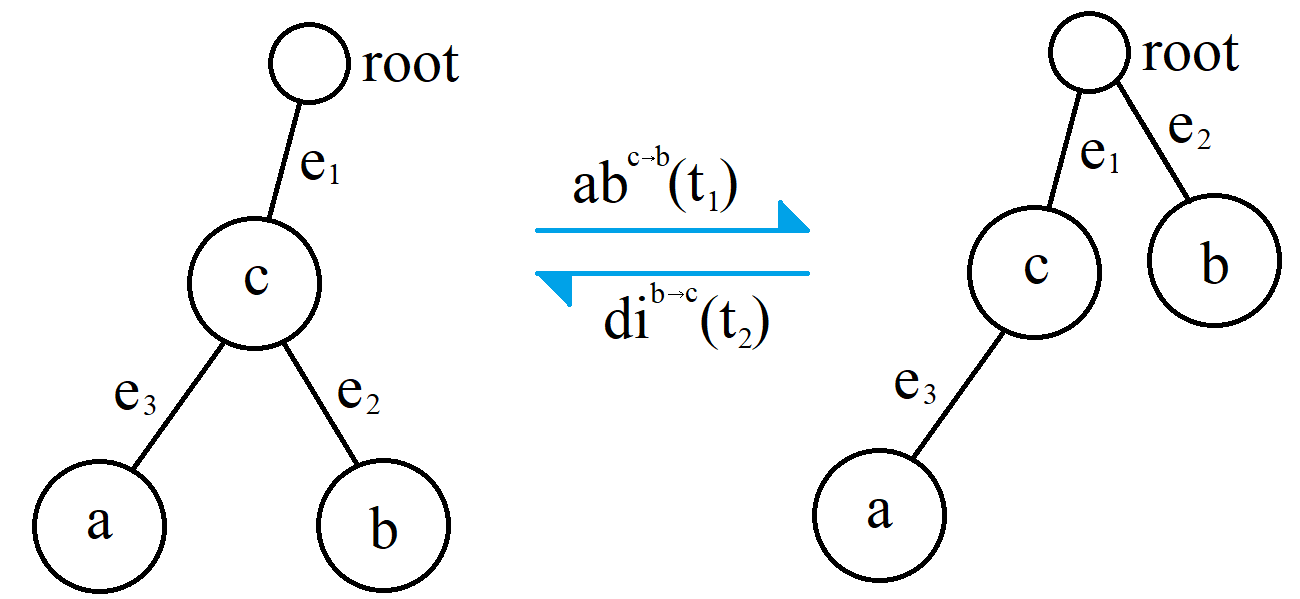}
    \caption{}
    \label{fig_5}
\end{subfigure}
\hfill
\begin{subfigure}{.6\linewidth}
    \centering
    \includegraphics[height=.11\textheight]{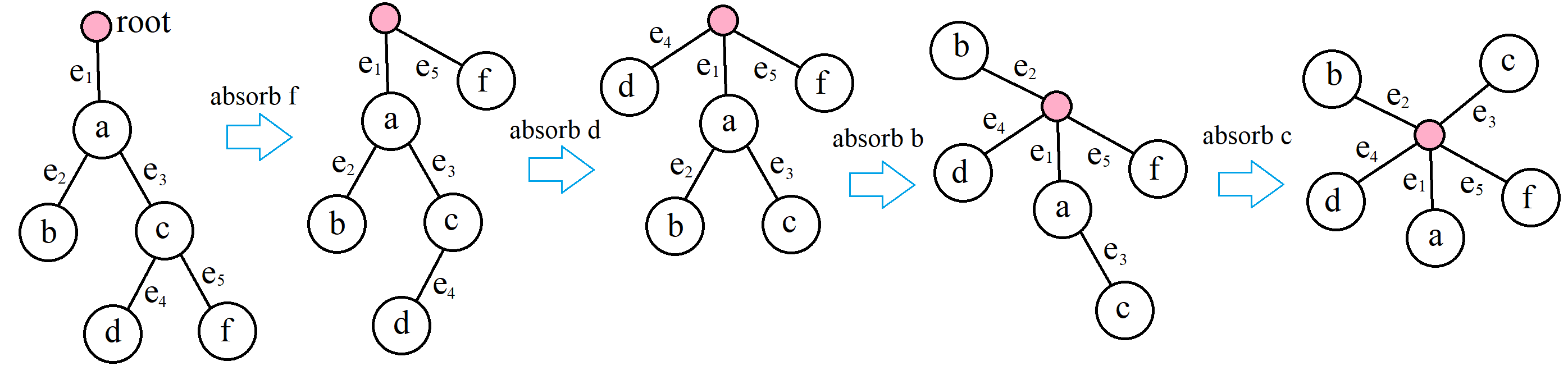}
    \caption{}
    \label{fig_55}
\end{subfigure}
\hfill
\caption{(a) Demonstration of dilation $di^{i\rightarrow j}\left ( t \right )$ and absorption $ab^{j\rightarrow i}\left ( t \right )$. (b) Demonstration of using absorption operation to transform a tree to a depth-1 tree.}
\end{figure*}


We define the set of all possible topologies of heterogeneous out-tree consisting of $N$ nodes as $\mathbb{T}_{N}$.  
\begin{definition}
We define two operations on set $\mathbb{T}_{N}$. For any heterogeneous out-tree $t\in \mathbb{T}_{N}$:\\
\textbf{Dilation}: Given a node $i$ whose direct parent node is root, direct parent edge $e_{i}$, and given another node $j$, the dilation operation $di^{i\rightarrow j}\left ( t \right )$ moves $e_{i}$ to the direct child edge of $j$. \\
\textbf{Absorption}: Given a leaf node $i$, and direct parent edge is $e_{i}$ whose direct parent node is $j$ (non-root), the absorption operation $ab^{j\rightarrow i}\left ( t \right )$ moves $e_{i}$ to the direct child edge of root.
\label{def_5}
\end{definition}
Figure \ref{fig_5} illustrates an example of dilation and absorption. The following properties are proved in Appendices \ref{sec_APP6}-\ref{sec_APP8}.
\begin{lemma}
$ab^{j\rightarrow i}\left ( t \right )$ and $di^{i\rightarrow j}\left ( t \right )$ on heterogeneous out-tree set are a pair of inverse operation.
\label{lemma_2}
\end{lemma}
\begin{proposition}
Any tree $t$ can be transformed to a depth-1 tree by a series of absorption and from the depth-1 tree, reconstruct the original tree $t$.
\label{prop_3}
\end{proposition}
\begin{corollary}
Tree set $\mathbb{T}_{N}$ is closed under dilation and absorption.
\label{cor_1}
\end{corollary}

In the following sections, we will shown how the heterogeneous dependency matrix and heterogeneous out-tree are related.
\subsection{Tree-Matrix Equivalence}\label{sec_TME}
\begin{figure}[!t]
 \centering
 \includegraphics[width=.5\linewidth]{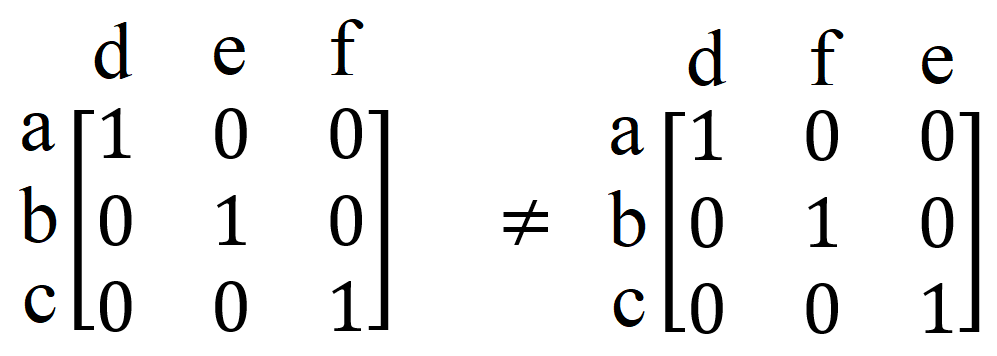}
 \caption{Different permutation matrices}
 \label{fig_8}
\end{figure}

We now discover if heterogeneous dependency matrix and heterogeneous out-tree have some shared properties. We define the set of all binary matrices $\left \{ 0,1 \right \}^{N\times N}$ as $\mathbb{S}_{N}$. Clearly, the matrices with corresponding heterogeneous out-trees consist of only a subset of $\mathbb{S}_{N}$. For example matrix
\begin{equation}
    \begin{bmatrix}
1 & 1 & 0\\ 
0 & 1 & 1\\ 
1 & 0 & 1
\end{bmatrix}
\end{equation}
has no corresponding heterogeneous out-tree. We define set of all binary matrices of $N\times N$ having corresponding heterogeneous out-trees as $\mathbb{D}_{N}$. $\mathbb{D}_{N}\subseteq \mathbb{S}_{N}$.
\begin{definition}
We define two operations on set $\mathbb{S}_{N}$. For any matrix $\mathbf{S}\in \mathbb{S}_{N}$: \\
\textbf{Absorption}: For any row $i$ and $j$, $i\neq j$, perform an XOR operation, $\mathbf{s}'=\mathbf{S}_{i}\oplus \mathbf{S}_{j}$. If $\sum \mathbf{s}'=1$ and $\sum \mathbf{S}_{i} > \sum \mathbf{S}_{j}$, absorption $ab^{j\rightarrow i}\left ( \mathbf{S} \right )$ makes $\mathbf{S}_{i}:= \left (\mathbf{S}_{i}- \mathbf{S}_{j}  \right )mod2$; otherwise, do nothing. \\ 
\textbf{Dilation}: For any row $i$ and $j$, $i\neq j$, if $\sum \mathbf{S}_{i}=1$ and $\mathbf{S}_{i}\cap \mathbf{S}_{j}=0$, dilation $di^{i\rightarrow j}\left ( \mathbf{S} \right )$ makes $\mathbf{S}_{i}:= \left (\mathbf{S}_{i}+ \mathbf{S}_{j}  \right )mod2$; else, do nothing. \\ 
\label{def_6}
\end{definition}
In other words, if $\mathbf{S}_{i}$ inherits all ``1''s in $\mathbf{S}_{j}$ and has one more ``1'' than $\mathbf{S}_{j}$, absorption makes $\mathbf{S}_{i}$ get rid of ``1''s that $\mathbf{S}_{i}$ and $\mathbf{S}_{j}$ have in common and only leaves the ``1'' that $\mathbf{S}_{i}$ has but $\mathbf{S}_{j}$ not. Dilation can work only when $\mathbf{S}_{i}$ and $\mathbf{S}_{j}$ shares no common ``1''s and there is only one ``1'' in $\mathbf{S}_{i}$. An example performing dilation and absorption in shown in Figure \ref{fig_10}.
The following tree-matrix equivalence holds (proof in Appendix \ref{sec_APP9}):
\begin{proposition}
For any heterogeneous out-tree $t\in \mathbb{T}_{N}$ and its generated heterogeneous dependency matrix $\mathbf{D}\in \mathbb{D}_{N}$, absorption operation on $t$ is equivalent to absorption operation on $\mathbf{D}$. Similarly, this is true for dilation.
\label{prop_4}
\end{proposition}
\begin{figure}[!t]
 \centering
 \includegraphics[width=\linewidth]{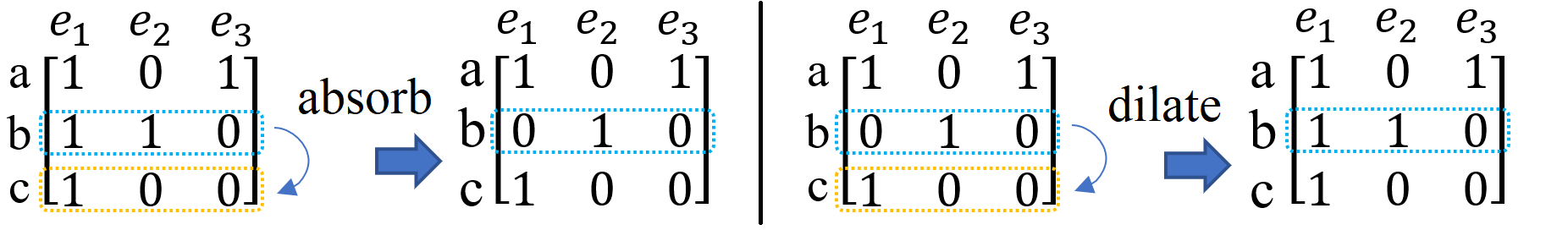}
 \caption{Absorption and dilation on matrix set. Matrices correspond to trees in Figure \ref{fig_5}.}
 \label{fig_10}
\end{figure}
\begin{proposition}
The tree to matrix translation in Lemma \ref{lemma_1} is bijective.
\label{prop_5}
\end{proposition}
From proposition \ref{prop_5} (proved in Appendix \ref{sec_APP10}), we know heterogeneous dependency matrix is a complete representation of heterogeneous out-tree; we 
interchange both names in later sections. In the previous section, we only consider $\mathbf{D}$ generated from a tree, but without discussing what properties $\mathbf{D}$ should have. The following theorem tells us under what condition a binary matrix is a heterogeneous dependency matrix.
\begin{theorem}
For any binary matrix $\mathbf{D}\in \left \{ 0,1 \right \}^{N\times N}$, it has unique corresponding heterogeneous out-tree iff it can be transformed to an permutation matrix $\mathbf{I}$ by a sequence of absorption and dilation.
\label{theo_1}
\end{theorem}
The proof is in Appendix \ref{sec_APP11}. The theorem \ref{theo_1} does not imply that all trees can be reduced to the same permutation matrix. As the heterogeneous dependency matrices are labelled, permutation matrices in Figure \ref{fig_8} are different instances. 

\begin{figure}[!t]
\centering
\begin{subfigure}{.4\linewidth}
    \centering
    \includegraphics[height=.1\textheight]{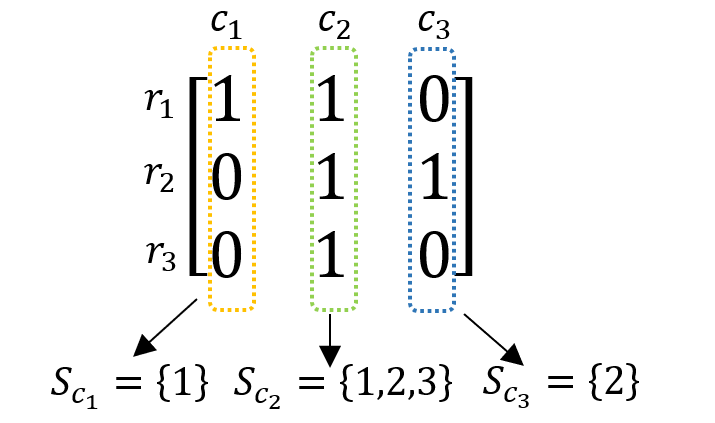}
    \caption{}
    \label{fig_9}
\end{subfigure}
\begin{subfigure}{.4\linewidth}
    \centering
    \includegraphics[height=.1\textheight]{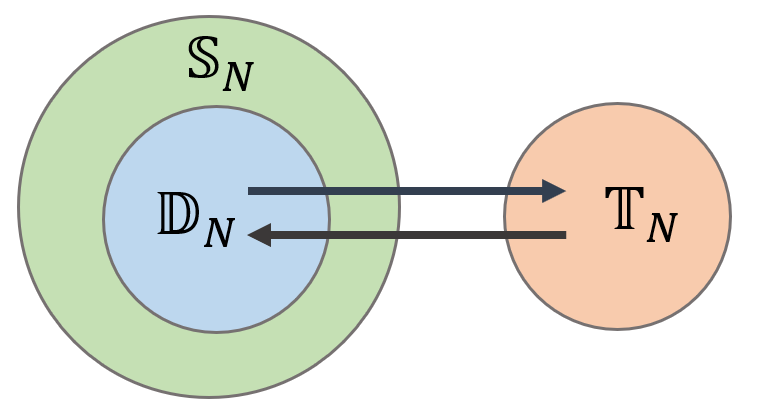}
    \caption{}
    \label{fig_61}
\end{subfigure}
\caption{(a) Column set example. (b) Relation among heterogeneous dependency matrix set $\mathbb{D}_{N}$, heterogeneous out-tree set $\mathbb{T}_{N}$ and binary matrix set $\mathbb{S}_{N}$.}
\end{figure}

We now show more properties of $\mathbf{D}$.
\begin{definition}
For a column $c$, its \emph{Column set} $S_c$ is defined as the set of rows that intersect column $c$. $S_c = \left\{r \in row(\mathbf{D}) \big | \mathbf{D}_{(r,c)} = 1 \right\}$.
\label{def_8}
\end{definition}
Each column has its own column set, which is the collection of the corresponding row labels that the elements in the column is 1, as shown in Figure \ref{fig_9}. It represents all the nodes in the sub-tree rooted in column $c$.
\begin{definition}
Define $J_{S_c}$ the \emph{Unique Element Set} of $S_c$,  if $J_{S_c} = S_c \backslash \bigcup_{c^\prime \in col(\mathbf{D}) : S_{c^\prime} \subsetneq S_{c} }$
\label{def_9}
\end{definition}
$J_{S_c}$ is the set of $r \in row(\mathbf{D})$ such that $r$ is in $S_c$ but not in any $S_{c^\prime}$ which is a proper subset of $S_c$. For example in Figure \ref{fig_9}, $J_{S_c1}=\left \{ 1 \right \}$, $J_{S_c3}=\left \{ 2 \right \}$, because $S_c1$ and $S_c3$ has no proper subset. $S_c1$ and $S_c3$ are the proper subsets of $S_c2$, so $J_{S_c2}$ should only include row labels that only appear in $S_c2$ but not in $S_c1$ or $S_c3$, which is $J_{S_c2}=\left \{ 3 \right \}$.\\
\begin{definition}
Consider the following conditions for $\mathbf{D}$:
\begin{enumerate}
    \item[] 1) There are no all-0 rows in $\mathbf{D}$.
    \item[] 2) There are no all-0 columns in $\mathbf{D}$.
    \item[] 3) There are no duplicate rows in $\mathbf{D}$.
    \item[] 4) There are no duplicate columns in $\mathbf{D}$.
    \item[] 5) $\forall c_{1},c_{2}\in col\left ( \mathbf{D} \right ),S_c1\subseteq S_c2 \  or \ S_c1\supseteq  S_c2 \  or \ S_c1 \cap  S_c2=\varnothing $
    \item[] 6) $\forall c,\left | J_{S_c} \right |=1$
\end{enumerate}
\label{def_10}
\end{definition}
We also define some sets of conditions:
\[ P=\left \{ 1^{\circ} ,2^{\circ},3^{\circ},4^{\circ},5^{\circ}\right \} \qquad P_{0}=\left \{ 1^{\circ} ,2^{\circ},5^{\circ},6^{\circ}\right \} \]
We denote a matrix $\mathbf{D}$ subject to a set of conditions $P$ as $\mathbf{D}:P$

\begin{proposition}
A matrix $\mathbf{D} \in \left\{0,1\right\}^{n \times n}$ satisfies $P_{0}$ $\Leftrightarrow\mathbf{D}$  satisfies $P$.
\label{prop_6}
\end{proposition}
\begin{theorem}
A matrix $\mathbf{D}$ satisfies $P$ is equivalent to matrix $\mathbf{D}$can be translated to a heterogeneous out-tree.
\label{theo_2}
\end{theorem}
Proofs can be found in Appendix \ref{sec_APP12} and \ref{sec_APP14}. Proposition \ref{prop_6} tells us condition sets $P$ and $P_0$ are equivalent, $\mathbf{D}:P\Leftrightarrow\mathbf{D}:P_{0}$.
Theorem \ref{theo_2} then shows that such $\mathbf{D}$ under condition set $P$ (or $P_0$) is equivalent to a heterogeneous out-tree.
If a binary matrix fulfills condition set $P$, it is a heterogeneous dependency matrix, and can be grown to a heterogeneous out-tree.  We propose an algorithm to do the matrix to out-tree translation, as shown in Algorithm \ref{algo_3}. Such translation can be found in pipeline Figure \ref{fig_34} right, where a heterogeneous out-tree is output at the final stage of our system. A demonstration of running the algorithm translating an example matrix to an out-tree is in Appendix \ref{sec_APP13}.
\begin{algorithm}[h]
   \caption{Matrix to Out-tree Translation Algorithm}
   \label{algo_3}
\begin{algorithmic}
   \STATE {\bfseries Input:} row set $\mathbf{r}=\left \{ r_{i} \right \}$ in $\mathbf{D}$, root node, explored set $E=\varnothing $.
   \WHILE{$\mathbf{r}\neq \varnothing $}
     \STATE find $r_{min}$ in $\mathbf{r}$ with minimal depth.
     \STATE pop $r_{min}$ from $\mathbf{r}$
       \IF{depth of $r_{min}$ is 1} 
         \STATE The column label with the "1" entry is $e$. Attach $r_{min}$ to root by edge $e$.
        \ELSE 
        \STATE find one $r'$ in $E$ that $r_{min}-r'=1$. The column label with the "1" entry is $e$. Attach $r_{min}$ to $r'$ by edge $e$.
       \ENDIF
       \STATE push $r_{min}$ in $E$.
   \ENDWHILE
   \STATE {\bfseries Output:} heterogeneous out-tree from root node
\end{algorithmic}
\end{algorithm}

The core idea of this section is shown in Figure \ref{fig_61}. Proposition \ref{prop_5} shows that the bi-directional arrows hold, the out-tree set $\mathbb{T}_{N}$ and matrix set $\mathbb{D}_{N}$ are different representations of the same thing. Theorem \ref{theo_1} and \ref{theo_2} tells us when any binary matrix in $\mathbb{S}_{N}$ is a heterogeneous dependency matrix in $\mathbb{D}_{N}$. They are proven from different perspectives.
\subsection{Partial Observability}\label{sec_PO}
In this section, we will discuss the partial observability problem. In the ideal case, each link will have some sensors mounted on, so that the sensors and links are one-one pair as Figure \ref{fig_3} shows. The heterogeneous dependency matrix is a square matrix of $N\times N$. However, when there are some links with no sensor mounted on, the movement or the dependency of such link is unobservable. The heterogeneous dependency matrix will have fewer rows than columns. If we need to extract a valid heterogeneous out-tree, we need to complete such matrix to square and also ensure it follows condition in theorem \ref{theo_1} or \ref{theo_2}. In this section, we will discuss under what condition the matrix can be completed to a square matrix and how such completion can be performed.

Let $\mathbf{D}^{-}$ be a sub-matrix of $\mathbf{D}:P$ of size $K*N$, where $K<N$. We now consider if such $\mathbf{D}^{-}$ has a corresponding tree. The first problem is whether we can fill $\mathbf{D}^{-}$ to a $\mathbf{D}$ of size $N\times N $, that is to get rid of partial observability. Because multiple different $\mathbf{D}$ can be cut to the same sub-matrix $\mathbf{D}^{-}$, we assume such filling may not be unique. Notice that we are discussing general $\mathbf{D}^{-}$ and $\mathbf{D}$, not necessarily compliant to any constraint set $P$. The following theorem shows when such filling exists and when the filling is unique. We define another set of condition $P^{-}=\left \{ 1^{\circ} ,3^{\circ},5^{\circ} \right \}$.
\begin{theorem}
A $\mathbf{D}^{-}:P^{-}$ can be filled to a $\mathbf{D}:P$. Such filling is unique when $N-K=1$ and there is one $J_{S_c}$ is $\varnothing$.
\label{theo_3}
\end{theorem}
Proof can be found in Appendix \ref{sec_APP16}. Theorem \ref{theo_3} tells us, a $\mathbf{D}^{-}$ can be filled to $\mathbf{D}:P$ if and only if $\mathbf{D}^{-}:P^{-}$. The filling is not unique unless $N-K=1$ and there is one $J_{S_c}$ is $\varnothing$. In that case, there will be a unique corresponding tree and the topology of the robot is fully determined.
\subsection{Matrix Completion Algorithm}
Knowing the conditions that a $\mathbf{D}^{-}$ can be filled to a $\mathbf{D}$ is not enough. We need to design a Matrix Completion Algorithm (Algorithm \ref{algo_2}) to perform such filling. We proposed an algorithm to fill $\mathbf{D}$ from $\mathbf{D}^{-}$ if $\mathbf{D}^{-}$ fulfills condition set $P^{-}$, as is shown in Algorithm \ref{algo_2}. We showed an example running such algorithm to fill $\mathbf{D}^{-}$ in Appendix \ref{sec_APP15}.
\begin{algorithm}[h]
   \caption{Matrix Completion Algorithm}
   \label{algo_2}
\begin{algorithmic}
   \STATE {\bfseries Input:} partial matrix $\mathbf{D}^{-}$ of $K\times N$.
   \STATE Calculate $S_c$ and $J_{S_c}$ for each column $c$ in $\mathbf{D}^{-}$.
   \STATE Randomly pick $K$ different columns with different $J_{S_c}$, and push the rest columns' labels (edge) in set $a=\varnothing $.
   \STATE Push the $N-K$ unobserved rows' labels (nodes) in set $b=\varnothing $
   \WHILE{$a\neq \varnothing $}
     \STATE Pull one edge $c$ from $a$ and one node  $v$ from $b$.
     \STATE Add a new row to $\mathbf{D}^{-}$.
     \FOR{all columns $c'$ in $\mathbf{D}^{-}$} 
       \IF{$S_c\subseteq S_{c'}$} 
         \STATE Set $\mathbf{D}^{-}_{v, c'}=1$
       \ENDIF
     \ENDFOR
   \ENDWHILE
   \STATE {\bfseries Output:} Complete matrix $\mathbf{D}$.
\end{algorithmic}
\end{algorithm}

\subsection{Contradiction}\label{sec_contra}
In this section, we discuss matrices that do not follow Theorem \ref{theo_1} or Theorem \ref{theo_2}. Such violation will not happen if such matrix is directly extracted from a heterogeneous tree. However, if such matrix is learned from some noisy observations or data-driven methods, the extracted matrix can be erroneous. For example, matrix
\begin{equation}
    \begin{bmatrix}
1 & 1 & 0\\ 
0 & 1 & 1\\ 
1 & 0 & 1
\end{bmatrix}
\end{equation}
does not have any corresponding heterogeneous out-tree (certainly, it also does not follow theorem \ref{theo_1} and \ref{theo_2}). We still need to infer the tree structure by finding the most similar matrix $\mathbf{D}\in \mathbb{D}_{N}$ to it which encodes the most possible tree structure. The most straightforward way is to enumerate all possible tree structures spanned by $N$ nodes and edges, compare each one with the erroneous matrix, and choose the one with minimal difference (e.g. Hamming distance). However due to Property \ref{per_5}, the enumerating will lead to an astronomical number when $N$ grows. So an algorithm efficiently recovers the correct matrix is needed.

Inspired by Proposition \ref{prop_3}, we know each tree corresponds to a depth-1 tree, or by Theorem \ref{theo_1}, a permutation matrix. Such permutation matrices (Figure \ref{fig_8}) encode the pairing of nodes and edges. Dilation or absorption on such permutation matrix will change the tree's topology but not such pairing. Thus, $\mathbb{D}_{N}$ can be segmented by such $N!$ (pairing $N$ edges with $N$ nodes) different permutation matrices, each of them can grow to a class of trees that have different topology but the same nodes-edges pairing. Notice the "1"s in the permutation matrix should also appear in the matrix of the tree it grows to. So we can first find the most similar permutation matrix of the erroneous matrix, and from such permutation matrix to grow the most similar tree.

\begin{algorithm}[h]
   \caption{Matrix Correction Algorithm}
   \label{algo_1}
\begin{algorithmic}
   \STATE {\bfseries Input:} Erroneous matrix $\mathbf{D}$ of $N\times N$.
   \STATE Calculate the most similar permutation matrix $\mathbf{I}$ by MILP as Equation \ref{eq_6}.
   \STATE $b=\left \{ \mathbf{I} \right \}$, $g=+\infty $.
   \WHILE{true}
   \STATE $b_{set}=\varnothing$, $d_{set}=\varnothing$
   \FOR{each $\mathbf{M}$ in $b$} 
   \FOR{$i,j\in N,i\neq j$} 
   \STATE $\tilde{\mathbf{M}}=di^{i\rightarrow j}\left ( \mathbf{M} \right )$
   \STATE $d_{\tilde{\mathbf{M}}}=$ Hamming distance between $\tilde{\mathbf{M}}$ and $\mathbf{D}$.
   \STATE $b_{set}=d_{set}\cup \tilde{\mathbf{M}}$, $d_{set}=d_{set}\cup d_{\tilde{\mathbf{M}}}$
   \ENDFOR
   \ENDFOR
   \STATE $\tilde{g}=\textup{min}\left (d_{set} \right )$
   \IF{$\tilde{g}=g$} 
   \STATE {break} 
   \ELSE 
   \STATE{$g:=\tilde{g}$, $b:=b_{set}$ whose ones have $d_{\tilde{\mathbf{M}}}=\tilde{g}$}
   \ENDIF
   \ENDWHILE
   \STATE {\bfseries Output:} Correct matrix candidates $b$.
\end{algorithmic}
\end{algorithm}

Finding the most similar permutation matrix can be solved by enumerating all possible permutation matrices and comparing the overlapping. However, there are still $N!$ possible permutation matrices and will be less efficient when $N$ grows. We turn such problem into a mixed-integer linear programming (MILP) problem. Though MILP is NP-Hard, in practice, we found it is very fast in finding the permutation matrix.

Let $\mathbf{I}$ be the permutation matrix we want to find. We know that the original matrix $\mathbf{D}$ and $\mathbf{I}$ must share the positions where the "1"s are in $\mathbf{I}$. We represent such overlapping  by another matrix $\mathbf{W}_{ij}=\textup{min}\left ( \mathbf{D}_{ij},\mathbf{I}_{ij} \right )$. That means only when both $\mathbf{D}_{ij}$ and $\mathbf{I}_{ij}$ are "1", $\mathbf{W}_{ij}$ becomes "1" (overlapping). The optimization target becomes maximizing the total number of overlapping, which is:
\begin{equation}
    \begin{split}
       & \underset{\mathbf{I}_{ij},\mathbf{W}_{ij}}{\textup{max}}\sum _{i,j}\mathbf{W}_{ij}\\
        s.t. &\quad \sum _{i}\mathbf{I}_{ij}=1,\quad 
        \sum _{j}\mathbf{I}_{ij}=1,\quad 
        \mathbf{I}_{ij}\in \left \{ 0,1 \right \}, \\
        &\mathbf{W}_{ij}\leq \mathbf{I}_{ij}, \quad 
        \mathbf{W}_{ij}\leq \mathbf{D}_{ij}
    \end{split}
    \label{eq_6}
\end{equation}
$\mathbf{W}_{ij}$ is the auxiliary variable in this case. The first two constraints ensure $\mathbf{I}$ is a permutation matrix, the third constraint ensures $\mathbf{I}$ to be binary. The last two constraints come from $\mathbf{W}_{ij}=\textup{min}\left ( \mathbf{D}_{ij},\mathbf{I}_{ij} \right )$. In such way we turn such "min" into two conditions in the optimization. Such optimization can be efficiently solved by MILP algorithms. The algorithm returns $\mathbf{I}$ and $\mathbf{W}$ simultaneously and we only use $\mathbf{I}$ in the following steps.

With the most similar permutation matrix, the next step is to generate a tree from such matrix by a sequence of dilation. From Property \ref{per_5}, we know there are $\left ( N+1 \right )^{N-1}$ trees can be generated, which is still a large number. We leverage the Trellis algorithm that only performs one-step dilation and retain the matrices with the smallest Hamming distance. We cut the branches with high Hamming distances that they are unlikely to generate the tree we need. The algorithm is summarized in algorithm \ref{algo_1}. We show an example running such algorithm in Appendix \ref{sec_APP17}.

\subsection{Application in Robotics}

Here we briefly explain how such theory can be used in real robot applications. In section \ref{sec_HDM} and \ref{sec_HOT}, we have shown how an open-chain robot can be reduced to a heterogeneous out-tree, and such out-tree is a valid representation of robot topology. However, if we cannot directly observe such topology, we have to infer such topology by some observations from other sources. In this paper, we assume the "other sources" are exteroception and proprioception data, which are measured by full-body IMUs and joint encoders. From such measurements, we can extract a binary heterogeneous dependency matrix. Theorem \ref{theo_1} and \ref{theo_2} tells us such matrix is a complete representation of heterogeneous out-tree we want to know, which represents the robot topology. Then we discussed the partially observable case, which is that for some links, there is no sensor mounted on. Theorem \ref{theo_3} tells us when such partial observed data can be reconstructed to a heterogeneous out-tree and when such reconstruction is unique. The matrix correction algorithm ensures when the heterogeneous dependency matrix is erroneous, how we can still correct it and extract the tree structure.

Such mechanism could have significant potential in robotics. First, the mechanism can monitor the body structure of the robot, when the robot body is damaged, our algorithm will identify that the body topology has changed and give a damage warning. Also, for some robots with unknown body structures or some robots whose FKs are hard to model (like soft robots), our work can find the most similar rigid body topology that describes the unknown body structure.

\section{EXPERIMENTS} \label{sec_EXP}
\subsection{Simulation}
\begin{figure}[!t]
 \centering
 \includegraphics[width=\linewidth]{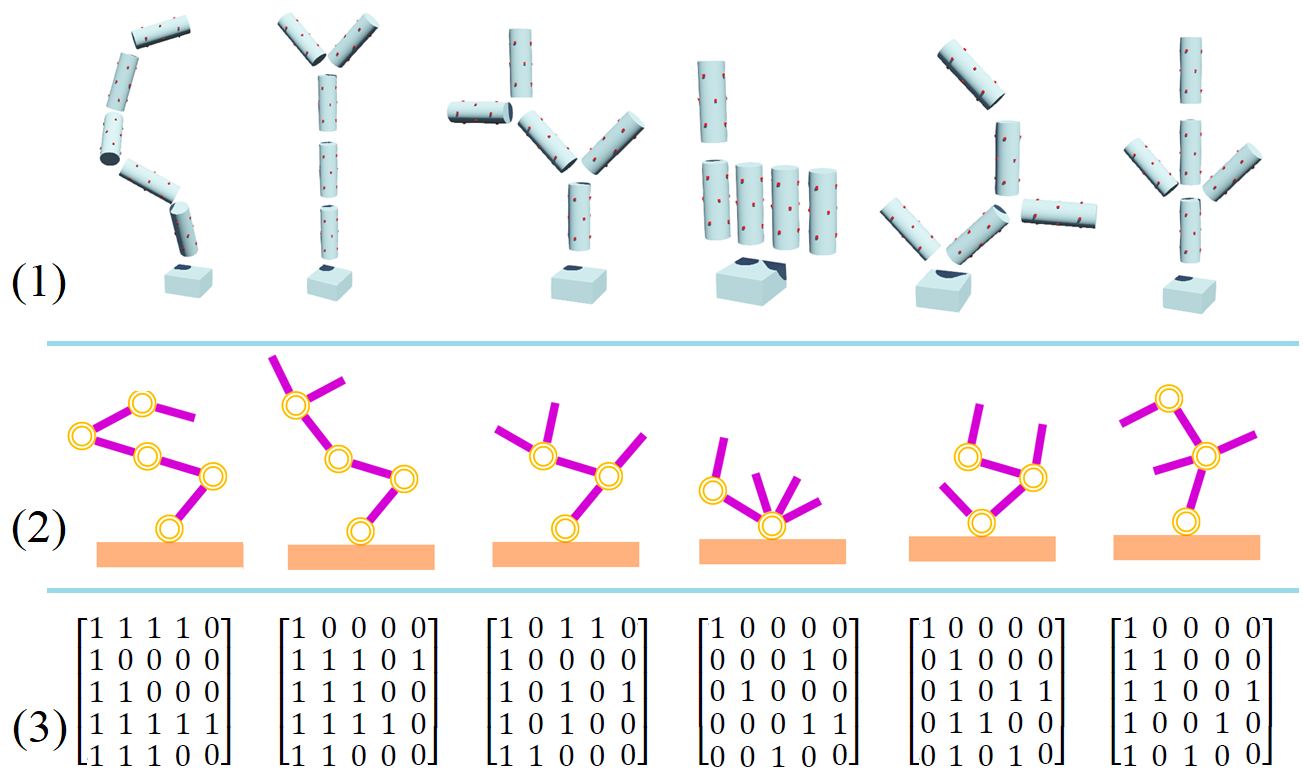}
 \caption{(1) Different robot topology in experiment. Red dots on robot body are IMUs. (2) the recognized robot body structures; (3) the extracted heterogeneous dependency matrices.}
 \label{fig_54}
\end{figure}

\noindent
For the experiments, we designed 6 open-chain robots in Webots simulator with different body structures (Figure \ref{fig_54} (1)). Each robot has 5 links except the base link and with 5 joints. 12 IMUs are evenly distributed on the surface of each link. The z-axis of IMU is pointing outwards the link along the surface normal while x and y axes are chosen arbitrarily. To record data, we perform random motion on each joint, which means every 0.2 seconds, we cast a uniformly distributed torque on each joint and keep the torque constant in such period. The sampling frequency is 100 Hz.
The result of running our algorithm is shown in Figure \ref{fig_54} (2-3). Our algorithm correctly identify the ground-truth robot body structures as (1) by extracting heterogeneous dependency matrices as shown in (3).
\subsection{Baselines Comparison}
\begin{figure}[!t]
 \centering
 \includegraphics[width=\linewidth]{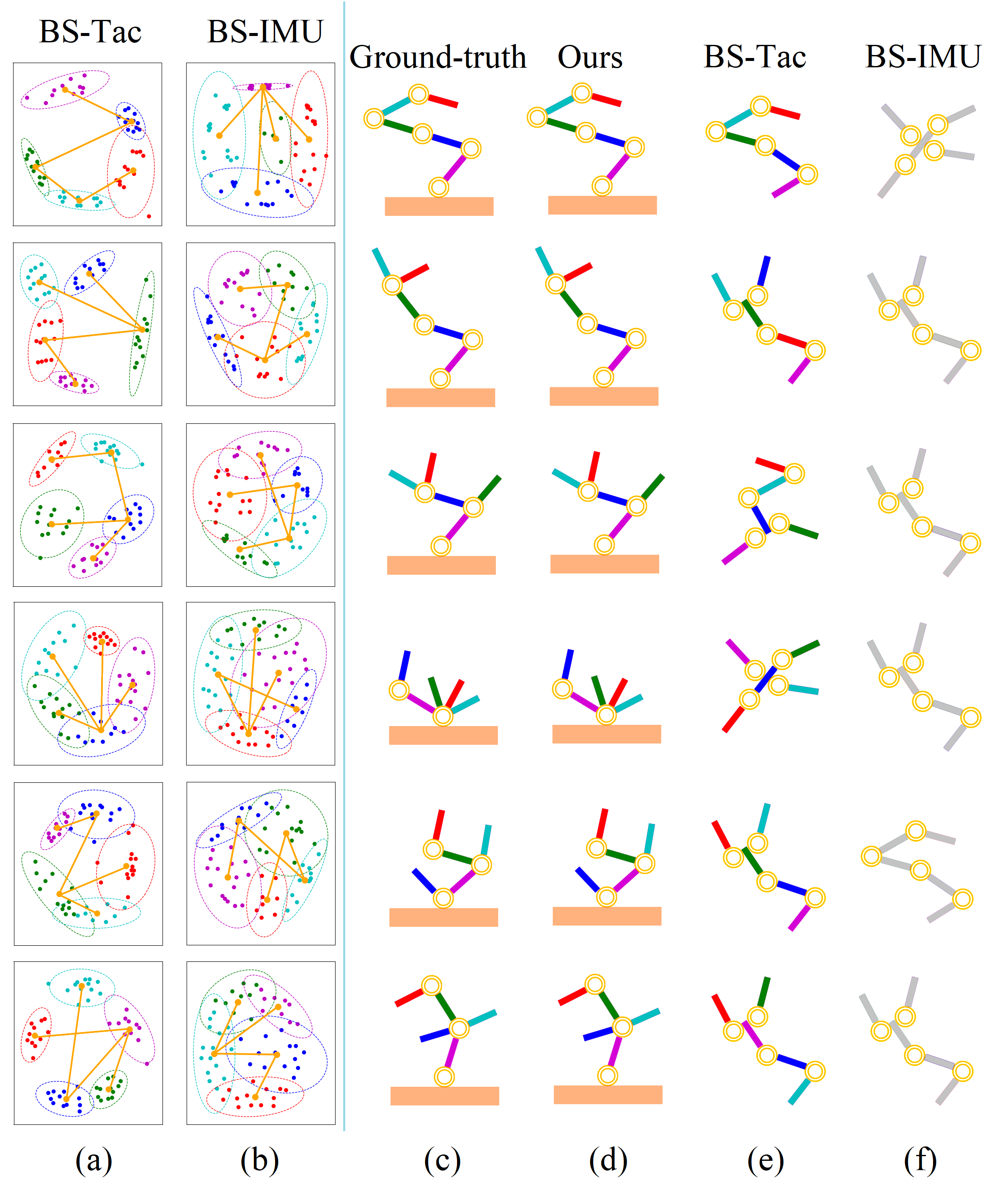}
 \caption{Body schema recognition: (a) clustering and body schema recognition results of \cite{mimura2017bayesian} using tactile signal; (b) results of \cite{mimura2017bayesian} using IMU signal. Yellow nodes represent recognized links and yellow edges are recognized joints; (c) ground-truth topology; (d) results of our solution; (e) results of \cite{mimura2017bayesian} using tactile signal; (f) results of \cite{mimura2017bayesian} using IMU signal. Colors show the link correspondence.}
 \label{fig_69}
\end{figure}

We compared work from \cite{mimura2017bayesian} with the six robots in the last sub-section, the result is shown in Figure \ref{fig_69}. Results of their method using their original tactile data is marked as BS-Tac and using our IMU data is marked as BS-IMU. The dimension of Multidimensional scaling (MDS) in their method is set as $d=10$, which means the original signal is mapped to 10 dimension space. We use mutual information to estimate the information metric $D$ in their work. 

The sensor clustering and topology recognition are shown in Figure \ref{fig_69} (a, b). The yellow nodes represent recognized links and yellow edges are recognized joints. Due to the original 10-D mapping is then reduced to 2-D for plotting, each cluster may not connect to its nearest cluster in 2D space. Comparing BS-Tac and BS-IMU, we can find that their algorithm  clustering in BS-Tac but the clustering is not clear in BS-IMU, which means their work has weaker distinguishing ability with local information such as IMU signal. 

The right side of Figure \ref{fig_69} shows the recognized topology of our solution and theirs. Different from our solution, their work finds the undirected graphs as Figure \ref{fig_69} (e,f) while ours finds the out-tree structure Figure \ref{fig_69} (d). The color of links are used to mark different labels, and they are consistent in all figures. In the last column, all links are marked grey, because the algorithm cannot cluster the sensors and find their correspondence correctly. Clearly, our work reconstructs the topology precisely, but the baselines only captures the topology correctly with only a small fraction of body parts. Also, some links are wrongly connected, as robot 2 red and purple link. As IMU data is a highly non-linear mapping from joint values, it is arguable to use a information metric extracted from data to directly represent the interconnection between clustering, as their work uses to find the latent joints. This might be the reason that their work shows clustering performance with tactile data extracted from global information (see more details in \cite{mimura2017bayesian}), but in BS-IMU, clustering is less recognizable Figure \ref{fig_69} (b). 

\begin{figure*}[!t]
 \centering
 \includegraphics[width=.9\linewidth]{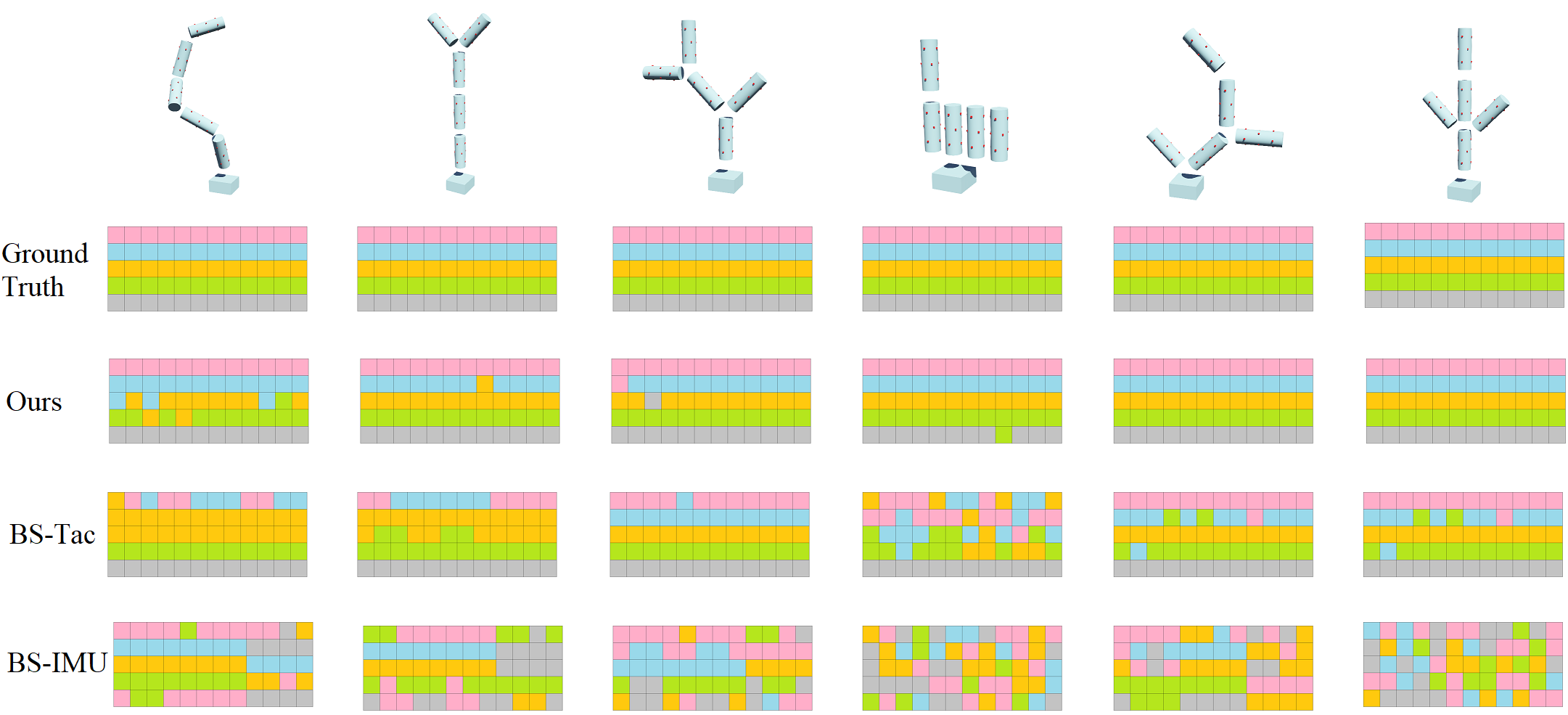}
 \caption{Clustering results of DPGMM components.}
 \label{fig_82}
\end{figure*}

The clustering result is more clear in Figure \ref{fig_82}. Both our solution and \cite{mimura2017bayesian} employed DPGMM, so we can compare the clustering results with the ground-truth clustering, as the correspondence between sensors and links is known (which sensor is connected to which link). From Figure \ref{fig_82}, we see our solution preserves the ground-truth clustering well. However the clustering in BS-IMU is almost random. It means that their method can hardly find the correspondence between sensors and links with only IMU data. It is the reason that in Figure \ref{fig_69}, the links are marked with grey.

\subsection{Real Robot Test}
\begin{figure}[!t]
\hfill
\begin{subfigure}{.45\linewidth}
    \centering
    \includegraphics[height=.1\textheight]{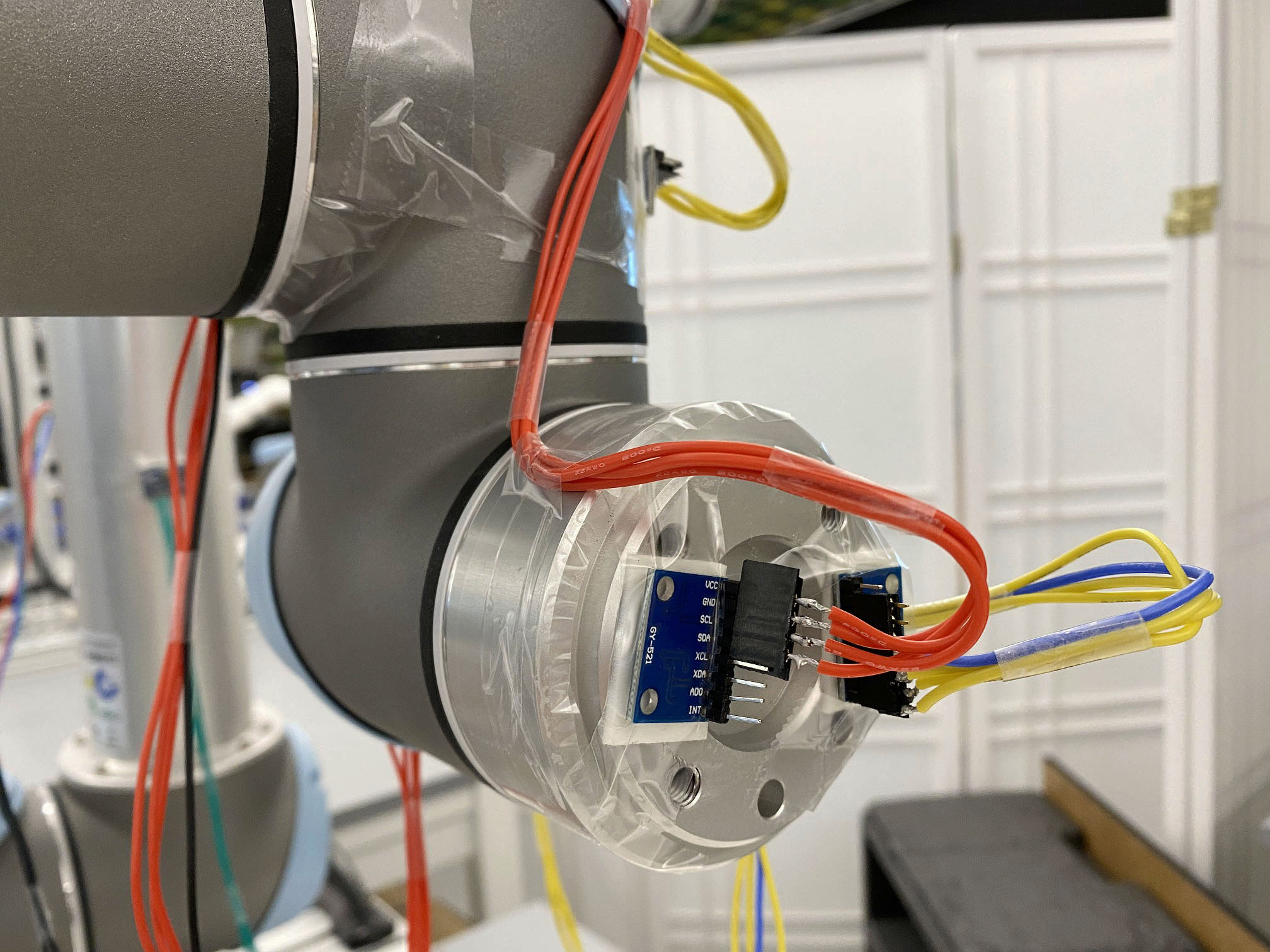}
    \caption{}
    \label{fig_86}
\end{subfigure}
\hfill
\begin{subfigure}{.45\linewidth}
    \centering
    \includegraphics[height=.1\textheight]{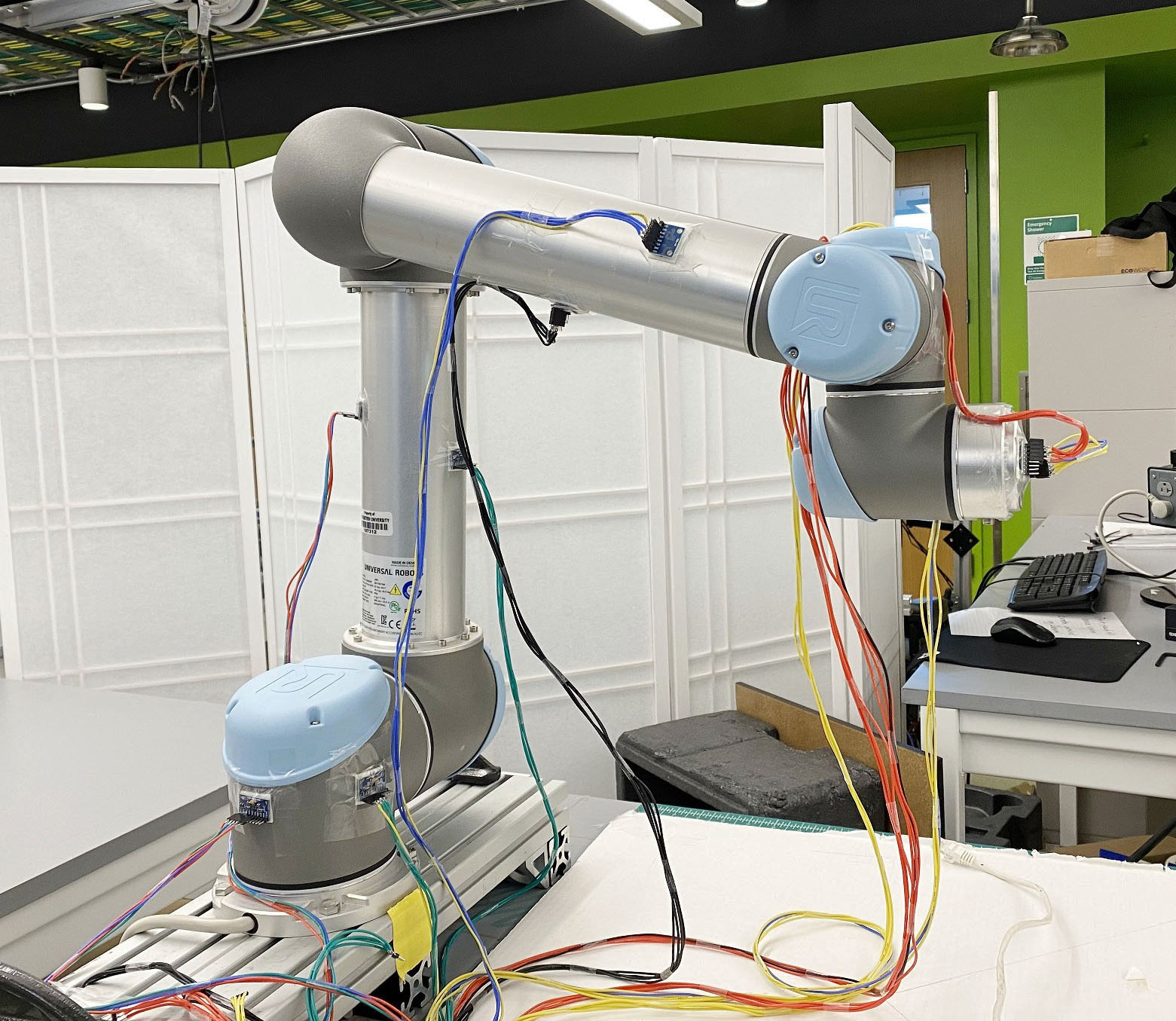}
    \caption{}
    \label{fig_87}
\end{subfigure}
\hfill
\begin{subfigure}{\linewidth}
    \centering
    \includegraphics[height=.11\textheight]{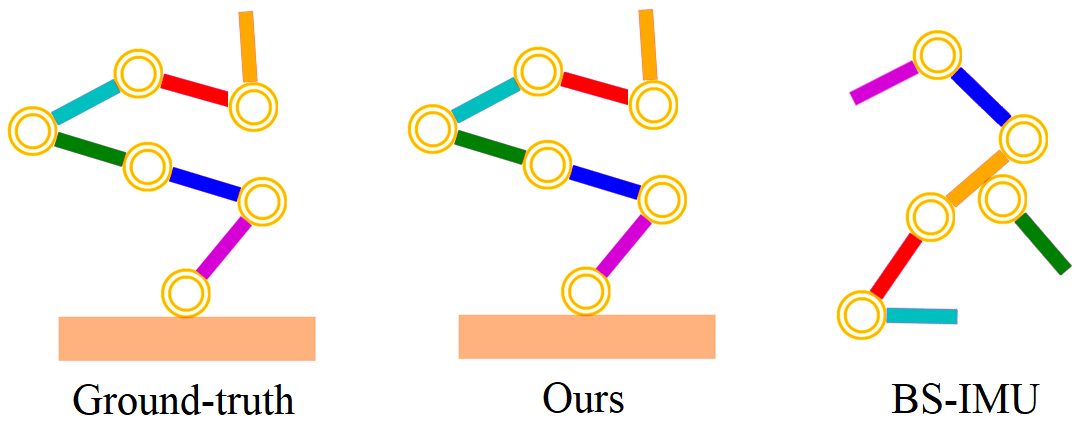}
    \caption{}
    \label{fig_92}
\end{subfigure}
\hfill
\caption{(a) MPU6050 mounting. (b) Robot with full-body IMUs. (c) Ground-truth and recognized robot body structure by our algorithm and baseline \cite{mimura2017bayesian}.}
\end{figure}

We tested our algorithm on a UR5e robot arm. We mount two MPU6050 sensors on each link (Figure \ref{fig_86}-\ref{fig_87}). The robot is programmed to track sinusoidal joint velocities as Equation \ref{eq_19}.
\begin{equation}
    \dot{q}\left ( t \right ) =A_{i}\cdot \textup{sin}\left ( 2\pi k_{i}\cdot t+y _{i} \right )
    \label{eq_19}
\end{equation}
The parameters $A_{i}$, $k_{i}$ and $y _{i}$ chosen are shown in Table \ref{tab_1}, $i$ is the index for the joints in Table \ref{tab_1}. As the real robot is affected by gravity that introduces additional acceleration term, Equation \ref{eq_20} shall be modified to cancel out the gravity effect, which becomes Equation \ref{eq_21}.
\begin{equation}
    \phi^{*} =\underset{\phi}{\textup{argmin}}\left [\left | \alpha  _{B}-\mathbf{R}^{T}\mathbf{g}-\mathbf{R}^{T}\mathbf{\ddot{b}} \right |-\textup{tr}\left (\mathbf{R}_{1}^{T}\mathbf{R}_{2}  \right )  \right ]
    \label{eq_21}
\end{equation}
$\mathbf{g}$ is the gravity vector that is usually set as $\mathbf{g}=\left [ 0,0,-9.8 \right ]^{T}$. The recognized body structure is shown in Figure \ref{fig_92}. From that we see our algorithm successfully recognized the UR5e robot body topology but baseline \cite{mimura2017bayesian} failed to.
\begin{table}[!t]
    \centering
    \begin{tabular}{ |c|c|c|c| } 
\hline
joint name & $A_{i}$ & $k_{i}$ & $y _{i}$ \\
\hline
shoulder\_pan\_joint & 0.2 & 0.1 &  0.1\\
\hline
shoulder\_lift\_joint & 0.2 & 0.13 &  0.2\\
\hline
elbow\_joint & 0.2 & 0.15 &  0.3\\
\hline
wrist\_1\_joint & 0.2 & 0.17 &  0.4\\
\hline
wrist\_2\_joint & 0.2 & 0.19 & 0.5\\
\hline
wrist\_3\_joint & 0.2 & 0.21 &  0.6\\
\hline
\end{tabular}
    \caption{Real robot joint velocity trajectory parameters.}
    \label{tab_1}
\end{table}

\subsection{Neural Network Structure Comparison}
\begin{figure*}[!t]
\hfill
\begin{subfigure}{.25\linewidth}
    \centering
    \includegraphics[height=.25\textheight]{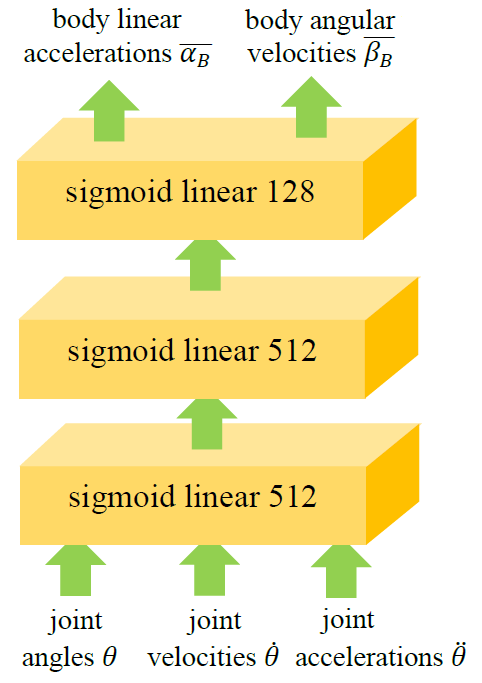}
    \caption{}
    \label{fig_88}
\end{subfigure}
\hfill
\begin{subfigure}{.7\linewidth}
    \centering
    \includegraphics[height=.22\textheight]{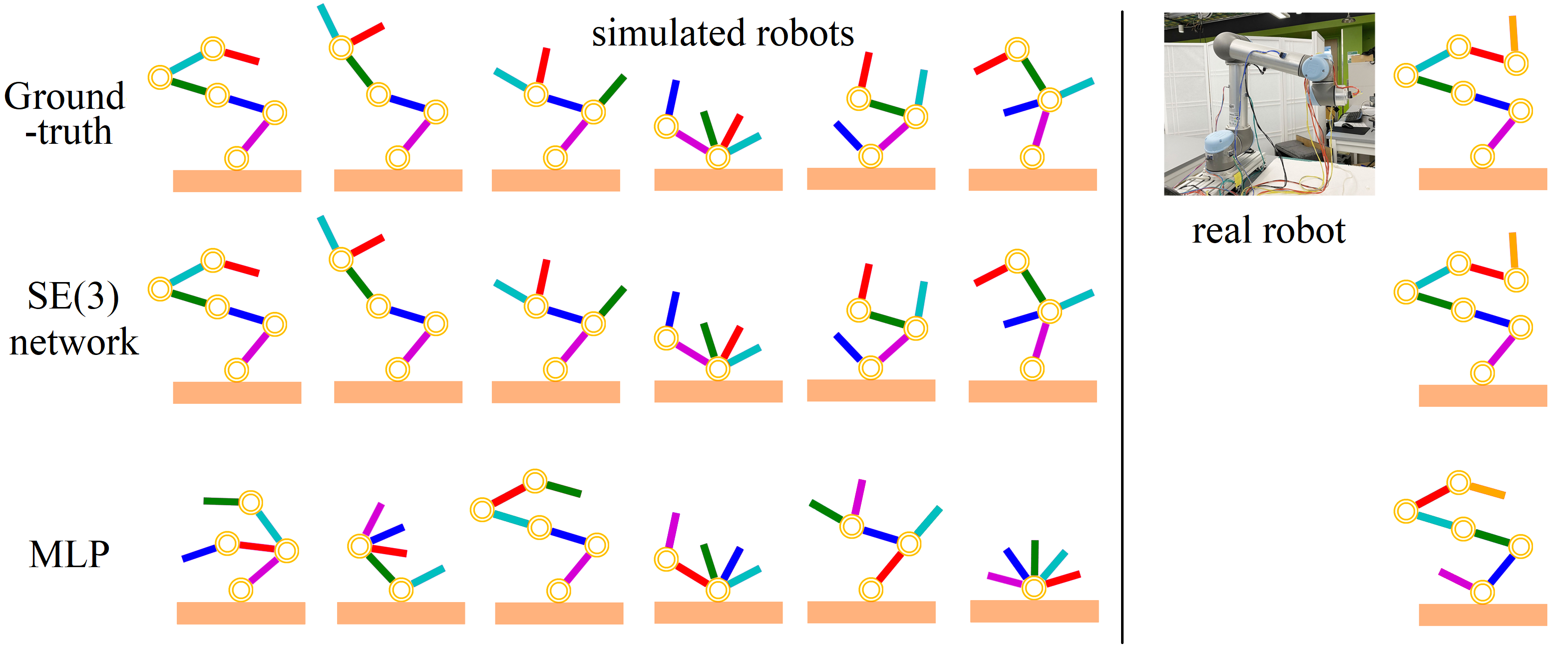}
    \caption{}
    \label{fig_89}
\end{subfigure}
\hfill
\caption{(a) Neural network structure of multi-layer perceptrons (MLP) regressor. (b) Heterogeneous Out-tree extraction results of simulated robots and real robot. (top row) ground-truth tree structure; (middle row) tree structure extracted using the proposed SE(3) network; (bottom row) tree structure extracted using MLP network.}
\end{figure*}

To show the advantage of our neural network structure proposed in Figure \ref{fig_35}, we compared with a multi-layer perceptrons (MLP) regressor. For each sensor on the robot, there is an independent MLP regressor attached to it. The network structure is shown in Figure \ref{fig_88}. The input of neural network is joint angles $\theta $, velocities $\dot{\theta }$ and accelerations $\ddot{\theta }$, the outputs are predicted body linear accelerations $\overline{\alpha _{B}}\in \mathbb{R}^{3}$ and body angular velocities $\overline{\beta  _{B}}\in \mathbb{R}^{3}$. The neural network weights are trained by minimizing the following loss function 
\begin{equation}
    \phi ^{*}=\underset{\phi }{\textup{argmin}}\left ( \left \| \alpha _{B}-\overline{\alpha _{B}} \right \| +\left \| \beta _{B}-\overline{\beta _{B}} \right \|\right )
\end{equation}
where $\alpha _{B}$, $\beta  _{B}$ are measured body linear accelerations and body angular velocities from IMUs. The Jacobian of such network is $\frac{\partial \left ( \overline{\alpha _{B}}, \overline{\beta  _{B}}\right )}{\partial \left ( \theta, \dot{\theta }, \ddot{\theta } \right )}\in \mathbb{R}^{6\times 3n}$. We only use the joint angles part $\frac{\partial \left ( \overline{\alpha _{B}}, \overline{\beta  _{B}}\right )}{\partial \theta}\in \mathbb{R}^{6\times n}$. The sliced Jacobians are then processed by Equation \ref{eq_22}-\ref{eq_23} to extract the corresponding Heterogeneous Dependency Matrix of the robot. The results are shown in Figure \ref{fig_89}. From the result, we see MLP is not capable of learning the correct tree structures.

\section{CONCLUSION}
\noindent
In this work, we proposed an algorithm for an open-chain robot with full-body exteroception and proprioception sensors to recognize its body structure. The exteroception signal is measured by IMUs and the proprioception signal is measured by joint encoders. We assigned each on-body sensor a neural network and proposed a training method to learn the global poses of the sensors. The Jacobian of such neural networks encodes dependency information which can be formed as a heterogeneous dependency matrix. We can also extract a heterogeneous out-tree structure from each robot that represents its body structure. We showed such matrix and out-tree are equivalent representations of the robot body topology. As such matrix comprises a subset of all binary square matrices, we proved two theorems that give the condition when a binary square matrix is a heterogeneous dependency matrix and discussed the remedy to fix such matrix when it is contaminated by partial observability and data noise. The simulation result showed that our algorithm successfully identified the body structures of 6 different robots with different skeletons and was validated on an UR5e real platform. The potential application of our algorithm is to identify the forward kinematics of robots whose forward kinematics is not accessible or hard to model. It is a big step for a robot to recognize itself instead of only interacting with the environment. The future work includes extending such algorithm to locomotion robot scenarios, shortening the the time of training to make it more fast-adaptive.

\appendix
\begin{appendices}
\section{Detailed Derivation of Exteroception Estimation} \label{sec_APP1}
\noindent
Assuming we have a second-order differentiable function approximator $f_{\phi }\left ( \theta  \right )$
\begin{equation}
    f_{\phi }\left ( \theta  \right )=\mathbf{T}_{\phi }\left ( \theta  \right )=\begin{bmatrix}
\mathbf{R}\left ( \theta  \right ) & \mathbf{b}\left ( \theta  \right ) \\ 
\mathbf{0} & 1
\end{bmatrix}
\end{equation}
which is parameterized by $\phi$. The function approximator is used to map the input joint angle vector $\theta$ to a homogeneous transformation $\mathbf{T}_{\phi }\left ( \theta  \right )$ in SE(3) that characterizes the transformation that sensor frame $B$ described in global frame $G$. Reader should keep in mind that for each sensor on the robot body, there will be a unique $f_{\phi }\left ( \theta  \right )$ to represent the sensor, we only discuss one of them, and the rest are used as the same way. Due to lacking of the knowledge of the robot body structure, the ground-truth homogeneous transformation is unknown, that is why we use such an approximator. We denote the first and second-order derivative of $f_{\phi }\left ( \theta  \right )$ as $\frac{\partial f_{\phi }\left ( \theta  \right )}{\partial \theta}=\mathbf{J}_{\phi}\left ( \theta  \right )\in \mathbb{R}^{4\times 4\times N}$ and $\frac{\partial^{2} f_{\phi }\left ( \theta  \right )}{\partial \theta^{2}}=\mathbf{H}_{\phi}\left ( \theta  \right )\in \mathbb{R}^{4\times 4\times N\times N}$, where $N$ is the number of joints. We can also measure the proprioception as the angles, angular velocities and angular accelerations for all the joint angles (though we do not know their hierarchy) as $\theta$, $\dot{\theta }$, $\ddot{\theta }$. The exteroception data are the vectors $\left [ \alpha _{B},\beta   _{B}\right ]^{T}$ of linear acceleration and angular velocity (by roll-pitch-yaw) described in body frame, where $\alpha _{B}\in \mathbb{R}^{3}$, $\beta   _{B}\in \mathbb{R}^{3}$. We will ignore the symbol of dependency of $\theta$ for all variables to make the result neat. The first and second-order time derivative of $f_{\phi }\left ( \theta  \right )$ are shown in Equations \ref{eq_15} and \ref{eq_16}.
\begin{equation}
    \frac{\partial f_{\phi } }{\partial t}=\frac{\partial f_{\phi }}{\partial \theta }\frac{\partial \theta }{\partial t}=\mathbf{J}_{\phi }\dot{\theta}=\begin{bmatrix}
\mathbf{\dot{R}} & \mathbf{\dot{b}}\\ 
\mathbf{0} & 1
\end{bmatrix}
\label{eq_15}
\end{equation}
\begin{equation}
    \begin{split}
        \frac{\partial ^{2}f_{\phi }}{\partial t^{2}}&=\frac{\partial }{\partial t}\left ( \frac{\partial f_{\phi } }{\partial t} \right )\\
        &=\frac{\partial }{\partial t}\left ( \frac{\partial f_{\phi } }{\partial \theta }\frac{\partial \theta }{\partial t} \right )\\
        &=\frac{\partial }{\partial t}\left ( \mathbf{J}_{\phi }\dot{\theta} \right )\\
        &=\left (\frac{\partial \mathbf{J}_{\phi }}{\partial t}  \right )\dot{\theta} +\mathbf{J}_{\phi }\ddot{\theta}\\
        &=\left (\frac{\partial \mathbf{J}_{\phi }}{\partial \theta } \frac{\partial \theta }{\partial t} \right )\dot{\theta} +\mathbf{J}_{\phi }\ddot{\theta}
    \end{split}
\end{equation}
\begin{equation}
    =\begin{bmatrix}
\mathbf{\ddot{R}} & \mathbf{\ddot{b}}\\ 
\mathbf{0} & 1
\end{bmatrix}=\mathbf{H}_{\phi }\dot{\theta}\dot{\theta}+\mathbf{J}_{\phi }\ddot{\theta}
\label{eq_16}
\end{equation}
The two equations show we can find $\mathbf{\dot{R}}$, $\mathbf{\dot{b}}$, $\mathbf{\ddot{R}}$ and $\mathbf{\ddot{b}}$ from exteroception and proprioception data. The next step is to find the relation between $\mathbf{R}$, $\mathbf{b}$, $\mathbf{\dot{R}}$, $\mathbf{\dot{b}}$, $\mathbf{\ddot{R}}$,  $\mathbf{\ddot{b}}$ and  $\left [ \alpha  _{B}, \beta  _{B}\right ]^{T}$.\\

For any point $r$ in space, we have
\begin{equation}
    r_{G}=\mathbf{T}_{\phi }\cdot r_{B}
\end{equation}
$r_{G}$ and $r_{B}$ are coordinates of $r$ in global and sensor frame respectively. Take the second-order time derivative of such relation, we have
\begin{equation}
    a_{G} = \ddot{r}_{G}=\mathbf{\ddot{T}}_{\phi }\cdot r_{B}=\begin{bmatrix}
\mathbf{\ddot{R}} & \mathbf{\ddot{b}}\\ 
\mathbf{0} & 0
\end{bmatrix}\cdot r_{B}
\end{equation}
so the acceleration of point $r$ in body frame is
\begin{equation}
    \begin{split}
        a_{B} &= \mathbf{T}_{\phi }^{-1}a_{G}=\mathbf{T}_{\phi }^{-1}\cdot \mathbf{\ddot{T}}_{\phi }\cdot r_{B}\\
&=\begin{bmatrix}
\mathbf{R}^{T} & -\mathbf{R}^{T}\mathbf{b}\\ 
\mathbf{0} & 1
\end{bmatrix}\cdot \begin{bmatrix}
\mathbf{\ddot{R}} & \mathbf{\ddot{b}}\\ 
\mathbf{0} & 0
\end{bmatrix}\cdot r_{B}\\
&=\begin{bmatrix}
\mathbf{R}^{T}\mathbf{\ddot{R}} & \mathbf{R}^{T}\mathbf{\ddot{b}}\\ 
\mathbf{0} & 0
\end{bmatrix}\cdot r_{B}
    \end{split}
\end{equation}
Define operation
\begin{equation}
    \begin{bmatrix}
w_{1}\\ 
w_{2}\\ 
w_{3}
\end{bmatrix}^{\wedge }=\begin{bmatrix}
0 & -w_{3} & w_{2}\\ 
w_{3} & 0 & -w_{1}\\ 
-w_{2} & w_{1} & 0
\end{bmatrix}
\end{equation}
and
\begin{equation}
    \begin{bmatrix}
0 & -w_{3} & w_{2}\\ 
w_{3} & 0 & -w_{1}\\ 
-w_{2} & w_{1} & 0
\end{bmatrix}^{\vee }=\begin{bmatrix}
w_{1}\\ 
w_{2}\\ 
w_{3}
\end{bmatrix}
\end{equation}
We conclude that the linear part $\alpha  _{B}$ equals $\mathbf{R}^{T}\mathbf{\ddot{b}}$ because they are both in $\mathbb{R}^{3}$. To find the expression of $\beta_{B}$, from $r_{G}=\mathbf{T}_{\phi }\cdot r_{B}$, we have 
\begin{equation}
    \dot{r}_{G}=\mathbf{\dot{T}}_{\phi }\cdot r_{B}=\mathbf{\dot{T}}_{\phi }\cdot\mathbf{T}^{-1}_{\phi }\cdot r_{G}=\begin{bmatrix}
\mathbf{\dot{R}}\mathbf{R}^{T} &\mathbf{\dot{b}}-\mathbf{\dot{R}}\mathbf{R}^{T}\mathbf{b} \\ 
\mathbf{0} & 0
\end{bmatrix}\cdot r_{G}
\end{equation}
Define global angular velocity matrix as $\omega  _{G}^{\wedge }=\mathbf{\dot{R}}\mathbf{R}^{T}$. Apparently $\omega  _{G}^{\wedge }\in \mathfrak{so}\left ( 3 \right )$ and the corresponding global angular velocity is $\omega  _{G}\in \mathbb{R}^{3}$. We know for any vector $v$,
\begin{equation}
    \left ( \mathbf{R}\cdot v \right )^{\wedge }=\mathbf{R}\cdot v^{\wedge }\cdot \mathbf{R}^{T} 
    \label{eq_17}
\end{equation}
We know the relation $\omega  _{G}=\mathbf{R}\omega  _{B}$ as $\omega  _{G}$ and $\omega  _{B}$ are the same vector expressed in different frames. By Equation \ref{eq_17}, we have 
\begin{equation}
    \omega  _{B}^{\wedge }=\mathbf{R}^{-1}\omega  _{G}^{\wedge }\mathbf{R}=\mathbf{R}^{-1}\mathbf{\dot{R}}\mathbf{R}^{T}\mathbf{R}=\mathbf{R}^{T}\mathbf{\dot{R}}
\end{equation}
However, the corresponding vector $\omega  _{B}$ is not $\beta _{B}$ because $\omega  _{B}$ is the rotation velocity vector described by axis-angle system and $\beta _{B}$ is the rotation velocity vector described by roll-pitch-yaw system. To align the two representations, we know the two velocity vectors should cause the same rotation effect in a short period of time $T_{s}$. For $\omega  _{B}$, it can be decomposed to a time derivative of angle $\dot{\vartheta }\in \mathbb{R}$ and a rotation axis vector $\vec{u}\in \mathbb{R}^{3}$, with $\omega  _{B}=\dot{\vartheta }\vec{u}$ and $\left |\vec{u}   \right |=1$. The rotated angle in $T_{s}$ can be approximated as $\dot{\vartheta }T_{s}$. By Rodrigues' rotation formula, the corresponding rotation matrix in $T_{s}$ is 
\begin{equation}
    \mathbf{R}_{1}=\mathbf{I}+\textup{sin}\left ( \dot{\vartheta }T_{s} \right )\omega  _{B}^{\wedge }+\left [ 1-\textup{cos}\left (  \dot{\vartheta }T_{s}\right ) \right ]\omega  _{B}^{\wedge }\cdot \omega  _{B}^{\wedge }
\end{equation}
For $\beta _{B}$, it represents the roll, pitch, yaw velocities as $\beta _{B} = \left [  \beta _{B1} ,\beta _{B2} ,\beta _{B3} \right ]^{T}$. The rotation angle in $T_{s}$ can be approximated by $\left [  \beta _{B1} T_{s} ,\beta _{B2}T_{s} ,\beta _{B3}T_{s} \right ]^{T}$, so the rotation matrix is calculated from standard roll-pitch-yaw matrices as
\begin{equation}
    \begin{split}
        \mathbf{R}_{2}&=\mathbf{R}_{z}\left ( \beta _{B3}T_{s} \right )\mathbf{R}_{y}\left ( \beta _{B2}T_{s} \right )\mathbf{R}_{x}\left ( \beta _{B1}T_{s} \right )\\
&=\begin{bmatrix}
a_{00} & a_{01} & a_{02}\\ 
a_{10} & a_{11} &a_{12} \\ 
a_{20} & a_{21} & a_{22}
\end{bmatrix}
    \end{split}
\end{equation}
with
\begin{equation}
\footnotesize
\begin{split}
    a_{00}&=\textup{cos}\left ( \beta _{B3}T_{s} \right )\textup{cos}\left ( \beta _{B2}T_{s} \right )\\
    a_{01}&=\textup{cos}\left ( \beta _{B3}T_{s} \right )\textup{sin}\left ( \beta _{B2}T_{s} \right )\textup{sin}\left ( \beta _{B1}T_{s} \right )-\textup{sin}\left ( \beta _{B3}T_{s} \right )\textup{cos}\left (  \beta _{B1}T_{s} \right )\\
    a_{02}&=\textup{cos}\left ( \beta _{B3}T_{s} \right )\textup{sin}\left ( \beta _{B2}T_{s} \right )\textup{cos}\left ( \beta _{B1}T_{s} \right )+\textup{sin}\left ( \beta _{B3}T_{s} \right )\textup{sin}\left (  \beta _{B1}T_{s} \right )\\
    a_{10}&=\textup{sin}\left ( \beta _{B3}T_{s} \right )\textup{cos}\left ( \beta _{B2}T_{s} \right )\\
    a_{11}&=\textup{sin}\left ( \beta _{B3}T_{s} \right )\textup{sin}\left ( \beta _{B2}T_{s} \right )\textup{sin}\left ( \beta _{B1}T_{s} \right )+\textup{cos}\left ( \beta _{B3}T_{s} \right )\textup{cos}\left (  \beta _{B1}T_{s} \right )\\
    a_{12}&=\textup{sin}\left ( \beta _{B3}T_{s} \right )\textup{sin}\left ( \beta _{B2}T_{s} \right )\textup{cos}\left ( \beta _{B1}T_{s} \right )-\textup{cos}\left ( \beta _{B3}T_{s} \right )\textup{sin}\left (  \beta _{B1}T_{s} \right )\\
    a_{20}&=-\textup{sin}\left ( \beta _{B2}T_{s} \right )\\
    a_{21}&=\textup{cos}\left ( \beta _{B2}T_{s} \right )\textup{sin}\left ( \beta _{B1}T_{s} \right )\\
    a_{22}&=\textup{cos}\left ( \beta _{B2}T_{s} \right )\textup{cos}\left ( \beta _{B1}T_{s} \right )
\end{split}
\end{equation}
\section{Proof of Proposition \ref{prop_2}} \label{sec_APP2}
\noindent
\textbf{Proposition \ref{prop_2}} \textit{Transform Invariant Jacobian is invariant to any transformation $\mathbf{Y}$.}\\

\textit{Proof:} Similarly, we represent sensor frame $\mathbf{T}$ as
\begin{equation}
\mathbf{T}=\begin{bmatrix}
\mathbf{n}_{x} & \mathbf{n}_{y} & \mathbf{n}_{z} & \mathbf{b}\\ 
0 & 0 & 0 & 1
\end{bmatrix}
\end{equation}
$\mathbf{n}_{x}$, $\mathbf{n}_{y}$, $\mathbf{n}_{z}$ are unit vectors of each axis, and $\mathbf{b}$ is the global position. We have an arbitrary transformation $\mathbf{Y}$ as
\begin{equation}
    \mathbf{Y}=\begin{bmatrix}
\mathbf{R}' & \mathbf{b}'\\ 
\mathbf{0} & 1
\end{bmatrix}
\end{equation}
The transformed new frame is 
\begin{equation}
    \mathbf{T}' = \mathbf{Y}\cdot \mathbf{T}=\begin{bmatrix}
\mathbf{n}'_{x}\left ( \theta  \right ) & \mathbf{n}'_{y}\left ( \theta  \right ) & \mathbf{n}'_{z}\left ( \theta  \right ) & \mathbf{b}''\left ( \theta  \right )\\ 
0 & 0 & 0 & 1
\end{bmatrix}
\end{equation}
with $\mathbf{n}'_{x}\left ( \theta  \right )=\mathbf{R}'\cdot \mathbf{n}_{x}\left ( \theta  \right )$, \\$\mathbf{n}'_{y}\left ( \theta  \right )=\mathbf{R}'\cdot \mathbf{n}_{y}\left ( \theta  \right )$, \\$\mathbf{n}'_{z}\left ( \theta  \right )=\mathbf{R}'\cdot \mathbf{n}_{z}\left ( \theta  \right )$,\\$\mathbf{b}''\left ( \theta  \right )=\mathbf{R}'\cdot \mathbf{b}\left ( \theta  \right )+\mathbf{b}'$\\
~\\
The new Jacobians are:\\
$\frac{\partial \mathbf{n}'_{x}\left ( \theta  \right )}{\partial \theta }=\mathbf{R}'\frac{\partial \mathbf{n}_{x}\left ( \theta  \right )}{\partial \theta }$\\
$\frac{\partial \mathbf{n}'_{y}\left ( \theta  \right )}{\partial \theta }=\mathbf{R}'\frac{\partial \mathbf{n}_{y}\left ( \theta  \right )}{\partial \theta }$\\
$\frac{\partial \mathbf{n}'_{z}\left ( \theta  \right )}{\partial \theta }=\mathbf{R}'\frac{\partial \mathbf{n}_{z}\left ( \theta  \right )}{\partial \theta }$\\
$\frac{\partial \mathbf{r}'\left ( \theta  \right )}{\partial \theta }=\mathbf{R}'\frac{\partial \mathbf{r}'\left ( \theta  \right )}{\partial \theta }$\\
~\\
Because $\mathbf{R}'$ is unitary matrix, we have $\left |\frac{\partial \mathbf{n}'_{x}\left ( \theta  \right )}{\partial \theta }  \right |=\left |\frac{\partial \mathbf{n}_{x}\left ( \theta  \right )}{\partial \theta }  \right |$, \\
$\left |\frac{\partial \mathbf{n}'_{y}\left ( \theta  \right )}{\partial \theta }  \right |=\left |\frac{\partial \mathbf{n}_{y}\left ( \theta  \right )}{\partial \theta }  \right |$, \\
$\left |\frac{\partial \mathbf{n}'_{z}\left ( \theta  \right )}{\partial \theta }  \right |=\left |\frac{\partial \mathbf{n}_{z}\left ( \theta  \right )}{\partial \theta }  \right |$, \\
$\left |\frac{\partial \mathbf{b}''\left ( \theta  \right )}{\partial \theta }  \right |=\left |\frac{\partial \mathbf{b}\left ( \theta  \right )}{\partial \theta }  \right |$\\
\textit{end of proof.}\\
\section{Proof of Property \ref{per_3}} \label{sec_APP3}
\noindent
\textbf{Property \ref{per_3}} \textit{Rows and columns of $\mathbf{D}$ are non-duplicate.}\\

\textit{Proof:} When we construct matrix $\mathbf{D}$, we eliminate all duplicate rows, so there are no duplicate rows. Because each element $\mathbf{D}_{ij}$ records if the path from node $i$ to root passes edge $j$ or not, any column $j$ of $\mathbf{D}$ represents the collection of indices of all descendants of edge $j$. Because one edge is only connecting to one child node directly, such child node must be the root node of the sub-tree constructed by this collection (though we don't exactly know which node it is connecting, it should be one in the collection). If another edge $k$ has the same column, similarly, the child node connects to edge $k$ must be the root node of the sub-tree constructed by that collection. We know two edges $j$ and $k$ cannot connect to the same child node, which means duplicate column instroduces two root nodes to the same sub-tree, which is not possible.\\
\textit{end of proof.}
\section{Proof of Property \ref{per_5}} \label{sec_APP4}
\noindent
\textbf{Property \ref{per_5}} \textit{There are $\left ( N+1 \right )^{N-1}\cdot N!$ different tree structures with $N$ nodes.}\\

\textit{Proof:} From Cayley's Formula, the number of trees without edge labels is $\left ( N+1 \right )^{N-1}$. Then we take node-edge pairing, there are $N!$ different ways of pairing $N$ nodes and $N$ edges.\\
\textit{end of proof.}
\section{Proof of Lemma \ref{lemma_1}} \label{sec_APP5}
\noindent
\textbf{Lemma \ref{lemma_1}} \textit{From each heterogeneous out-tree, one can extract a determined heterogeneous dependency matrix.}\\

\textit{Proof:} Because it is an out-tree, for each node $i$ in the tree, one can find a unique route from $i$ to the root. As all rows of the matrix are labelled, each node $i$ corresponds to a particular row in the matrix. All edges $j$ that the route passes will set the $j$-th element in row $i$ to one while keeping the unvisited edges zero. In such way, one can construct a determined heterogeneous dependency matrix.\\ 
\textit{end of proof.}
\section{Proof of Lemma \ref{lemma_2}} \label{sec_APP6}
\noindent
\textbf{Lemma \ref{lemma_2}} \textit{$ab^{j\rightarrow i}\left ( t \right )$ and $di^{i\rightarrow j}\left ( t \right )$ on heterogeneous out-tree set are a pair of inverse operation.}\\

\textit{Proof:} Absorption moves a leaf node and its direct parent edge to the root while dilation moves a leaf node and its direct parent edge back to the original position without changing the topology of other parts of the tree.\\
\textit{end of proof.}
\section{Proof of Proposition \ref{prop_3}} \label{sec_APP7}
\noindent
\textbf{Proposition \ref{prop_3}} \textit{Any tree $t$ can be transformed to a depth-1 tree by a series of absorption and from the depth-1 tree, reconstruct the original tree $t$.}\\
\begin{figure}[!t]
 \centering
 \includegraphics[width=\linewidth]{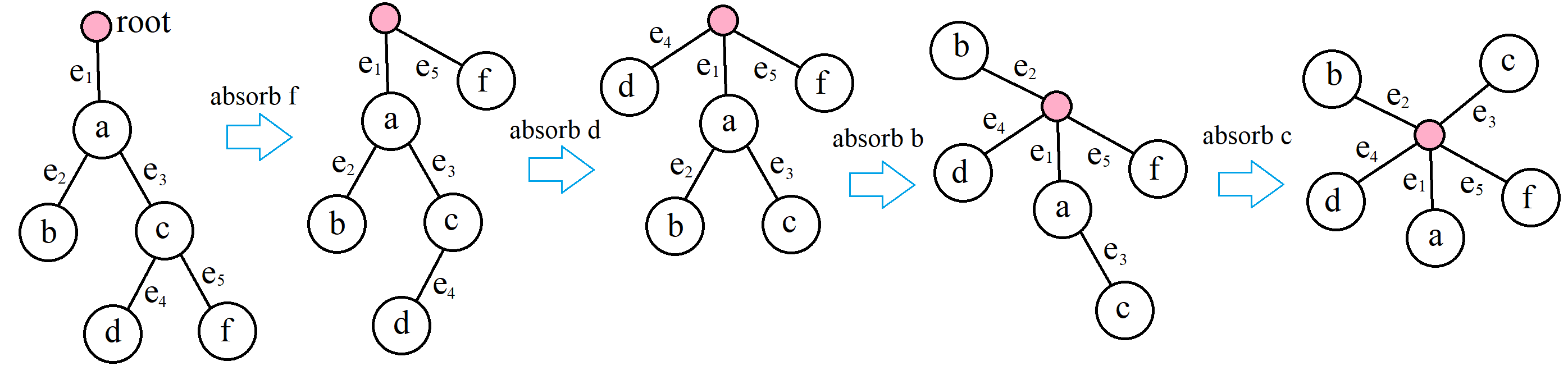}
 \caption{Demonstration of using absorption operation to transform a tree to a depth-1 tree}
 \label{fig_55}
\end{figure}

\textit{Proof:} A depth-1 tree is a star shape graph that each node is directly connecting to the root. We know that for any heterogeneous out-tree, the absorption operation is to move a node with its parent edge to the root. We can design an algorithm that first moves all leaf nodes to the root. Then the old leaf nodes are all depth-1, and some new nodes become leaf nodes. We repeat such operation of absorbing leaf nodes until all nodes are depth-1 (Figure \ref{fig_55}). Due to Lemma \ref{lemma_2},\\
~\\
Lemma \ref{lemma_2}: \textit{$ab^{j\rightarrow i}\left ( t \right )$ and $di^{i\rightarrow j}\left ( t \right )$ on heterogeneous out-tree set are a pair of inverse operation.}\\
~\\
we can find the reversed sequence of dilation operation to reconstruct such tree.\\
\textit{end of proof.}
\section{Proof of corollary \ref{cor_1}} \label{sec_APP8}
\noindent
\textbf{Corollary \ref{cor_1}} \textit{Tree set $\mathbb{T}_{N}$ is closed under dilation and absorption.}\\

\textit{Proof:} From the definition of dilation and absorption, moving a deeper leaf node to the root or moving a root-connected leaf to a deeper position will still result in a tree.\\
\textit{end of proof.}
\section{Proof of Proposition \ref{prop_4}} \label{sec_APP9}
\noindent
\textbf{Proposition \ref{prop_4}} \textit{For any heterogeneous out-tree $t\in \mathbb{T}_{N}$ and its generated heterogeneous dependency matrix $\mathbf{D}\in \mathbb{D}_{N}$, absorption operation on $t$ is equivalent to absorption operation on $\mathbf{D}$. So as for dilation.}\\

\textit{Proof:} Because $\mathbf{D}\in \mathbb{D}_{N}\subseteq \mathbb{S}_{N}$, operations defined on $\mathbb{S}_{N}$ also apply to $\mathbf{D}$. We are aware of a fact that if node $i$ is a child of node $j$, row $\mathbf{D}_{i}$ will have one more "1" than row $\mathbf{D}_{j}$ and all other entries the same. $\mathbf{S}_{i}- \mathbf{S}_{j}$ in absorption operation makes $\mathbf{D}_{j}$ get rid of all the dependencies introduced by $\mathbf{D}_{j}$, and only leave the dependency of $\mathbf{D}_{i}$'s parent edge. This operation makes the new $\mathbf{D}_{i}$ has only one "1", which indicates it directly connects to the root. The operation is equivalent to move node $i$ from child of $j$ to the root along with its edge.

For dilation operation, we ensure there is only one "1" in $\mathbf{D}_{i}$ so that $\mathbf{D}_{i}$ is originally connected to the root. $\mathbf{S}_{i}+ \mathbf{S}_{j}$ in dilation operation makes $\mathbf{D}_{i}$ inherits all dependencies of $\mathbf{D}_{j}$. Because we originally ensure $\mathbf{S}_{i}\cap \mathbf{S}_{j}=0$, the new $\mathbf{D}_{i}$ will have one more "1" than $\mathbf{D}_{j}$, which means $depth\left (\mathbf{D}_{i}  \right )=depth\left (\mathbf{D}_{j}  \right )+1$, $\mathbf{D}_{i}$ becomes the direct child of $\mathbf{D}_{j}$. The operation is equivalent to move node $i$ from the root to the child of $j$ along with its edge. The equivalence of absorption and dilation on tree set and matrix set are shown in Figure \ref{fig_5} and \ref{fig_10}.\\
\begin{figure}[!t]
 \centering
 \includegraphics[width=.8\linewidth]{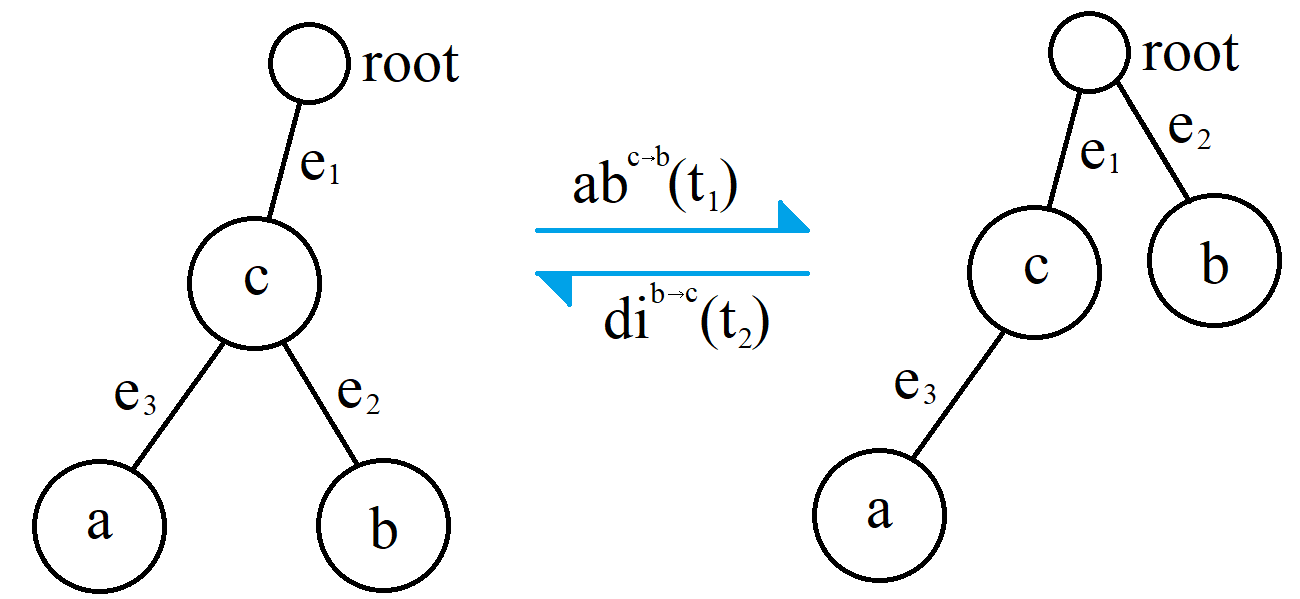}
 \caption{Demonstration of dilation $di^{i\rightarrow j}\left ( t \right )$ and absorption $ab^{j\rightarrow i}\left ( t \right )$.}
\end{figure}
\begin{figure}[!t]
 \centering
 \includegraphics[width=\linewidth]{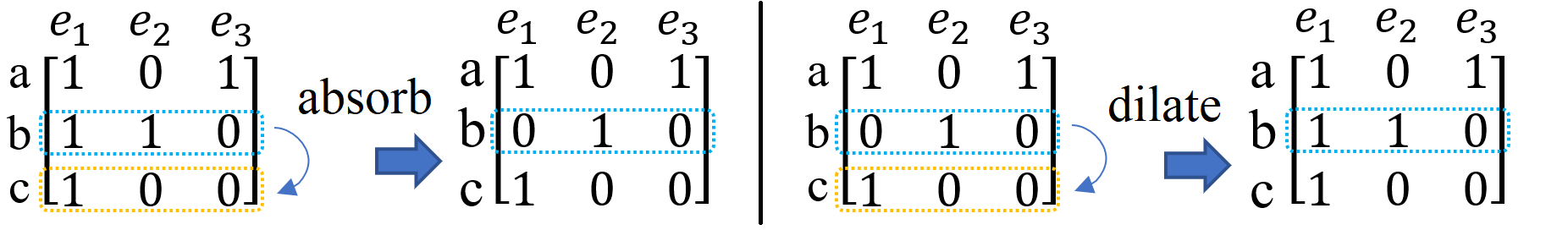}
 \caption{Absorption and dilation on matrix set. Matrices correspond to trees in Figure \ref{fig_5}.}
 \label{fig_10}
\end{figure}
Thus, we can see absorption and dilation defined in $\mathbb{D}_{N}$ and $\mathbb{S}_{N}$ are equivalent.\\
\textit{end of proof.}
\section{Proof of Proposition \ref{prop_5}} \label{sec_APP10}
\noindent
\textbf{Proposition \ref{prop_5}} \textit{The tree to matrix translation in Lemma \ref{lemma_1} is bijective.}\\

\textit{Proof:}
\begin{figure}[!t]
 \centering
 \includegraphics[width=.8\linewidth]{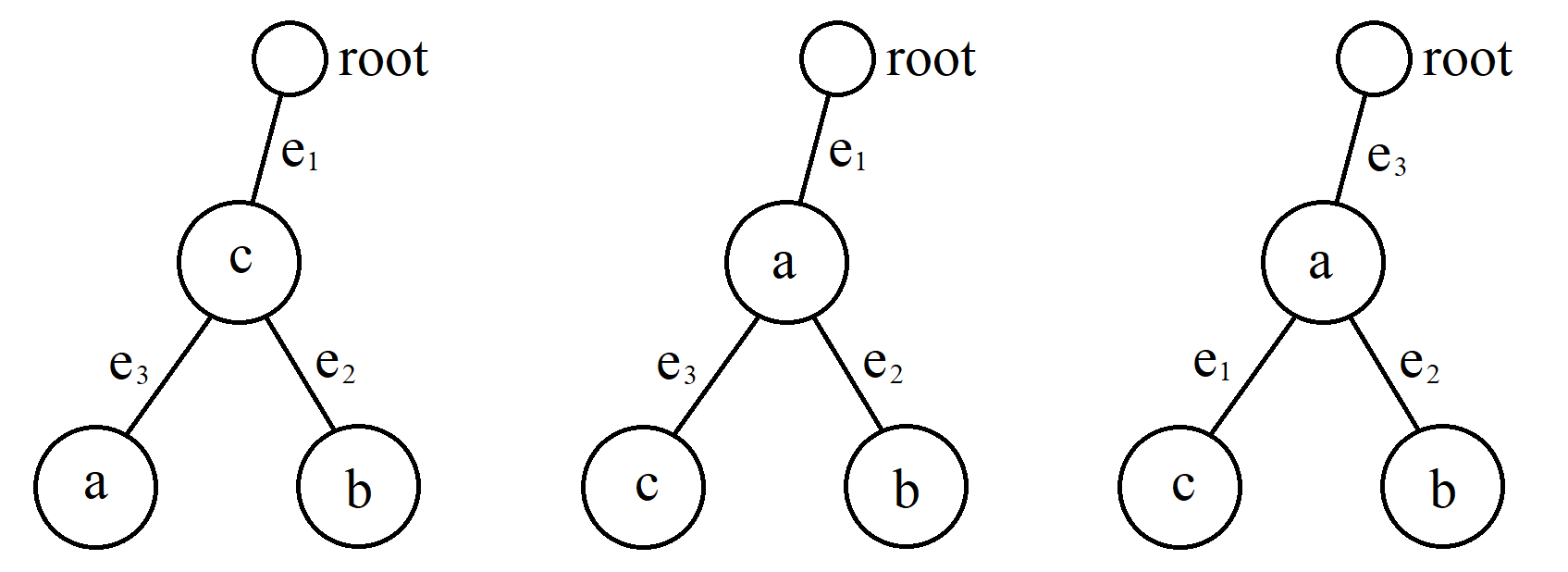}
 \caption{Examples of heterogeneous out-tree.}
\end{figure}
Injective:\\
From Lemma \ref{lemma_1},\\
~\\
Lemma \ref{lemma_1}: \textit{From each heterogeneous out-tree, one can extract a determined heterogeneous dependency matrix.}\\
~\\
we know for each $t\in \mathbb{T}_{N}$, there is $\mathbf{D}\in \mathbb{D}_{N}$. What we need to show is for $t_{1},t_{2}\in \mathbb{T}_{N}$ and $t_{1}\neq t_{2}$, their counterparts $\mathbf{D}_{1},\mathbf{D}_{2}\in \mathbb{D}_{N}$ are different, which is $\mathbf{D}_{1}\neq \mathbf{D}_{2}$. To show such distinction, we first segment $\mathbb{D}_{N}$ by semi-equivalent defined in Definition \ref{def_7}.\\
~\\
Definition \ref{def_7}: \textit{Two heterogeneous out-trees are semi-equivalent if any node with the same label in both trees has the same parent node.}\\
~\\
For example, tree 1 in Figure \ref{fig_4} is in one subset and tree 2, 3 in are in another subset. In each subset, there is a set of trees share the same node-only topology as traditional out-tree. If the two trees $t_{1}$ and $t_{2}$ are different out-tree, at some where in the tree should we find the first pair of parent $a$ and its child $b$ (by search algorithm as BFS) that only found in $t_{1}$ but in $t_{2}$.
\begin{itemize}
  \item in $t_{1}$: $a\to b$
  \item in $t_{2}$: $a\not\to b$
\end{itemize}
We can separate this into two conditions:\\

1) $b$ is a descendant of $a$ (not child) in $t_{2}$ but $b$ is a child of $a$ in $t_{1}$: in $t_{1}$ case, $depth\left ( b \right )=depth\left ( b \right )+1$. In $t_{2}$ case, $b$ is attached to the sub-tree rooted on $a$, $depth\left ( b \right )> depth\left ( b \right )+1$. By property \ref{per_4}, \\
~\\
Property \ref{per_4}: \textit{$\sum _{j}\mathbf{D}_{ij}$ is the depth of node $i$ in the tree.}\\
~\\
there must be $\sum _{j}\mathbf{D_{2}}_{b,j}>\sum _{j}\mathbf{D_{1}}_{b,j}$. Thus $\mathbf{D}_{1}\neq \mathbf{D}_{2}$.\\

2) $b$ is not a descendant of $a$ in $t_{2}$ but $b$ is a child of $a$ in $t_{1}$. in $t_{1}$ case, $b$ inherits dependency of $a$, so $\mathbf{D_{1}}_{b,a}=1$. In $t_{2}$ case, because $a$, $b$ is the first error pair found, $b$ cannot be an ascendant of $a$. So, $b$ must be on a other sub-tree which does not include $a$, which means $b$ has no dependency on $a$, $\mathbf{\mathbf{D_{2}}}_{b,a}=0$. $\mathbf{D}_{1}\neq \mathbf{D}_{2}$

Next, we discuss trees in each subset considering its dependency on edges. This problem is that given an out-tree without edge labels, how to fit different edge label to each edge. Apparently there are $N!$ possible assignments considering all permutations of edge labels as shown in Figure \ref{fig_8}. From property \ref{per_3},\\
~\\
Property \ref{per_3}: \textit{Rows and columns of $\mathbf{D}$ are non-duplicate.}\\
~\\
we know that no columns are duplicate, so each of the assignment will result in a unique matrix $\mathbf{D}$.

To summarize, subset in $\mathbb{T}_{N}$ separated by semi-equivalence will result in non-duplicate $\mathbf{D}$ while in each subset, all the elements have non-duplicate $\mathbf{D}$. This implies for each tree, the extraction will map it to a unique matrix $\mathbf{D}\in \mathbb{D}_{N}$.\\
~\\
Surjective:\\
From the definition of $\mathbf{D}$, we know there is must be a tree that generate such matrix, so each matrix must have a preimage in $\mathbb{T}_{N}$.\\
\begin{figure}[!t]
 \centering
 \includegraphics[width=.5\linewidth]{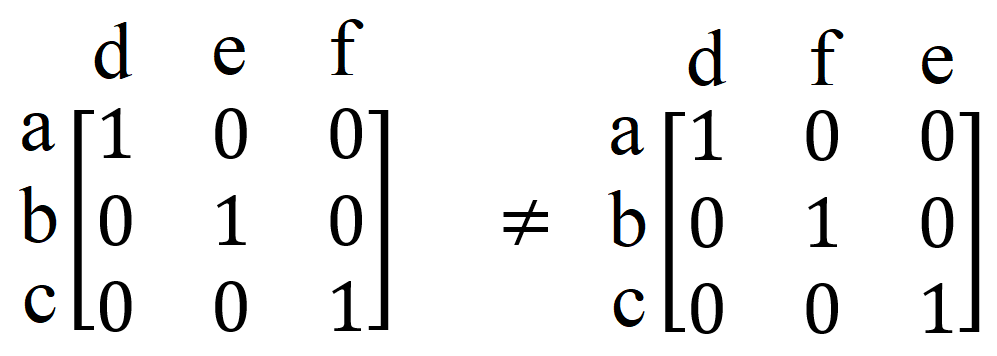}
 \caption{Different permutation matrices}
\end{figure}
\textit{end of proof.}
\section{proof of theorem \ref{theo_1}} \label{sec_APP11}
\noindent
We first prove a lemma.\\

\begin{lemma}
A heterogeneous dependency matrix as a permutation matrix can be extracted from a depth-1 tree.
\label{lemma_3}
\end{lemma}

\textit{Proof:} From Lemma \ref{lemma_1}, \\
~\\
Lemma \ref{lemma_1}: \textit{From each heterogeneous out-tree, one can extract a determined heterogeneous dependency matrix.}\\
~\\
we can extract a determined heterogeneous dependency matrix from depth-1 tree. From property \ref{per_4}, \\
~\\
Property \ref{per_4}: \textit{$\sum _{j}\mathbf{D}_{ij}$ is the depth of node $i$ in the tree.}\\
~\\
we know each row has only one element as 1 and others are all zero. From property \ref{per_3}, \\
~\\
Property \ref{per_3}: \textit{Rows and columns of $\mathbf{D}$ are non-duplicate.}\\
~\\
we know the element "1"s in different rows are separated into different columns. Thus, by property \ref{per_1}, \\
~\\
Property \ref{per_1}: \textit{Rows and columns of $\mathbf{D}$ are interchangeable.}\\
~\\
we can rearrange the matrix to a permutation matrix.\\
\textit{end of proof.}\\
~\\
\textbf{Theorem \ref{theo_1}} \textit{For any binary matrix $\mathbf{D}\in \left \{ 0,1 \right \}^{N\times N}$, it has unique corresponding heterogeneous out-tree iff it can be transformed to an permutation matrix $\mathbf{I}$ by a sequence of adsorption and dilation.}\\

\textit{Proof:} $t\Rightarrow \mathbf{I}$:\\
Combining proposition \ref{prop_3}, lemma \ref{lemma_3} and proposition \ref{prop_4}, \\
~\\
Proposition \ref{prop_3}: \textit{Any tree $t$ can be transformed to a depth-1 tree by a series of absorption and from the depth-1 tree, reconstruct the original tree $t$.}\\
~\\
Lemma \ref{lemma_3}: \textit{A heterogeneous dependency matrix as a permutation matrix can be extracted from a depth-1 tree.}\\
~\\
Proposition \ref{prop_4}: \textit{For any heterogeneous out-tree $t\in \mathbb{T}_{N}$ and its generated heterogeneous dependency matrix $\mathbf{D}\in \mathbb{D}_{N}$, absorption operation on $t$ is equivalent to absorption operation on $\mathbf{D}$. So as for dilation.}\\
~\\
we know if $\mathbf{D}$ is generated by a tree, its corresponding tree must be able to be reduced to a depth-1 tree by absorption. In such case, each node matches with one edge, such that for each row or column, there is only one element is 1. Property \ref{per_1}\\
~\\
Property \ref{per_1}: \textit{Rows and columns of $\mathbf{D}$ are interchangeable.}\\
~\\
shows such matrix can be rearranged to an permutation matrix by switching rows and columns with their labels.\\

$\mathbf{I}\Rightarrow t$:\\
Assuming there is a matrix $\mathbf{D}\notin \mathbb{D}_{N}$ but can be reduced to a permutation matrix by a sequence $s$ of the two operations. Such operation must has a corresponding depth-1 tree in $\mathbb{T}_{N}$. We know the inverse operation of switching rows is switching rows (the same pair), the inverse operation of switching columns is switching columns. Dilation and absorption are inverse operation for each other. So by proposition \ref{prop_4}, in domain $\mathbb{T}_{N}$, there should be a reversed sequence $\bar{s}$ that grows a depth-1 tree. Corollary \ref{cor_1}\\
~\\
Corollary \ref{cor_1}: \textit{Tree set $\mathbb{T}_{N}$ is closed under dilation and absorption.}\\
~\\
shows such grown tree should also be a heterogeneous out-tree. By lemma \ref{lemma_1}, \\
~\\
Lemma \ref{lemma_1}: \textit{From each heterogeneous out-tree, one can extract a determined heterogeneous dependency matrix.}\\
~\\
such tree is equivalent to a matrix in $\mathbb{D}_{N}$. Due from proposition \ref{prop_5}, \\
~\\
Proposition \ref{prop_5}: \textit{The tree to matrix translation in Lemma \ref{lemma_1} is bijective.}\\
~\\
such tree only have one corresponding matrix $\mathbf{D}\in \mathbb{D}_{N}$, it is contradictory to $\mathbf{D}\notin \mathbb{D}_{N}$. So such $\mathbf{D}$ does not exist.\\
\textit{end of proof.}
\section{proof of Proposition \ref{prop_6}} \label{sec_APP12}
\noindent
We first prove a lemma and its corollary.
\begin{lemma} 
If $\mathbf{D} \in \left\{0,1\right\}^{n \times n}$ satisfies $\left \{ 1^{\circ},5^{\circ} \right \}$, then for any $r \in row(\mathbf{D})$, there exists a $c \in col(\mathbf{D})$ such that $r \in J_{S_c}$.
\label{Lemma_4}
\end{lemma}

\textit{Proof:} For any $r \in row(\mathbf{D})$, by $1^{\circ}$, there exist $c \in col(\mathbf{D})$ such that $r \in S_c$. From all the columns $c$ that "includes" row $r$, we pick the $c$ with the minimum number of elements, which is $c_{min} = \arg\min_{c \in col(\mathbf{D}), r \in S_c} |S_c|$. By $5^{\circ}$, such $c_{min}$ is unique. Because if we have another column $c'$, there are three cases:\\
1) $S_{c_{min}}\subseteq S_{c^{'}}$, in such case, $\left |S_{c^{'}}  \right |\geq  \left | S_{c_{min}} \right |$, equality holds when $S_{c^{'}}  = S_{c_{min}}$. So, as we shall pick the set with minimum number of elements, we will still pick $S_{c_{min}}$ unless they are the same (where it is still unique).\\
2) $S_{c_{min}}\supseteq S_{c^{'}}$. In this case, because $c_{min}$ can be chosen arbitrarily, we can choose $c^{'}$ as the new $c_{min}$, which is still unique.\\
3) $S_{c_{min}}\cap S_{c^{'}}=\varnothing$, apparently $r\notin S_{c^{'}}$\\
As it has the minimum number of elements, $S_c$ cannot have a proper set that also includes row $r$ (though it can still have proper sets). Thus, $r$ only belongs to $c$ but not to any proper set of $c$, $r \in J_{S_c}$.\\
\textit{end of proof.}\\

\begin{corollary}
Each row $r$ in $\mathbf{D}:\left \{ 1^{\circ},5^{\circ},6^{\circ} \right \}$ or $\mathbf{D}:\left \{ 1^{\circ},3^{\circ} ,5^{\circ} \right \}$ corresponds to a unique $J_{S_c}$.
\label{cor_2}
\end{corollary}
\textit{Proof:} 1) If $\mathbf{D}:\left \{ 1^{\circ},5^{\circ},6^{\circ} \right \}$: Assuming two rows $r_1$, $r_2$ correspond to the same $J_{S_c}$, we will have $\left |J_{S_c}  \right |\geq 2$, which violates $6^{\circ}$.\\
~\\
2) If $\mathbf{D}:\left \{ 1^{\circ},3^{\circ} ,5^{\circ} \right \}$: Assuming two rows $r_1$, $r_2$ correspond to the same $J_{S_c}$, we discuss the relation between $S_c$ and another column set $S_{c'}$. Due to $5^{\circ}$, there are following relations:\\
1) $S_{c}\subseteq S_{c^{'}}$: both $r_1$, $r_2$ belong to $S_{c'}$.\\
2) $S_{c}\supset S_{c^{'}}$: both $r_1$, $r_2$ not belong to $S_{c'}$.\\
3) $S_{c}\cap S_{c^{'}}=\varnothing$: both $r_1$, $r_2$ not belong to $S_{c'}$.\\
4) $S_{c}= S_{c^{'}}$: both $r_1$, $r_2$ belong to $S_{c'}$.\\
So for any $S_{c^{'}}\neq S_{c}$, $r_1$, $r_2$ simultaneously belong to or not belong to $S_{c^{'}}$, which means $r_1$ and $r_2$ are equivalent, which violates $3^{\circ}$\\
\textit{end of proof.}\\
~\\
\textbf{Proposition \ref{prop_6}} \textit{A matrix $\mathbf{D} \in \left\{0,1\right\}^{n \times n}$ satisfies $P_{0}$ $\Leftrightarrow\mathbf{D}$  satisfies $P$.}\\

\textit{Proof:} Such proposition is equivalent to prove $\left \{ 6^{\circ} \right \}\Leftrightarrow \left \{ 3^{\circ},4^{\circ} \right \}$.\\

``$\Rightarrow$'': Let matrix $\mathbf{D}$ satisfies $P_0$, which has $r_1,r_2 \in row(\mathbf{D}), r_1 \neq r_2$, $c_1,c_2 \in col(\mathbf{D}), c_1 \neq c_2$. What we show is when $3^{\circ}$ or $4^{\circ}$ does not hold, $6^{\circ}$ does not hold.

1) When $3^{\circ}$ not holds: Assume $r_1 = r_2$, e.e., then for any $c \in col(\mathbf{D})$, $r_1$ and $r_2$ must be simultaneously present or absent in $S_c$. Then by Lemma \ref{Lemma_4}, $r_1, r_2$ must be simultaneously present in $J_{S_{c^\prime}}, \exists c^\prime \in col(\mathbf{D})$, contradicts with condition $6^{\circ}$, that each $J_{S_{c^\prime}}$ can only has one element. Thus, $3^{\circ}$ must hold.

2) When $4^{\circ}$ not holds: Assume $c_1 = c_2$, e.e., we must have $J_{S_c1}=J_{S_c2}$, which violates corollary \ref{cor_2} that $J_{S_c}$ for all rows are unique. Thus, $4^{\circ}$ must hold.\\

``$\Leftarrow$'': Let matrix $\mathbf{D}$ satisfies $P$ and assume $6^{\circ}$ does not hold. We discuss two different cases of violation of $6^{\circ}$ separately:

1) If there exists $c$ such that $|J_{S_c}| \geq 2$, due to corollary \ref{cor_2}'s proof, it leads to violation to $3^{\circ}$

2) If there exists $c$ such that $|J_{S_c}| < 1$, by Lemma \ref{Lemma_4}, there are totally $n$ rows belongs to $n-1$ non-empty Unique Element Sets (eliminate column $c$), implying at least one of $J_{S_c}$ have at least 2 rows, which is the same case as $|J_{S_c}| \geq 2$.\\
\textit{end of proof.}
\section{Demonstration of Matrix to Out-tree Translation Algorithm} \label{sec_APP13}
\begin{figure*}[!t]
 \centering
 \includegraphics[width=.7\linewidth]{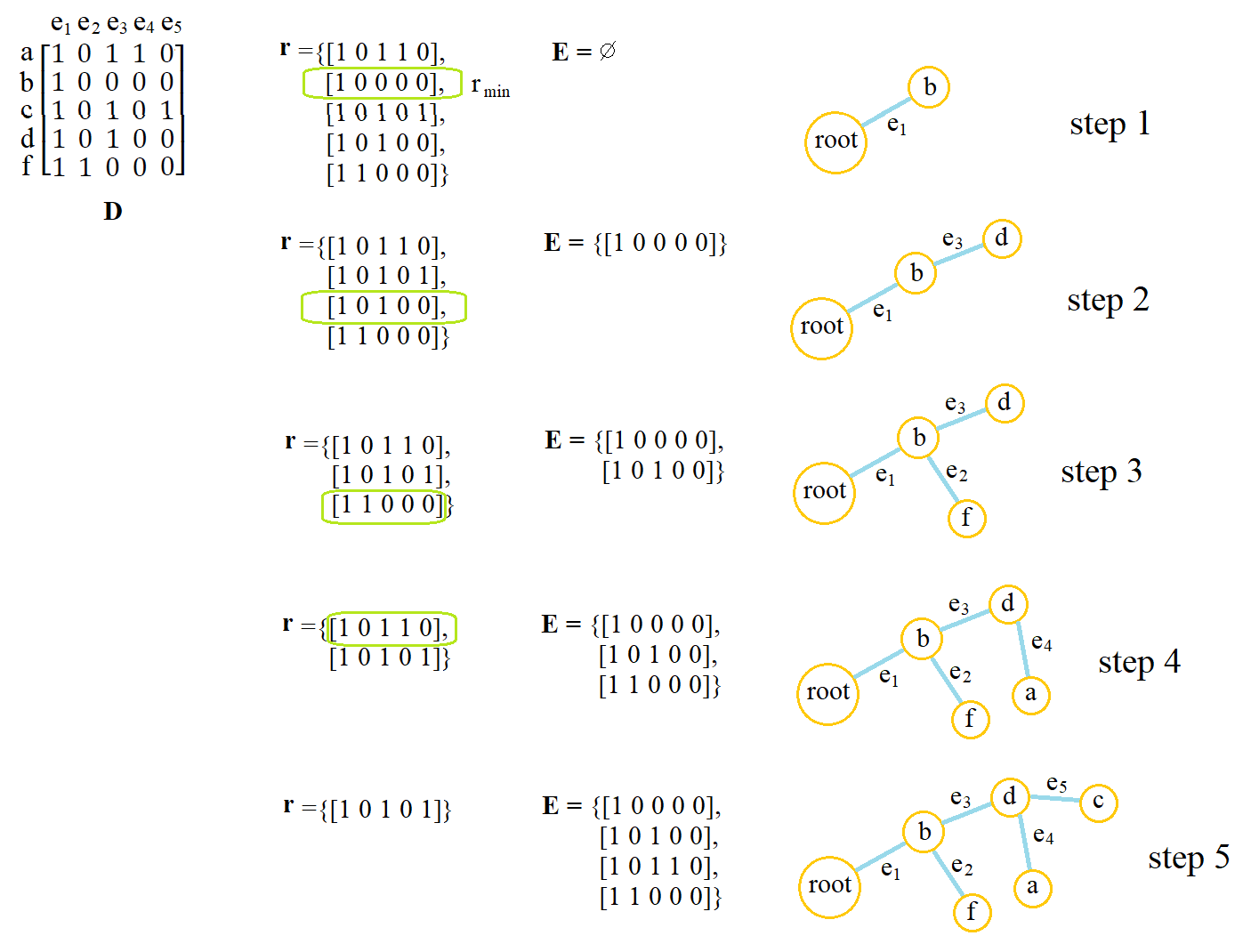}
 \caption{Demonstration of Matrix to Out-tree Translation Algorithm.}
 \label{fig_56}
\end{figure*}

\noindent
We show a demonstration of running the algorithm in Figure \ref{fig_56}.
Assuming we have a heterogeneous matrix 
\begin{equation}
    \mathbf{D}=\begin{bmatrix}
1 & 0 & 1 & 1 & 0\\ 
1 & 0 & 0 & 0 & 0\\ 
1 & 0 & 1 & 0 & 1\\ 
1 & 0 & 1 & 0 & 0\\ 
1 & 1 & 0 & 0 & 0
\end{bmatrix}
\end{equation}
with row labels $\left \{ a,b,c,d,f \right \}$ and column labels $\left \{ e_{1},e_{2},e_{3},e_{4},e_{5} \right \}$. The row set $r$ is the result of decomposing matrix $\mathbf{D}$ by rows, $r=\bigl\{ \left[ 1 , 0 , 1 , 1 , 0 \right] $, $\left [ 1 , 0 , 0 , 0 , 0 \right ]$, $\left[ 1 , 0 , 1 , 0 , 1 \right]$, $\left[ 1 , 0 , 1 , 0 , 0 \right]$, $\left[ 1, 1 , 0 , 0 , 0 \right] \}$, explored set $E=\varnothing $. We can calculate the depths of $ a,b,c,d,f$ as the sum of their corresponding rows $r_{a}=3$, $r_{b}=1$, $r_{c}=3$, $r_{d}=2$, $r_{f}=2$.\\
~\\
\textbf{step 1:} pick the node with minimum depth $r_{min} = r_{b}$. As the depth $r_{b}=1$, we directly attach $b$ to the root node. In $r_{b}$, the column label with '1' indicates the parent edge of $b$, which is $e_{1}$. Remove $r_{b}$ from $r$ and add it to $E$.\\
~\\
\textbf{step 2:} pick the node with minimum depth $r_{min} = r_{d}$ (also can pick $f$, does not affect the result). Search in $E$, where $r_{d}$ has one more '1' than any row in $E$. it gives $r_{b}$, which means node $d$ should be a child of node $b$. The unique '1' $r_{d}$ compared with $r_{b}$ indicates the parent edge of $d$, which is $e_{3}$. Remove $r_{d}$ from $r$ and add it to $E$.\\
~\\
\textbf{step 3:} pick the node with minimum depth $r_{min} = r_{f}$. Search in $E$, where $r_{f}$ has one more '1' than any row in $E$. it gives $r_{b}$, which means node $f$ should be a child of node $b$. The unique '1' $r_{f}$ compared with $r_{b}$ indicates the parent edge of $f$, which is $e_{2}$. Remove $r_{f}$ from $r$ and add it to $E$.\\
~\\
\textbf{step 4:} pick the node with minimum depth $r_{min} = r_{a}$. Search in $E$, where $r_{a}$ has one more '1' than any row in $E$. it gives $r_{d}$, which means node $a$ should be a child of node $d$. The unique '1' $r_{a}$ compared with $r_{d}$ indicates the parent edge of $a$, which is $e_{4}$. Remove $r_{a}$ from $r$ and add it to $E$.\\
~\\
\textbf{step 5:} pick the node with minimum depth $r_{min} = r_{c}$. Search in $E$, where $r_{c}$ has one more '1' than any row in $E$. it gives $r_{d}$, which means node $c$ should be a child of node $d$. The unique '1' $r_{c}$ compared with $r_{d}$ indicates the parent edge of $c$, which is $e_{5}$. Remove $r_{c}$ from $r$ and add it to $E$.\\
~\\
\textbf{return:} heterogeneous out-tree.
\section{proof of Theorem \ref{theo_2}} \label{sec_APP14}
\noindent
\textbf{Theorem \ref{theo_2}} \textit{$\mathbf{D}:P\Leftrightarrow\mathbf{D}$ \textit{can be grown to a heterogeneous out-tree.}}\\

\textit{Proof:} ``$\Rightarrow$'': The meaning of $J_{S_c}$ is the node's label which is the direct child of column $c$ (or edge $c$). $|J_{S_c}| = 1$ indicates for each edge, its direct child has be determined. Thus, we can extract $N$ different pairs of "edge-node" from such $\mathbf{D}$. We can construct such tree by first attaching all edges that has maximum $\left | S_c \right |$ to the root. Then attach the edges with $S_c$ which has no intersection to other $S_{c'}$ also to the root. For any column $c$ left, whose $S_c$ must be a subset of some $S_{c'}$ that already connected to the root, because we have already attached all the edges whose $S_c$ share no intersection to others, that is the third case in $5^{\circ}$. Then we can attach the rest $c$ by a decreasing order of $\left | S_c \right |$ to the tree, specifically to $c'$ that $S_{c}\subseteq S_{c'}$. In such way, we can generate a tree from $\mathbf{D}:P$.\\
~\\
``$\Leftarrow$'': From Lemma \ref{lemma_1}\\
~\\
Lemma \ref{lemma_1}: \textit{From each heterogeneous out-tree, one can extract a determined heterogeneous dependency matrix.}\\
~\\
we know for a heterogeneous out-tree, there is a $\mathbf{D}$ can be generated. We show such $\mathbf{D}$ fulfills condition set $P$.\\
$1^{\circ}$: An all-0 row means a node that has no path to the root, it is not allowed in out-tree.\\
$2^{\circ}$: An all-0 column means an edge with no descendent node, such open edge is not allowed in out-tree.\\
$3^{\circ}$: Duplicate rows indicates the same node, which can be merged when constructing $\mathbf{D}$ without changing tree's topology.\\
$4^{\circ}$: Duplicate columns indicates the two sub-trees rooted at the two edges include the same set of nodes. As it is a tree structure, such sub-tree can only have one path to the root, which means the edges should be the same edges, they can still be merged.\\
$5^{\circ}$: If a node $a$ whose parent edge is $\theta _{1}$, is the descendent of another node $b$ whose parent edge is $\theta _{2}$, there should be $S_{c\theta _{1}}\subseteq S_{c\theta _{2}}$. If $b$ is a descendent of $a$, then $S_{c\theta _{1}}\supseteq  S_{c\theta _{2}}$. If $a$, $b$ are not on the same branch, apparently $\ S_c1 \cap  S_c2=\varnothing$. Because there cannot be a case that $\ S_c1$ and $\ S_c2$ share some intersections, at least for some common nodes in the two sub-trees from edges $\theta _{1}$ and $\theta _{2}$, the nodes has two parents $\theta _{1}$ and $\theta _{2}$.\\
\textit{end of proof.}
\section{Matrix Completion Algorithm example} \label{sec_APP15}
\noindent
We use an example here to show how such algorithm works. Considering matrix\\
\begin{equation}
    \mathbf{D}^{-}=\begin{bmatrix}
1 & 1 & 1 & 1 & 1 & 1\\ 
1 & 1 & 1 & 0 & 0 & 0
\end{bmatrix}
\end{equation}
\begin{figure}[!t]
 \centering
 \includegraphics[width=\linewidth]{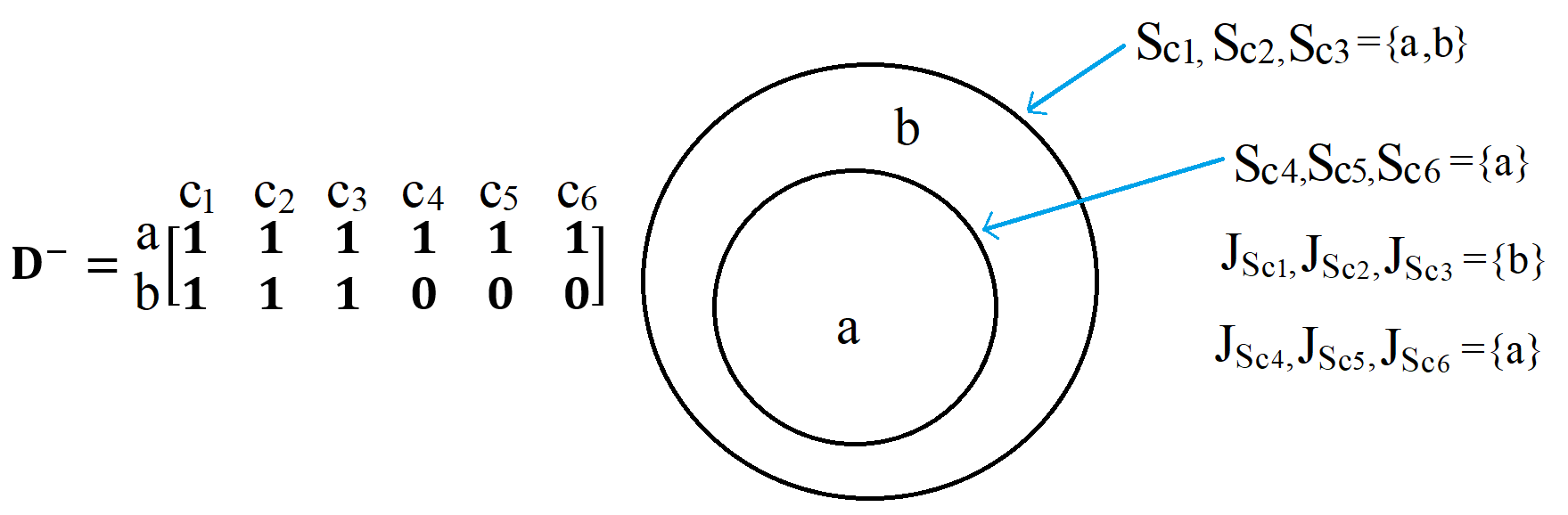}
 \caption{$\mathbf{D}^{-}$ (left) and $\mathbf{D}^{-}$ in $S_{c}$ space (right)}
 \label{fig_12}
\end{figure}
Such matrix can be extracted from robot with 6 joints (number of columns) and 2 sensors (two rows). One sensor depends on all 6 joints (it is mounted on the arm extremity) and the other sensor only depends on only the first three joints (it is mounted on the third link). There are 4 links that with no sensor mounted, so they are unobservable. The complete form of such matrix should be $6\times 6$, because in fully observable case, each link is at least with one sensor mounted, and 6 joints correspond to 6 links except root. So we assume the original fully-observable matrix is $6\times 6$. We can assign any name to rows and columns, e.g., we assign rows as $\left [ a,b,c,d,e,f \right ]$ and columns $\left [ c_{1},c_{2},c_{3},c_{4},c_{5},c_{6} \right ]$. Among these, rows $c,d,e,f$ are not observable. It is easy to find $\mathbf{D}^{-}$ fulfills $P^{-}$, so there has to be a way to fill it to a $\mathbf{D}:P$ of size $N*N$. We draw $\mathbf{D}^{-}$ in $S_{c}$ space with $S_{c}$ and $J_{S_c}$ for each column in Figure \ref{fig_12}.

We now have two sets with $K$ unique $J_{S_c}$, as $c_{1},c_{2},c_{3}$ have the same $J_{S_c}$ and $c_{4},c_{5},c_{6}$ have the same $J_{S_c}$. We can pick one with $\left |J_{S_c}  \right |>0$ from each set, e.g., $\left \{ c_{1},c_{4} \right \}$. The rest of other columns becomes another set $\left \{ c_{2},c_{3},c_{5} ,c_{6}\right \}$. According to Corollary \ref{cor_2}, $J_{S_c}$ of different columns should be different, which is not true for now, we need to assign dependency of unused nodes to the columns in set $\left \{ c_{2},c_{3},c_{5} ,c_{6}\right \}$ to make $J_{S_c}$ unique. Also, we have a set of unused nodes $\left \{ c,d,e,f\right \}$.

\begin{figure}[!t]
 \centering
 \includegraphics[width=\linewidth]{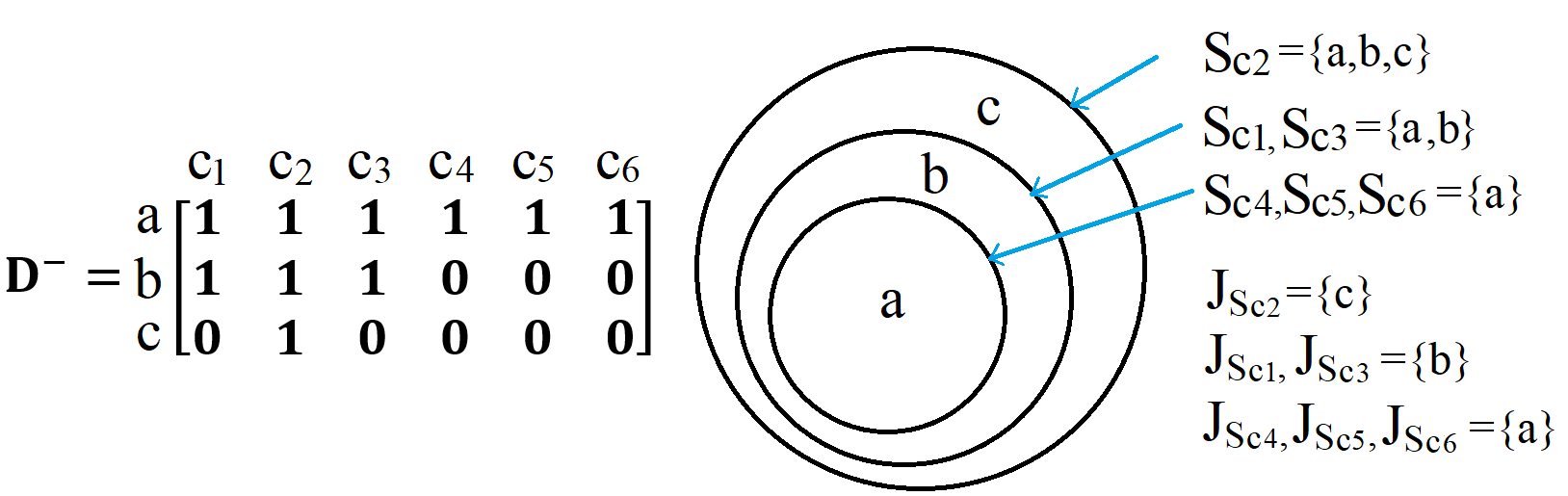}
 \caption{$\mathbf{D}^{-}$ (left) and $\mathbf{D}^{-}$ in $S_{c}$ space (right)}
 \label{fig_13}
\end{figure}

Then for each column in $\left \{ c_{2},c_{3},c_{5} ,c_{6}\right \}$, we find a unused node (a row) in unused set $\left \{ c,d,e,f\right \}$ and assign dependency to the column $c_{i}$ and all $c_{j}$s as $S_{c_{j}}$ is the proper super-set of $S_{c_{i}}$. For example, we first have $c_{2}$, and pick one node from unused node set, e.g. node $c$. Then We add dependency of $c$ with $c_{2}$ and columns that have proper super-set of $S_{c_{2}}$. From Figure \ref{fig_12}, we see, $S_{c_{2}}$ has no proper super-set. So the new matrix and new $S_c$ representation becomes Figure \ref{fig_13}:\\

\begin{figure}[!t]
 \centering
 \includegraphics[width=\linewidth]{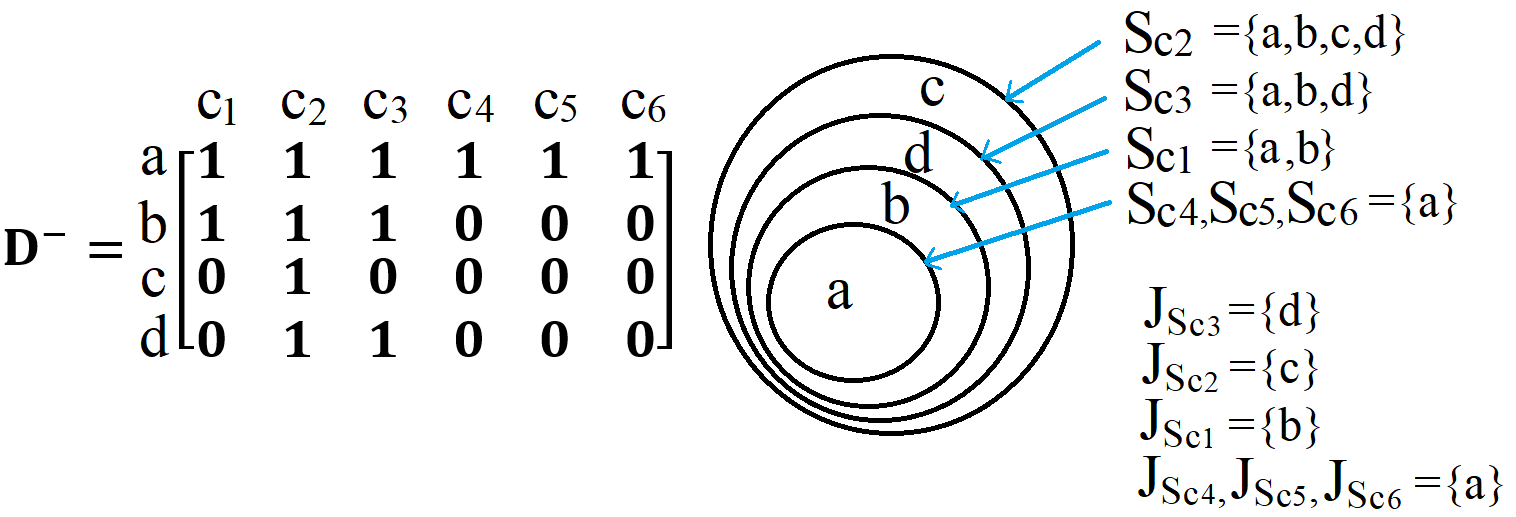}
 \caption{$\mathbf{D}^{-}$ (left) and $\mathbf{D}^{-}$ in $S_{c}$ space (right)}
 \label{fig_15}
\end{figure}

Eliminating $c$ and $c_{2}$ from sets above, we have to pick a new pair from $\left \{c_{3},c_{5} ,c_{6}\right \}$ and $\left \{ d,e,f\right \}$. Let us pick $c_{3}$ and $d$. Then we can add dependency of $d$ with $c_{3}$ and columns that have proper super-set of $S_{c_{3}}$. From Figure \ref{fig_13}, we see the proper super-set of $S_{c_{3}}$ is $S_{c_{2}}$, so we add a new row to $\mathbf{D}^{-}$, the result is shown in Figure \ref{fig_15}\\

\begin{figure}[!t]
 \centering
 \includegraphics[width=\linewidth]{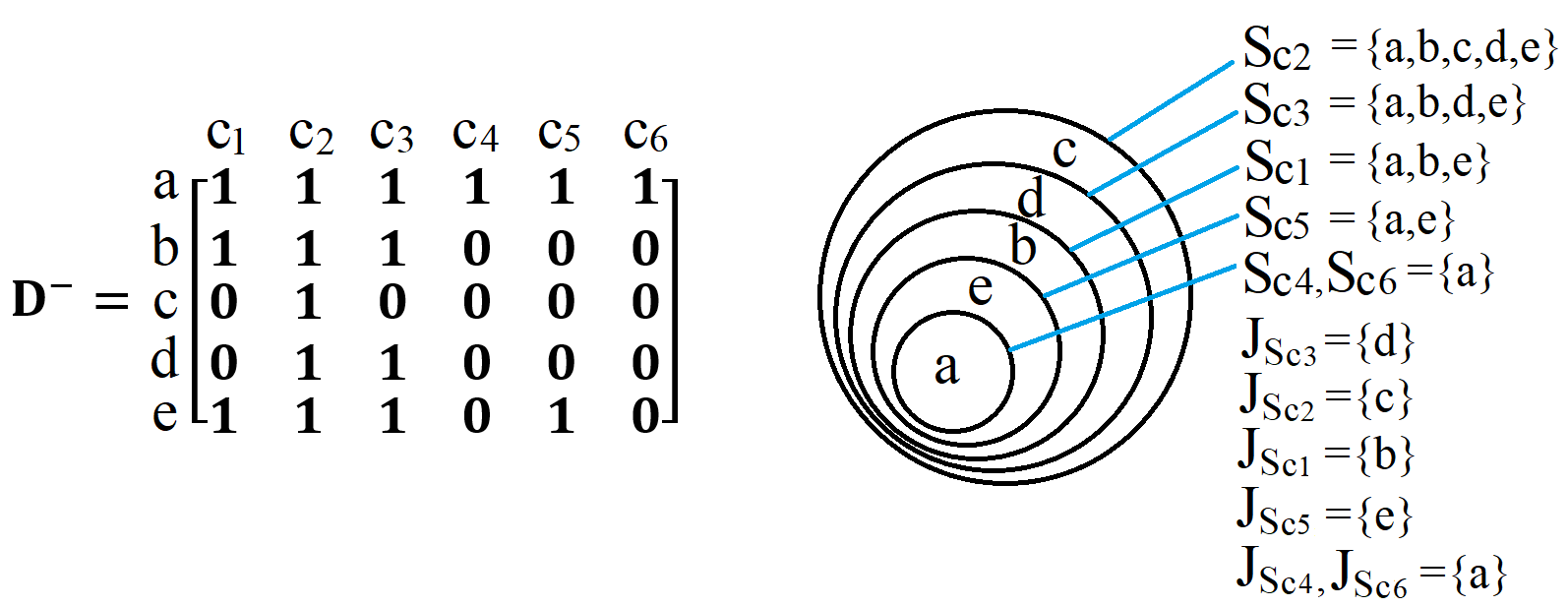}
 \caption{$\mathbf{D}^{-}$ (left) and $\mathbf{D}^{-}$ in $S_{c}$ space (right)}
 \label{fig_16}
\end{figure}

Eliminating $d$ and $c_{3}$ from sets above, we have to pick a new pair from $\left \{c_{5} ,c_{6}\right \}$ and $\left \{ e,f\right \}$. Let us pick $c_{5}$ and $e$. Then we add dependency of $e$ with $c_{5}$ and columns that have proper super-set of $S_{c_{5}}$. From Figure \ref{fig_13}, we see the proper super-set of $S_{c_{3}}$ are $S_{c_{1}}$, $S_{c_{2}}$ and $S_{c_{3}}$, so we add a new row to $\mathbf{D}^{-}$, the result is shown in Figure \ref{fig_16}\\

\begin{figure}[!t]
 \centering
 \includegraphics[width=\linewidth]{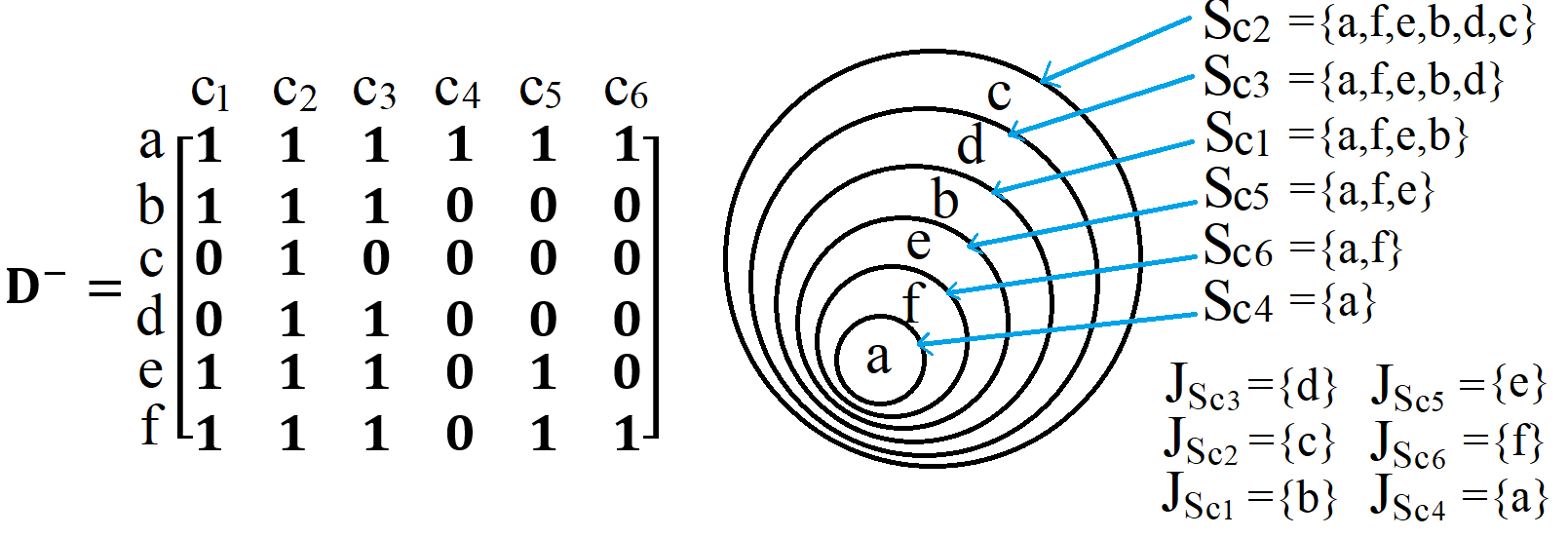}
 \caption{$\mathbf{D}^{-}$ (left) and $\mathbf{D}^{-}$ in $S_{c}$ space (right)}
 \label{fig_17}
\end{figure}

Eliminating $e$ and $c_{5}$ from sets above, we only have $c_{6}$ and $f$. Then we add dependency of $f$ with $c_{6}$ and columns that have proper super-set of $S_{c_{6}}$. From Figure \ref{fig_15}, we see the proper super-set of $S_{c_{6}}$ are $S_{c_{1}}$, $S_{c_{2}}$, $S_{c_{5}}$ and $S_{c_{3}}$, so we add a new row to $\mathbf{D}^{-}$, the result is shown in Figure \ref{fig_17}. Until now, we have successfully filled the $N*N$ matrix. Reader can check such matrix is under condition set $P$.\\
\section{proof of Theorem \ref{theo_3}} \label{sec_APP16}
\noindent
\begin{lemma}
If we have $\mathbf{D}^{-}:P^{-}$, for any column $c$ in $\mathbf{D}^{-}$, $|J_{S_c}| \leq 1$
\label{lemma_6}
\end{lemma}

\textit{Proof:} We prove it by showing when there is a column $c$ that $|J_{S_c}| \geq 2$, $P^{-}$ will be violated. If $|J_{S_c}| \geq 2$, there are two rows belong to the same $J_{S_c}$, this is equivalent to corollary \ref{cor_2}'s proof 2) case, it violates condition $3^{\circ}$ in $P^{-}$.\\
\textit{end of proof.}\\

\begin{corollary}
If we have $\mathbf{D}^{-}:P^{-}$, there exists $K$ columns that $|J_{S_c}| = 1$ and the $J_{S_c}$ are unique.
\label{cor_3}
\end{corollary}

\textit{Proof:} Because there are $K$ non-all-0 rows, by Lemma \ref{Lemma_4},\\
~\\
Lemma \ref{Lemma_4}: \textit{If $\mathbf{D} \in \left\{0,1\right\}^{n \times n}$ satisfies $\left \{ 1^{\circ},5^{\circ} \right \}$, then for any $r \in row(\mathbf{D})$, there exists a $c \in col(\mathbf{D})$ such that $r \in J_{S_c}$.}\\
~\\
there must be $K$ non-empty $J_{S_c}$. Also by Lemma \ref{lemma_6}, the $J_{S_c}$ have $|J_{S_c}| \leq 1$. Because they are non-empty, $|J_{S_c}| = 1$. So each $J_{S_c}$ only contains one row, and for every $J_{S_c}$, they are different.\\
\textit{end of proof.}\\
~\\
\textbf{Theorem \ref{theo_3}} \textit{
A $\mathbf{D}^{-}:P^{-}$ can be filled to a $\mathbf{D}:P$. Such filling is unique when $N-K=1$ and there is one $J_{S_c}$ is $\varnothing$.}\\

\textit{Proof:} ``$\Leftarrow$'': If the original $\mathbf{D}$ fulfills $P$, the sub-matrix $\mathbf{D}^{-}$ must fulfill $P^{-}$. As $\mathbf{D}^{-}$ is a sub-matrix of $\mathbf{D}:P$, we consider which conditions in $P$ is inherited to $P^{-}$. A sub-matrix is to eliminate several rows in original matrix. Because rows are independent, all the conditions for rules are inherited, $1^{\circ}$ ,$3^{\circ}$. Eliminating rows will cause columns to lose some elements, it is possible to create duplicate or all-0 columns, so $2^{\circ}$ ,$4^{\circ}$ do no more apply. Apparently if in original matrix if we have $5^{\circ}$, the three relations $S_c1\subseteq S_c2$ , $S_c1\supseteq  S_c2$, $\ S_c1 \cap  S_c2=\varnothing$ in $5^{\circ}$ should also fit in $\mathbf{D}^{-}$ Thus, $P^{-}=\left \{ 1^{\circ}, 3^{\circ}, 5^{\circ} \right \}$.\\
~\\
``$\Rightarrow$'': As we demonstrate a constructive algorithm in this section to fill $\mathbf{D}^{-}$, we only need to show the completed $\mathbf{D}$ is under condition $P$ or $P_{0}$. From Corollary \ref{cor_3} we know initially there are $K$ non-all empty rows, and there can be all-zero columns. As we pick $K$ columns at the first step of the algorithm with unique $J_{S_c}$, the rest $N-K$ columns should have the same $J_{S_c}$ with one of the picked $K$ columns (because $|J_{S_c}| \leq 1$ by Lemma \ref{lemma_6}) or it is empty. That means we have to assign the rest $N-K$ unused nodes to the $N-K$ columns to make their $J_{S_c}$ different from the picked $K$ columns, that ensures condition $1^{\circ}$ and $2^{\circ}$. In $S_c$ space, it is equivalent to assign nodes to the overlapped $S_c$ sets or sets with no $J_{S_c}$. As in Figure \ref{fig_17}, the sets should be non-intersection and has its only one $J_{S_c}$ node, which ensures condition $6^{\circ}$. As we only extend overlapping $S_c$ sets or sets with no $J_{S_c}$, which ensures condition $5^{\circ}$. Thus, the completed $\mathbf{D}$ is under condition set $P_{0}$, equivalently $P$.

Because we have to assign $N-K$ nodes to $N-K$ columns, there are different ways of such assignment. The permutation causes the assignment not unique. Such uniqueness is only fulfilled by assign one node to one column, that is $N-K=1$. Also there should be one $J_{S_c}$ is $\varnothing$, because if $J_{S_c}$ is not $\varnothing$, $c$ must be a duplicate column (has the same $S_c$) of some column $c'$ in the picked $K$ columns. As the two columns are indistinguishable, one can pick either $c$ or $c'$ in the first $K$ columns, that causes two different assignments. Thus, the ununiqueness comes from two sources, one is picking duplicate columns in the first $K$ columns, the second is assign the rest $N-K$ unused nodes to the $N-K$ columns.\\
\textit{end of proof.}\\
\section{Matrix Correction Algorithm Example} \label{sec_APP17}
\begin{figure}[!t]
 \centering
 \includegraphics[width=.5\linewidth]{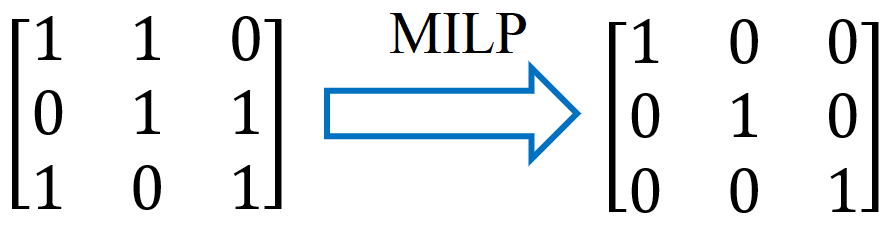}
 \caption{$\mathbf{D}$ and $\mathbf{I}$ by MILP}
 \label{fig_20}
\end{figure}

\noindent
We use an example to show how the algorithm works. The original matrix $\mathbf{D}$ and $\mathbf{I}$ calculated by MILP are shown in Figure \ref{fig_20}. It generates the most similar permutation matrix $\mathbf{I}$ to the original matrix $\mathbf{D}$.

\begin{figure}[!t]
 \centering
 \includegraphics[width=\linewidth]{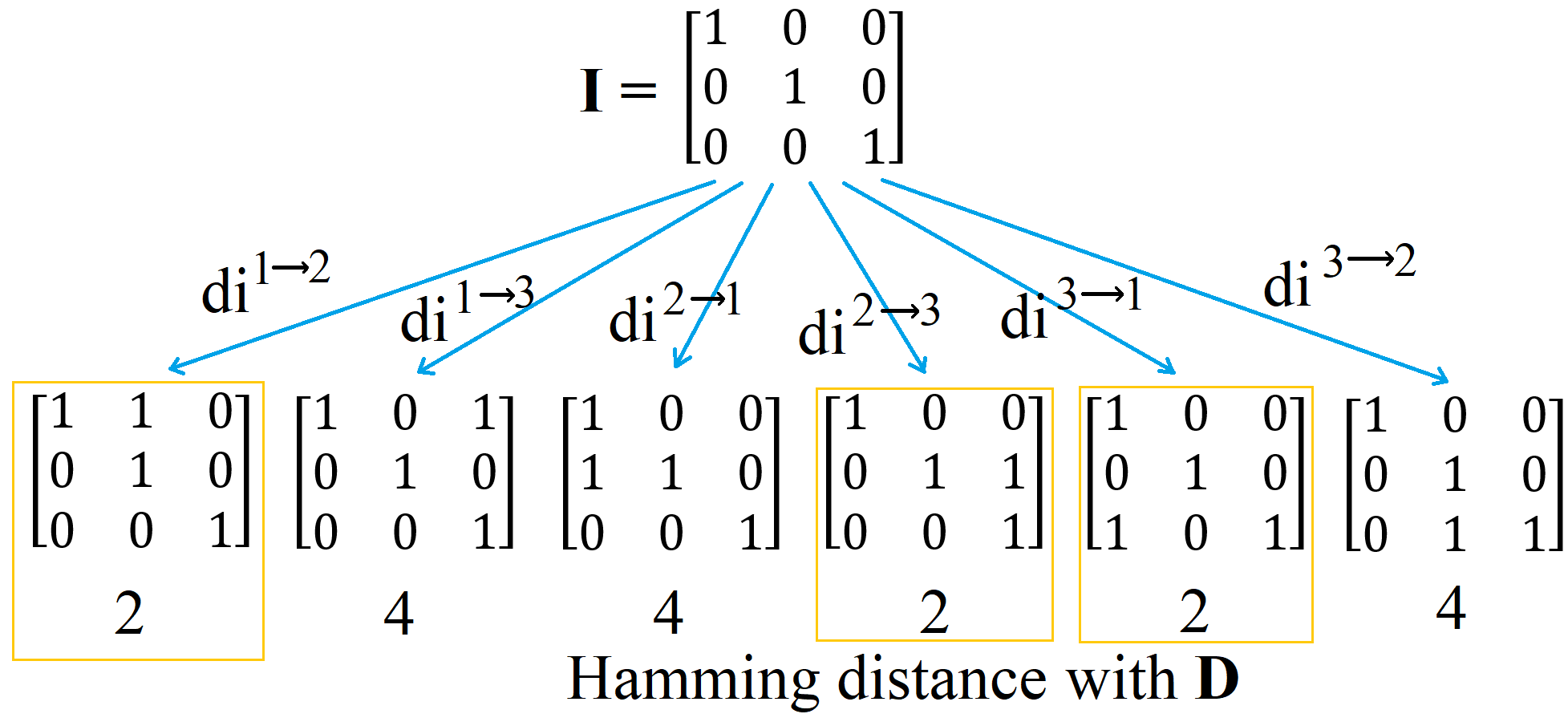}
 \caption{One-step dilation result from Trellis algorithm. Yellows are optimal candidates of $\mathbf{D}^{-}$}
 \label{fig_22}
\end{figure}

\begin{figure}[!t]
 \centering
 \includegraphics[width=\linewidth]{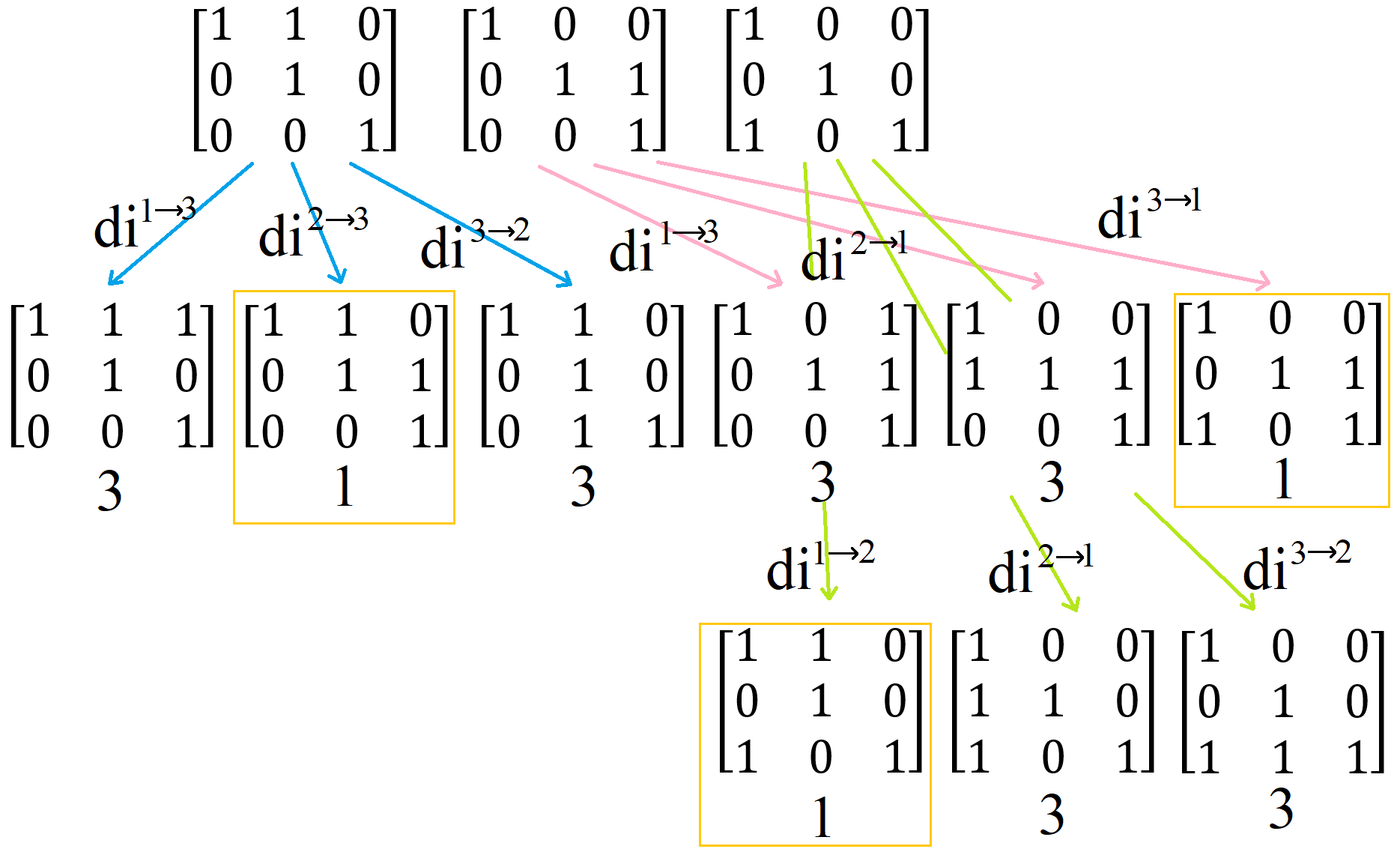}
 \caption{One-step dilation result from Trellis algorithm.}
 \label{fig_23}
\end{figure}

We then generate all possible one-step dilation matrices as in Figure \ref{fig_22}, calculate their Hamming distance to $\mathbf{D}$ and pick the matrices with smallest Hamming distance (yellow ones).
Running one more step dilation will lead to Figure \ref{fig_23}. The minimal Hamming distance now reduce to 1. Continuing with more dilation operations will not reduce the Hamming distance anymore, so the algorithm stops at this step. There are three candidate matrices with equal Hamming distances, we can pick any of these as the correct matrix. For a partially observed matrix $\mathbf{D}^{-}$, we can first fill it to a $N\times N$ matrix by set all missing elements as zero and then perform the correction algorithm.
\end{appendices}

\begin{IEEEbiography}[{\includegraphics[width=1in,height=1.25in,clip,keepaspectratio]{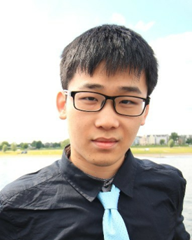}}]{Shuo Jiang}
Shuo Jiang studied control, Microsystems and Microelectronics at
the University of Bremen, Germany, and received the M.Sc. degree in 2018. Currently, he is working towards the Ph.D. degree at the Northeastern University, USA. His research interests lie in the areas of robot tactile perception, robot body schema learning. \end{IEEEbiography}

\begin{IEEEbiography}[{\includegraphics[width=1in,height=1.25in,clip,keepaspectratio]{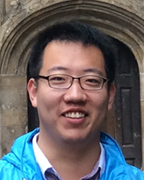}}]{Jinkun Zhang}
Jinkun Zhang studied Microelectronics at
the Fudan University, and received the B.Sc. degree in 2017. Currently, he is working towards the Ph.D. degree at the Northeastern University, USA. His research interests lie in the areas of network optimization, distributed storage and distributed computing. \end{IEEEbiography}

\begin{IEEEbiography}[{\includegraphics[width=1in,height=1.25in,clip,keepaspectratio]{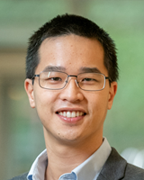}}]{Lawson L.S. Wong}
Lawson L.S. Wong is an assistant professor in the Khoury College of Computer Sciences at Northeastern University. His research focuses on learning, representing, and estimating knowledge about the world that an autonomous robot may find useful. Prior to Northeastern, Lawson was a postdoctoral fellow at Brown University. He completed his PhD at the Massachusetts Institute of Technology. He has received a Siebel Fellowship, AAAI Robotics Student Fellowship, and Croucher Foundation Fellowship for Postdoctoral Research. \end{IEEEbiography}

\end{document}